\documentclass[]{fairmeta}

\usepackage[utf8]{inputenc} 
\usepackage[T1]{fontenc}    
\usepackage{url}            
\usepackage{booktabs}       
\usepackage{amsfonts}       
\usepackage{nicefrac}       
\usepackage{microtype}      
\definecolor{bluelink}{RGB}{0,113,188}
\definecolor{greenlink}{RGB}{0,188,113}
\definecolor{PineGreen}{RGB}{0.0, 0.47, 0.44}
\definecolor{Gray}{RGB}{0.5,0.5,0.5}
\usepackage{listings} 
\usepackage{wrapfig}
\usepackage[most]{tcolorbox}

\newtcolorbox{samplebox}[1]{
    breakable, 
    colback=blue!5!white,
    colframe=blue!50!black, 
    fonttitle=\bfseries,
    title=#1
}
\usepackage{xcolor}
\definecolor{LangCol}{HTML}{F2A65A}
\definecolor{UndCol}{HTML}{5CAF8C}
\definecolor{GenCol}{HTML}{4A7FCE}
\definecolor{citecolor}{HTML}{0071bc}
\hypersetup{
    colorlinks=true,%
    citecolor=citecolor,%
    filecolor=citecolor,%
    linkcolor=citecolor,%
    urlcolor=citecolor
}
\usepackage{tabularx}
\usepackage[most]{tcolorbox}
\usepackage{amsmath}
\usepackage{multirow}
\usepackage{array}
\usepackage{caption}
\usepackage{wrapfig}
\usepackage{enumitem}
\usepackage{tikz}
\usepackage{lipsum}
\usepackage{subcaption}
\usepackage{multirow} 
\usepackage{adjustbox}
\usepackage[table]{xcolor}  
\usepackage{siunitx}          
\usepackage{graphicx}

\usepackage{amssymb}
\renewcommand{\paragraph}[1]{\vspace{1.25mm}\noindent\textbf{#1}}

\newcommand{\finding}[2]{
    \begin{tcolorbox}[
        colback=white!90!gray,     
        colframe=teal!60!black,     
        arc=5pt,                    
        boxsep=5pt,                 
        left=10pt,                  
        right=10pt,                 
        top=2pt,                    
        bottom=2pt,                 
        boxrule=0.8pt,              
        drop shadow=gray!50!white,  
        enhanced jigsaw             
    ]
    \vspace{-0.1cm}
        \paragraph{\textbf{\textit{Finding #1:}}} #2
    \vspace{-0.1cm}
    \end{tcolorbox}
    \vspace{-0.1cm}
}

\newcommand{\suggref}[2]{\hyperref[suggestion:#1]{\textcolor{blue!50!black}{\textit{#2}}}}

\usepackage{pgf}
\usepackage{colortbl}
\usepackage{tipa}
\usepackage{tcolorbox}
\usepackage{rotating}
\usepackage[abs]{overpic}
\usepackage{makecell}
\usepackage{longtable}
\usepackage{tocloft}  
\usepackage[english]{babel}
\usepackage{csquotes}
\usepackage{hyperref}

\newlength\savewidth

\title{Towards Physics of Multimodal Pretraining:
Knowledge Flow, Modality Synergy, Early Unification, and Recipes}

\author[1,3]{Junlin Han}
\author[1]{Shengbang Tong}
\author[1]{David Fan}
\author[2,3]{Minghao Chen}
\author[3]{Philip Torr}
\author[2]{Filippos Kokkinos}
\author[1]{\\[0.25em]Mike Lewis}

\affiliation[1]{FAIR, Meta}
\affiliation[2]{Reality Labs, Meta}
\affiliation[3]{University of Oxford}

\abstract{Vision offers a critical axis for advancing foundation models, driving a shift towards natively unified multimodal pretraining. Despite this momentum, the design space and the fundamental mechanisms of how modalities interact during unified training remain underexplored. We provide empirical clarity through a systematic exploration of multimodal pretraining. Our controlled experiments on both synthetic and large-scale real-world datasets yield four key insights into the physics of multimodal pretraining: \textbf{(i) Knowledge Flow:} We disentangle how language, visual understanding, and visual generation transfer knowledge across modalities, revealing distinct patterns of influence and asymmetry; \textbf{(ii) Synergy vs. Competition:} We show that data "complexity" largely determines whether modalities are synergistic, identify architectural choices that promote synergy—such as shared attention and normalization with modality-specific feed-forward layers, and find that these behaviors generalize across different visual tokenizer designs; \textbf{(iii) Early Unification:} Unifying modalities from the very early stages and training them jointly is shown to be more effective than late alignment or sequential training. This process uncovers a vision laziness phenomenon, where delayed integration leads models to rely on language priors; \textbf{(iv) Recipes:} We derive efficient pretraining recipes that achieve strong generative performance using only 5\% of the compute budget. These core findings are subsequently validated at scale by training multiple 13.5B MoE models on 2T tokens. We hope this study provides a principled foundation for understanding and scaling multimodal pretraining.}

\date{\today}
\correspondence{ \email{junlinhan@meta.com}}

\metadata[Project page]{\url{https://junlinhan.github.io/projects/physics_of_mm_pretrain/}}

\begin{document}

\maketitle

\vspace{3em}
\noindent
\begin{tcolorbox}[
    colback=gray!5!white,      
    colframe=green!40!black,
    arc=4pt,
    boxrule=0.5pt,
    boxsep=3pt,
    left=6pt, right=6pt, top=4pt, bottom=3pt,
    enhanced jigsaw,
    drop shadow=gray!40!white,
]
{\noindent\sffamily\bfseries Highlights at a Glance}\\[-2pt]
{\color{green!50!black}\hrule height 0.3pt}\vspace{4pt}
\begin{enumerate}[leftmargin=1.5em, itemsep=6pt, parsep=0pt, topsep=0pt, label={\small\textbf{S\arabic*.}}]
    \vspace{6pt}
    \item \textbf{Knowledge Flow}: Modality transfer is asymmetric; language and visual understanding act as strong priors to drive visual generation.
    \hfill \hyperref[section:flowstudy]{\textcolor{green!50!black}{\textit{\S\ref*{section:flowstudy}}}}
    
    \item \textbf{Modality Synergy}: Task complexity dictates interaction; shared attention and normalization promote synergy, while decoupled FFNs mitigate competition.
    \hfill \hyperref[sec:synergy]{\textcolor{green!50!black}{\textit{\S\ref*{sec:synergy}}}}
    
    \item \textbf{Early Unification}: Early joint training is critical for multimodal co-evolution; delaying integration triggers a "vision laziness" bias. 
    \hfill \hyperref[sec:earlyunification]{\textcolor{green!50!black}{\textit{\S\ref*{sec:earlyunification}}}}
    
    \item \textbf{Recipes}: Asymmetric data recipes combined with MoE and early unification scale unified pretraining with high efficiency.
    \hfill \hyperref[sec:recipes]{\textcolor{green!50!black}{\textit{\S\ref*{sec:recipes}}}}
\end{enumerate}
\end{tcolorbox}

\clearpage

\tableofcontents

\clearpage

\section{Introduction}
\label{section:intro}

The trajectory of foundation models is evolving from unimodal language to multimodal~\citep{OpenAI2024gpt4o,comanici2025gemini,team2026qwen3}. To integrate vision, the community initially relied on late-fusion~\citep{li2023blip,liu2023visual,alayrac2022flamingo,tong2024cambrian,bai2025qwen3,grattafiori2024llama3}, which aligns pretrained visual encoders with pretrained language models. However, this inherently bottlenecks capabilities, as rich visual signals are forced to conform to a pre-existing language space. By integrating vision from the very beginning of training, early-fusion allows visual and language representations to co-evolve, unlocking deeper and more native visual understanding capacities~\citep{team2026kimi,meta2025llama,comanici2025gemini,thinkingmachines2026interactionmodels}. Yet, the evolution does not stop here. The field is now moving toward unified multimodal models~\citep{yang2026omni,deng2025bagel,tong2026beyond,meituanlongcatteam2026longcatnextlexicalizingmodalitiesdiscrete,wang2026ernie,janus,liu2025tuna,han2026vision,team2024chameleon,li2025manzanosimplescalableunified,wang2026arm}. This new paradigm pushes the ambition beyond mere text and visual understanding, treating visual generation as a core, simultaneous objective within a single model.

As we enter unified pretraining, the complexity of the design space has further increased, leaving the field to navigate largely by heuristics~\citep{tong2026beyond}. The fundamental mechanisms, or the underlying ``physics''~\citep{AllenZhu-icml2024-tutorial}, that govern unified pretraining remain underexplored. This gap is further exacerbated by the prevailing paradigm, where most existing approaches construct unified models by retrofitting pretrained LLMs/MLLMs~\citep{lmfusion,deng2025bagel,showo,liu2026tuna,diao2026sensenovau1unifyingmultimodalunderstanding}. By predominantly focusing on appending vision capabilities to an LLM, they treat vision more as a module to be aligned post-hoc. Consequently, the dynamics of how vision should actively participate in and shape the foundational pretraining remain obscured.
In this work, we aim to replace intuition with evidence, providing a systematic, bottom-up exploration of unified pretraining. Through rigorously controlled experiments across synthetic environments and large-scale real-world datasets, we isolate the fundamental behaviors of multimodal learning. \emph{We seek to establish a principled foundation for how to design, unify, and scale the unified multimodal models.}

Our exploration yields four insights into multimodal pretraining: 

 \begin{itemize}

\item \textbf{Knowledge Flow (\S~\ref{section:flowstudy}).} We begin by examining the general knowledge flow among language, visual understanding, and visual generation using real-world data, revealing a starkly asymmetric transfer between these capabilities. However, the exact interplay between visual understanding and generation has been a subject of extensive debate~\citep{wise,zhang2025unified}. To cut through the noise and confounders of real-world distributions, we design a much cleaner setting using strictly controlled synthetic data~\citep{johnson2017clevr}. This allows us to precisely isolate specific capability transfers, demonstrating that cross-modal knowledge flow is also concept-dependent.


\item \textbf{Modality Synergy (\S~\ref{sec:synergy}).}
Next, we delve into the underlying conditions that dictate whether modalities compete or synergize. We first study data and task complexity, revealing that highly simplified language or vision tasks can promote cross-modal synergy and improve the counterpart modality. To dissect the mechanics behind this phenomenon, we analyze the model's Transformer architecture to pinpoint where synergies and collisions emerge within the network, identifying designs such as shared attention to foster synergy and separate feed-forward networks for competition isolation. Finally, we show that these synergies generalize across diverse visual tokenizers rather than depending on a specific visual representation.

\item \textbf{Early Unification (\S~\ref{sec:earlyunification}).}
We also investigate the temporal and strategic dimensions of training dynamics. By systematically varying the exact moment visual data is introduced, we demonstrate the necessity of early unification (when to introduce vision). Furthermore, we establish that simultaneous joint training is imperative (how to schedule modalities), proving that modalities must actively co-evolve rather than being trained in isolated curriculum steps. In doing so, we uncover a "vision laziness" phenomenon, where late alignment causes models to optimize less in vision components and over-rely on pre-existing language priors.

\item \textbf{Recipes (\S~\ref{sec:recipes}).}
Finally, we synthesize these insights into ready-to-use pretraining recipes. We first conduct extensive empirical searches over data mixing ratios to validate our findings on knowledge flow asymmetry. We then integrate this optimized asymmetric data mix with our insights on parameter-sharing architectural designs and early unified training. To evaluate these three main findings at scale, we train 13.5B MoE models on 2T tokens using controlled, single-variable comparisons. These scaled evaluations demonstrate that our recipes scale effectively, providing a practical baseline for future multimodal foundation models.
\end{itemize}

\section{Experimental Setup}
\label{section:formulation}
In this section, we introduce our default training and evaluation settings.

\subsection{Training protocol}

\paragraph{Pretraining setup.} We follow standard practices and pretrain decoder-only Transformer models that closely adhere to the \textbf{Llama-3} architecture~\citep{grattafiori2024llama3}, featuring SwiGLU, RoPE ($\theta=500{,}000$), pre-RMSNorm, grouped-query attention, QK-norm, and FlashAttention. To natively integrate vision and language, we adopt the \textbf{Transfusion} framework~\citep{zhou2024transfusion} that unifies discrete next-token prediction for text and continuous flow matching for visual generation within a single model. The default backbone of all controlled experiments is a 1.5B Llama-3-like model utilizing \textbf{modality-specific split FFNs} for text and image tokens, totaling 2.3B parameters. Specifically, this consists of 16 layers with a hidden dimension of 2048, using Grouped Query Attention (GQA) with 32 query heads, 8 key-value heads, and an FFN expansion multiplier of 1.5.
We use the Llama-3 BPE tokenizer with a vocabulary size of approximately $128{,}000$, augmented with a small set of multimodal control tokens. 

We support four visual tokenization configurations. For the diffusion-based configurations, models are trained with rectified flow and decoded with a 25-step Euler sampler with a classifier-free guidance scale of 5.0. Following recent practice~\citep{li2025jit,tong2026beyond} and our empirical validation, we adopt $x$-prediction: the network outputs the clean sample $x_0$, which is converted on-the-fly to a velocity $v=(x_0-x_t)/(1-t)$ for the loss and each Euler ODE step, rather than directly regressing velocity. (1) \textbf{RAE (default)}: a frozen SigLIP-2 ViT-400m/14 encoder~\citep{siglip,tschannen2025siglip} processes $224\times224$ images into a $16\times16$ grid of 256 semantic tokens used for both understanding and generation, with generation performing flow matching in the SigLIP latent space and decoded by a Representation Autoencoder~\citep{zheng2025diffusion}. (2) \textbf{Raw Pixels}: the encoder/decoder is removed; a $224\times224$ image is patchified by a single $14\times14$ convolution into a $16\times16$ grid of 256 patch tokens, projected by a lightweight MLP to the transformer hidden dimension (with bilinear resampling to align the token count when needed), and the same tokens are consumed for understanding and denoised for generation. (3) \textbf{CLIP\,+\,VAE}: SigLIP-2 is used for understanding at $224\times224$ (256 tokens), while generation operates at $256\times256$ in the Stable Diffusion~3 VAE latent space ($8\times$ downsampling, 16 channels), denoising a $32\times32$ latent grid. (4) \textbf{AR (UniTok)}: to evaluate whether our findings generalize beyond diffusion-based image modeling, we implement an autoregressive configuration using discrete visual codes from UniTok~\citep{jiao2025unitoken}. Images are represented by residual-quantized codes, which are predicted autoregressively across spatial positions. At each position, a dedicated causal depth head predicts the codebook factors sequentially under a cross-entropy objective.

Training is optimized using AdamW with $\beta_1=0.9$, $\beta_2=0.95$, weight decay of $0.1$, and gradient clipping at $1.0$, following a cosine decay schedule with a linear warm-up over the first 8000 steps. To balance cross-modal learning, the continuous diffusion flow-matching loss is up-weighted by a factor of 3.0 relative to the discrete text cross-entropy loss. Training timesteps $t\in[0,1]$ are sampled from a logit-normal distribution. Models are trained with a context length of 4096 tokens in bf16 precision under FSDP-2. Our controlled experiments span training budgets ranging from 100B to 2T tokens. 

\paragraph{Pretraining data.}  Our language data is sourced from DCLM~\citep{li2024datacomplm}. For vision text-image paired data, we use a collection of roughly 350M image-text pairs from Shutterstock-Image (SSTK) as our exclusive source for both image-to-text understanding and text-to-image conditional generation.

\paragraph{VQA fine-tuning setting and data.} Following the visual question answering (VQA) evaluation protocol established by Cambrian-1~\citep{tong2024cambrian}, Web-SSL~\citep{fan2025scaling}, LSBS~\citep{han2025learning}, and Beyond Language Modeling~\citep{tong2026beyond}, we perform one epoch of supervised fine-tuning on the Cambrian-7M instruction dataset after pretraining to obtain visual-understanding scores. Fine-tuning uses AdamW with a peak learning rate of $1\times10^{-5}$, cosine decay, weight decay of $0.1$, and an effective global batch size of 128 sequences. We use the full Cambrian-7M curation as the supervised fine-tuning corpus, mixing language-only and vision-language paired instructions in their original proportions.

For all experiments, we fix the random seed to $0$ for both model initialization and the data iterator, and set the decoding temperature to 0 for evaluation to obtain stable and reproducible comparisons.

\subsection{Evaluation protocol}

\paragraph{Language evaluation.} We measure language capability along two complementary axes: few-shot downstream
accuracy and validation perplexity. The downstream score is the unweighted
average over \textbf{11 standard benchmarks}, comprising nine multiple-choice
reasoning and commonsense tasks scored by accuracy, including ARC-Easy and
ARC-Challenge~\citep{arc-ce}, BoolQ~\citep{clark2019boolq},
CoQA~\citep{reddy2019coqa}, HellaSwag~\citep{zellers2019hellaswag},
OpenBookQA~\citep{openbookqa}, PIQA~\citep{bisk2020piqa},
SIQA~\citep{sap2019social}, and WinoGrande~\citep{sakaguchi2021winogrande}, and two open-ended QA tasks scored by exact match,
NaturalQuestions~\citep{kwiatkowski2019natural} and
TriviaQA~\citep{joshi2017triviaqa}. We also report validation average perplexity
on DCLM (in-distribution) and C4 (out-of-distribution).

\paragraph{VQA evaluation.} We comprehensively evaluate visual capabilities across understanding and generation. For visual understanding, we categorize our 16 benchmarks into four core evaluation axes:

\begin{itemize}[leftmargin=*, nosep]
    \item General: Focuses on basic visual perception and its alignment with commonsense, avoiding complex inferential tasks. Evaluated on GQA~\citep{hudson2019gqa}, MME~\citep{fu2023mme}, MMBench~\citep{liu2023mmbench}, and SEED~\citep{ge2023planting}.

    \item Knowledge: Probes the integration of visual cues with broad world knowledge, demanding multi-step reasoning for scientific and mathematical problem-solving. Covers ScienceQA~\citep{lu2022learn}, MMMU~\citep{yue2023mmmu}, AI2D~\citep{hiippala2021ai2d}, and MathVista~\citep{lu2023mathvista}. 
    
    \item OCR \& Chart VQA: Assesses high-resolution, fine-grained perception necessary for reading dense textual elements and decoding structured charts. Comprises TextVQA~\citep{singh2019towards}, ChartQA~\citep{masry2022chartqa}, DocVQA~\citep{mathew2021docvqa}, and OCRBench~\citep{liu2023hidden}.
    
    \item Vision-Centric: Tests native visual faculties such as spatial understanding, 3D perception, object counting, and abstract visual logic. Benchmarks include RealWorldQA~\citep{grok}, MMVP~\citep{tong2024eyes}, CV-Bench (derived from Cambrian-1~\citep{tong2024cambrian}, consisting of COCO~\citep{lin2014microsoft}, ADE~\citep{zhou2019semantic}, and Omni3D~\citep{brazil2023omni3d}). 
\end{itemize}

\paragraph{Visual generation evaluation.} We measure compositional text-to-image capability using GenEval~\citep{geneval}, and evaluate dense-prompt generation via DPG-Bench~\citep{dpgbench}. For text-image alignment, we report CLIP similarity scores~\citep{clipscore} across three prompt-length buckets—short (<10 words), medium (10-30 words), and long (30-50 words), each comprising a few hundred prompts. Finally, as a proxy for overall generation quality, we report the held-out diffusion loss measured on a validation set containing 1000 samples.

\section{Demystifying Modality Knowledge Flow}
\label{section:flowstudy}

Unified models jointly train on language, image understanding (image $\rightarrow$ text), and image generation (text $\rightarrow$ image). It is still not fully clear how knowledge flows between these three abilities, with recent studies often reaching different conclusions~\citep{zhang2025unified,metaquery,deng2025bagel,zhang2025unified,tong2026reversing,tong2026beyond,shi2026realunify,niu2025does}. In this section we systematically dissect these directions of transfer through controlled mixture experiments. We organize the study in two stages.

We \textbf{first} train on \emph{general, web-scale data} (SSTK image--text pairs) for vision and DCLM~\citep{li2024datacomplm} for language, so that conclusions are grounded in real data (\S~\ref{subsec:part1_start}). While the main paper focuses on RAE-based results, Appendix~\ref{appendix:modalitytransfer} demonstrates that the exact same trends hold for alternative vision encoders and decoders (e.g., Raw Pixels and CLIP + VAE).

\textbf{Second}, because real data conflates many factors at once, we replicate the analysis on a procedurally generated synthetic benchmark. The synthetic setting lets us remove or insert specific concepts at will, and obtain per-concept evaluation, which together turn correlational findings on real data into causal statements about modality transfer (\S~\ref{subsec:clevr}).

\subsection{Modality transfer on real-world data}
\label{subsec:part1_start}
\finding{1}{Modality knowledge flow is asymmetric: Language acts as a universal booster for all visual tasks; visual understanding serves as a strong prior for generation; however, visual generation yields neutral effects on other abilities.}

\subsubsection{Language priors for visual understanding and generation}
\label{subsubsec:part1}

\paragraph{Setting.} We ask whether adding language data on top of a fixed vision allocation helps either visual generation or visual understanding. Models are trained \emph{from scratch} on language and vision data. The vision allocation is held constant at 50B tokens, and language tokens are added on top so that the language ratio sweeps $\{0\%,20\%,40\%,60\%,80\%\}$, yielding added language tokens of $\{0, 12.5, 33, 75, 200\}$~B. For visual generation, we study both unconditional ($p_{\text{rot}}{=}0$, no text condition) and conditional (text $\rightarrow$ image) settings.

\begin{figure*}[!h]
    \centering
    \includegraphics[width=\linewidth]{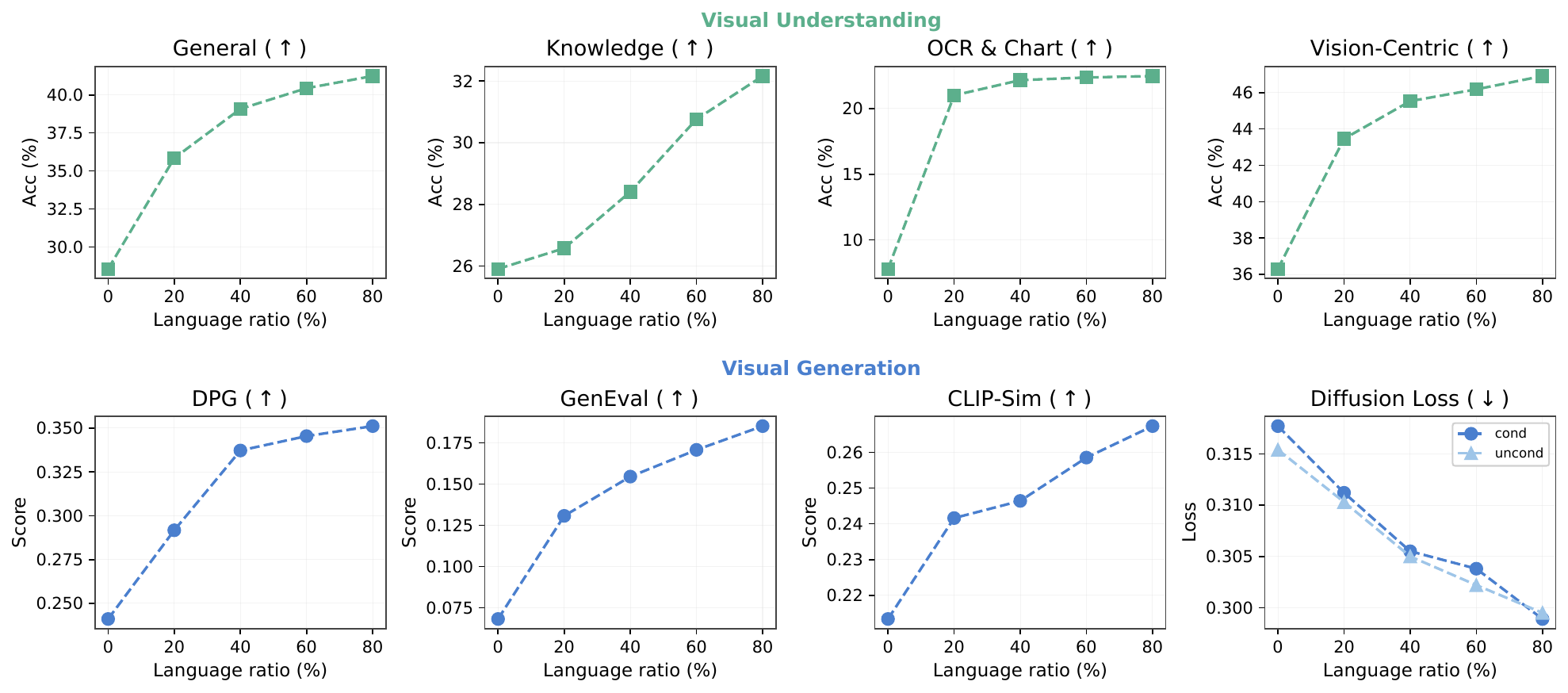}
    \caption{\textbf{Impact of scaling language data on visual understanding and generation.} Increasing the language ratio universally improves both vision capabilities.}
    \label{fig:part1}
\end{figure*}

\paragraph{Results.} As shown in Figure~\ref{fig:part1}, increasing the language data ratio from 0\% to 80\% yields monotonic improvements across all visual understanding evaluation axes. General and Vision-Centric tasks exhibit steady improvements, while Knowledge and OCR \& Chart capabilities see dramatic relative increases. For visual generation, a stronger language prior uniformly enhances text-to-image alignment and compositional generation quality. Notably, the diffusion loss decreases for both conditional and unconditional generation, indicating that joint language pretraining imparts benefits that universally improve the model's native visual modeling capacity. This aligns with recent observations of language visual priors~\citep{han2025learning,wang2025words} and supports the Platonic Representation Hypothesis~\citep{huh2024platonic,huang2025cross,ziyin2025proof}, suggesting that a mature language manifold may naturally capture an underlying, modality-agnostic world structure that bootstraps visual learning.

\subsubsection{Transfer from vision understanding to language and visual generation}
\label{subsubsec:part2}

\paragraph{Setting.} We study the effect of adding visual understanding data on top of a fixed language or visual generation allocation. The base allocation (either language or generation) is held constant at 50B tokens, and visual understanding tokens are added on top so that the understanding ratio sweeps $\{0\%,20\%,40\%,60\%,80\%\}$, yielding added understanding tokens of $\{0, 12.5, 33, 75, 200\}$~B. When studying the effect on language, models are trained from scratch on the mixed data. When studying visual generation, however, we initialize training from a 50B pretrained language model rather than from scratch; this ensures the model possesses very basic text capabilities. As before, we evaluate both unconditional and text-conditional generation variants.

\begin{figure*}[!h]
    \centering
    \includegraphics[width=\linewidth]{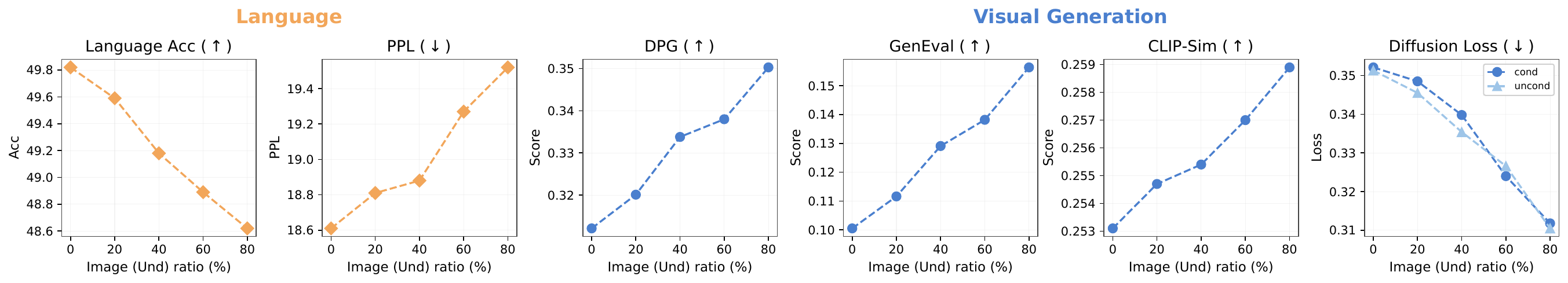}
    \caption{\textbf{Impact of scaling visual understanding data.} It significantly benefits visual generation but degrades pure language performance.}
    \label{fig:part2}
\end{figure*}

\paragraph{Results.} Figure~\ref{fig:part2} illustrates a mild capability trade-off on the language side. As the proportion of visual understanding tokens increases, language benchmark scores experience a slight drop and perplexity marginally worsens. This degradation is largely an artifact of the text distribution in standard vision-language datasets being fundamentally different from that of pure language datasets. Conversely, on the vision side, visual understanding serves as a powerful catalyst for visual generation. Increasing the understanding data ratio markedly improves generation evaluation metrics and significantly reduces both conditional and unconditional diffusion losses. This confirms that the rich, discriminative visual features learned during understanding tasks transfer highly effectively to generative processes.

\subsubsection{Does visual generation help understanding and language?}
\label{subsubsec:part3}

\paragraph{Setting.} We ask whether adding visual generation data on top of a fixed language or visual understanding allocation helps either capability. The base allocation (either language or understanding) is held constant at 50B tokens, and generation tokens are added on top so that the generation ratio sweeps $\{0\%,20\%,40\%,60\%,80\%\}$, yielding added generation tokens of $\{0, 12.5, 33, 75, 200\}$~B. When studying the effect on language, models are trained from scratch on the mixed data; the fixed language compute budget is jointly trained with the added generation tokens (in both unconditional and text-conditional variants). When studying visual understanding, however, we initialize training from a 50B pretrained language model to ensure foundational text abilities. The generation data is then mixed with the fixed understanding allocation, and each pretrained model is subsequently fine-tuned on Cambrian-VQA to measure downstream understanding performance. 

\begin{figure*}[!h]
    \centering
    \includegraphics[width=\linewidth]{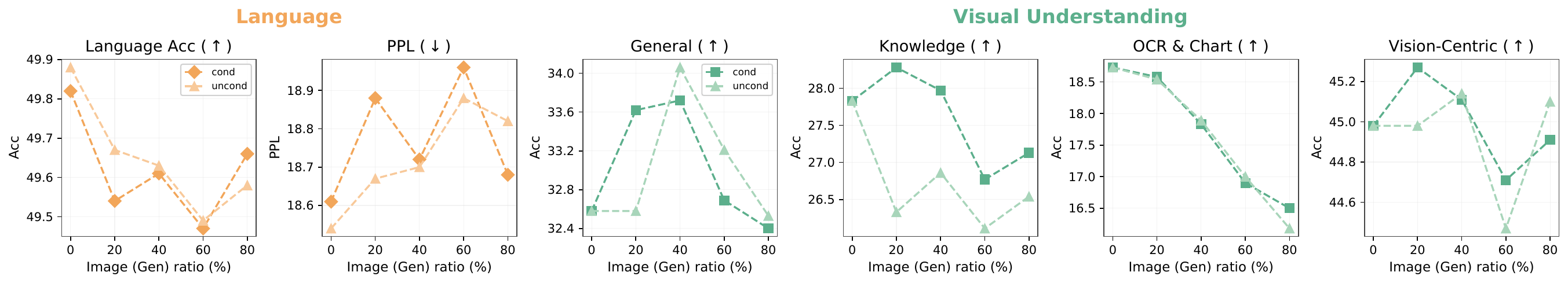}
    \caption{\textbf{Impact of scaling visual generation data.} Adding visual generation causes minor fluctuations in language and most understanding tasks.}
    \label{fig:part3}
\end{figure*}

\paragraph{Results.} As depicted in Figure~\ref{fig:part3}, introducing visual generation data does not demonstrate a positive transfer to other tasks. Scaling the generation ratio leads to minor fluctuations in language benchmark accuracy and perplexity, rather than a significant degradation. Similarly, for visual understanding, performance across the General, Knowledge, OCR \& Chart, and Vision-Centric axes exhibits marginal fluctuations without a distinct trend. Unlike other cross-modal directions, visual generation does not act as a strong catalyst to boost language or understanding performance. However, it also does not severely interfere with them. Generative training objectives (such as flow matching) do not inherently conflict with next token prediction, though they offer limited backward knowledge transfer.

\subsection{Concept transfer: A synthetic controlled study}
\label{subsec:clevr}
\finding{2}{Knowledge flow is concept-dependent. Zero-shot transfer strictly fails for low-level concepts in both directions. However, visual understanding provides zero-shot transfer for structural concepts to generation, while generation provides  latent priors that accelerate low-level understanding.}

\begin{figure*}[!h]
    \centering
    \includegraphics[width=\linewidth]{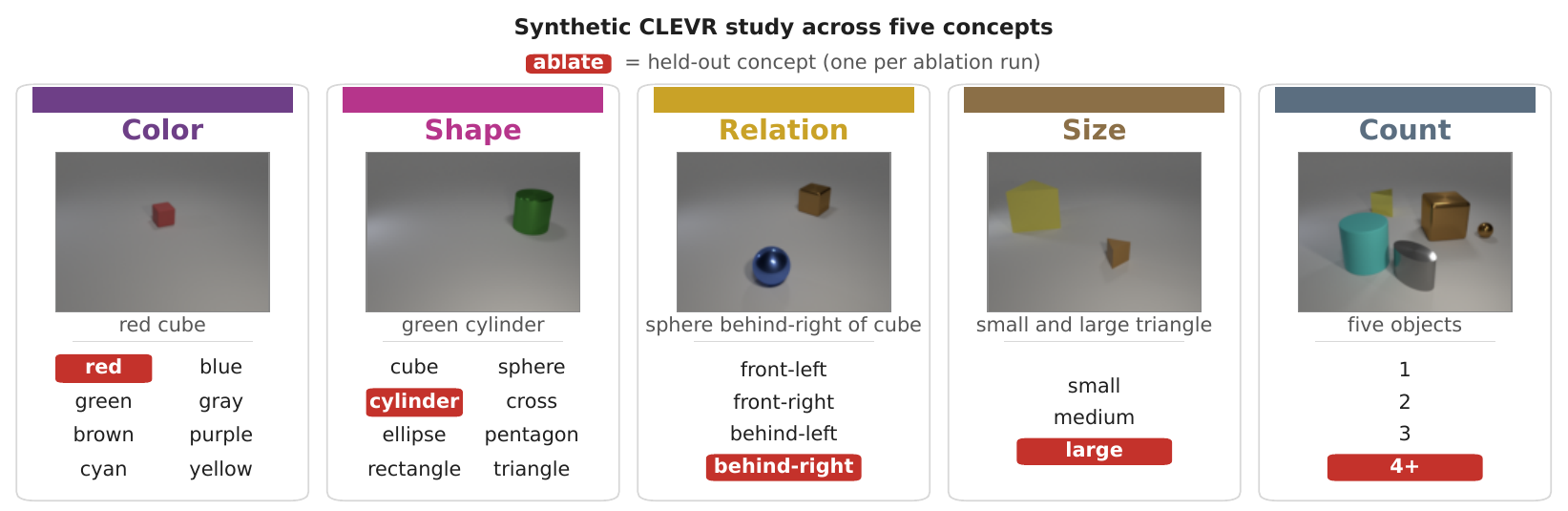}
    \caption{\textbf{Overview of the synthetic CLEVR testbed.} We extend the standard CLEVR vocabulary across five conceptual axes: color, shape, spatial relation, size, and object count. To isolate knowledge flow, specific target concepts (highlighted in red) are systematically ablated from targeted modality training streams.}
    \label{fig:clevr}
\end{figure*}

\subsubsection{Data pipeline and experiment setting}
\label{subsubsec:clevr_setting}
To rigorously test how knowledge transfers between modalities without the noise and confounders of real-world data, we create a strictly controlled synthetic setup. Our high-level questions are twofold: First, if a model learns a specific visual concept entirely through one objective (e.g., visual understanding), can it zero-shot transfer that concept to another objective (e.g., visual generation)? Second, if zero-shot transfer fails, does the initial exposure at least leave a usable latent prior that accelerates learning when the concept is explicitly introduced?

To answer this, we systematically ablate (remove) specific concepts from one modality's training stream, retain them in the other, and observe if cross-modality transfer occurs.

\paragraph{Data.} We build a controlled testbed using the procedurally generated CLEVR dataset~\citep{johnson2017clevr} and extend its shape library. For every rendered scene, we generate three parallel modality streams: a text-to-image pair generation target, an image-to-text descriptive caption, and a set of visual question answering (VQA) pairs. Because CLEVR is rendered from explicit scene graphs, we have full control over the visual concepts present in each example, including per-object color, shape, size, count, and pairwise spatial relations. Samples in this dataset are presented in Figure~\ref{fig:clevr}.

This level of control allows us to strictly filter out any example mentioning a specific target concept (e.g., removing all scenes with a yellow object, or scenes containing 4+ objects) from one modality stream, while leaving the other streams untouched. Any cross-modality performance difference on the ablated concept can therefore only be attributed to knowledge transfer, rather than dataset size or data leakage.

For each held-out concept, a regex word-boundary filter (cross-checked
against the renderer metadata for color, shape, count, and size) drops
every caption, question, or answer mentioning that concept from the
targeted stream(s). All other shards are kept identical, so any
cross-run difference can only be attributed to the content of
the ablated stream rather than to dataset size.

\paragraph{Setting.}
The study is intentionally minimal: in each experiment run, we ablate a specific concept from exactly one of the two main visual objectives. We test two symmetric directions:
\begin{itemize}
    \item \textbf{Understanding $\rightarrow$ Generation:} The concept is removed from the generation stream but kept in the understanding streams (caption and VQA).
    \item \textbf{Generation $\rightarrow$ Understanding:} The concept is removed from the understanding streams but kept in the generation stream.
\end{itemize}

We run this protocol over five distinct concept categories: color (e.g., yellow, red), shape (e.g., sphere, cylinder), spatial relation (e.g., front-left, behind-right), size (e.g., large), and count (e.g., 4+ objects), where the former two represent lower-level attributes and the latter three capture higher-level relational or abstract concepts. 
Figure~\ref{fig:clevr} presents the full suite of concepts. Each directional ablation is compared to a \textbf{baseline} (where the concept is seen in both modalities) and a \textbf{control} (where the concept is dropped from both streams, serving as the zero-exposure floor). All runs share an identical data and compute budgets.

\paragraph{Evaluation.} We evaluate the model's grasp of the held-out concept on both modalities. For understanding, we evaluate VQA accuracy on a held-out test set specifically targeting the ablated concepts. We measure whether the model can correctly answer questions about a concept it has never explicitly seen in its understanding training data. For generation, we prompt the model with a held-out prompt set explicitly requiring the target concept. We render the resulting images and score them using a strong vision-language model (\textbf{Qwen3-VL-8B-Instruct}~\citep{bai2025qwen3}) as an automatic judge to verify whether the generated scene accurately reflects the requested attribute (color, shape, relation, size, or count). Evaluation is conducted exclusively on the held-out concepts, using 100 generated prompts (for generation evaluation) and VQA questions per concept to ensure reliable measurement.  

\begin{figure*}[!h]
    \centering
    \includegraphics[width=\linewidth]{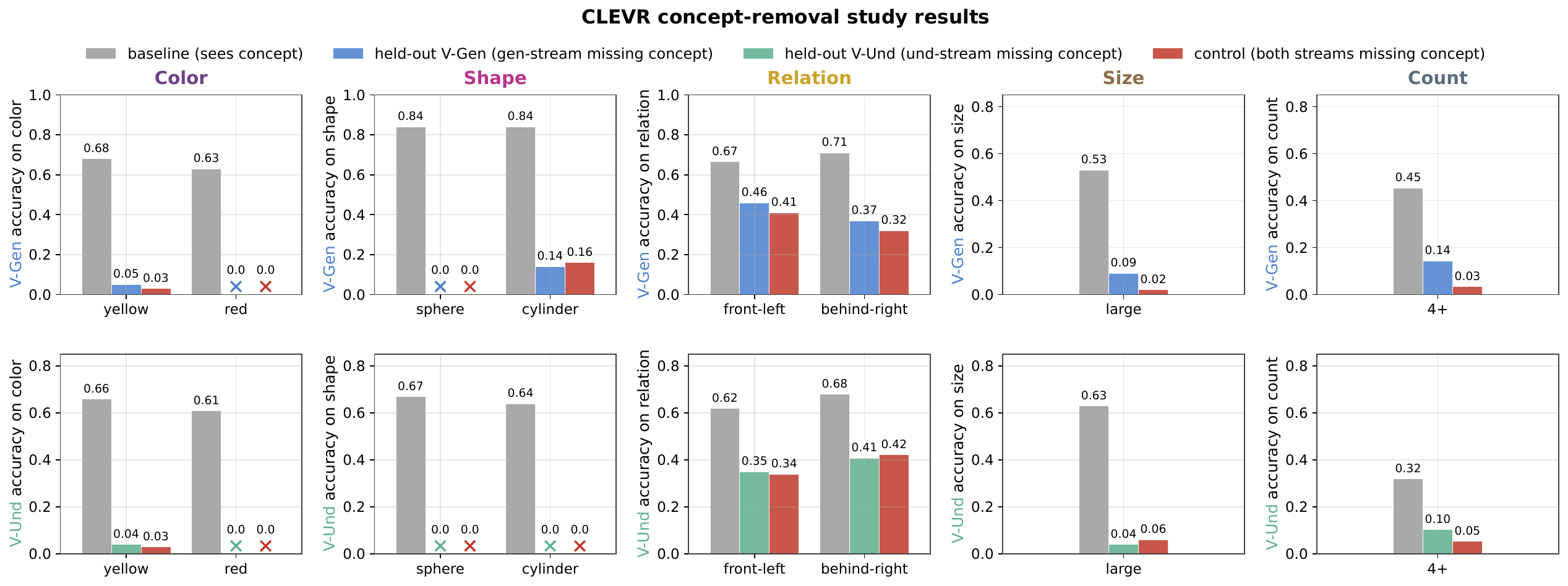}
    \caption{\textbf{Zero-shot concept transfer results on CLEVR.} \textbf{Left (Color, Shape):} Low-level (Color, Shape) attributes fail to transfer in either direction. \textbf{Right (Relation, Size, Count):} Structural concepts exhibit an asymmetric transfer. Understanding helps zero-shot generation, whereas generation largely fails to help understanding, with a minor exception for counting.}
    \label{fig:clevr_ablation}
\end{figure*}

\subsubsection{Zero-shot concept transfer results}
\label{subsubsec:clevr_zeroshot}
We present the concept-removal study results in Figure~\ref{fig:clevr_ablation}. The experiments reveal a nuanced, concept-dependent transfer behavior between visual understanding and generation:

\paragraph{Semantic attributes (Color, Shape) strictly do not transfer.} For more low-level concepts like color and shape, we observe a complete failure of zero-shot transfer in both directions. When a specific color or shape is removed from the generation stream (but kept in understanding), generation accuracy collapses to the zero-exposure control level (blue bars vs. red bars). Symmetrically, removing them from the understanding stream completely destroys VQA performance on those concepts (green bars vs. red bars). This indicates that fundamental visual vocabularies must be explicitly learned within each specific task objective. Such zero-shot transfer failure provides direct support for recent decoupled architectural designs~\citep{janus,li2025manzanosimplescalableunified}, which argue that the visual representations required for generation (dense, pixel-level) and understanding (sparse, semantic) are inherently misaligned and difficult to share directly.

\paragraph{Understanding transfers to generation for structural concepts.} For higher-level, structural concepts, such as spatial relations, size, and object counts, a clear asymmetric transfer emerges. Retaining these concepts exclusively in the understanding stream allows the model to perform better in zero-shot generating them. As seen in the top row of Figure~\ref{fig:clevr_ablation}, generation accuracy for relation, size, and count (blue bars) remains higher than the zero-exposure control (red bars), indicating that spatial and compositional knowledge learned via understanding can guide the generative process.

\paragraph{Generation provides minimal transfer to understanding.} Conversely, learning structural concepts solely through generation does not broadly equip the model to understand them. When relation and size are removed from the understanding stream, VQA accuracy drops to a level almost identical to the control. The only exception is count, where generation provides a slight but observable positive transfer to understanding (green bar $>$ red bar). Overall, this corroborates our real-world findings: understanding acts as a prior for generation, but generative modeling yields no clear zero-shot backward transfer. This strict boundary prompts us to investigate whether any latent knowledge is retained beneath the surface.

\subsubsection{Priors in concept transfer}

\paragraph{Settings.} 
For low-level concepts (such as color and shape) that completely failed to transfer zero-shot in \S\ref{subsubsec:clevr_zeroshot}, we investigate whether any latent knowledge is retained beneath the surface. Although these concepts cannot be utilized immediately without explicit task-specific exposure, training on one stream might establish supportive representations that can be unlocked via fine-tuning. 

We initialize training from the ablated checkpoints and fine-tune them for $2000$ steps on a mixed data stream that contains the previously held-out concept. To monitor the recovery trajectory and measure the acceleration provided by these latent priors, we evaluate the models every $200$ steps, recording both VQA and generation accuracy following the evaluation protocols detailed in \S\ref{subsubsec:clevr_setting}. These learning curves are compared against a control model that has never encountered the target concept in either the understanding or generation stream during pretraining, allowing us to determine if prior experience in the counterpart modality accelerates subsequent concept acquisition.

\begin{figure*}[!h]
    \centering
    \includegraphics[width=\linewidth]{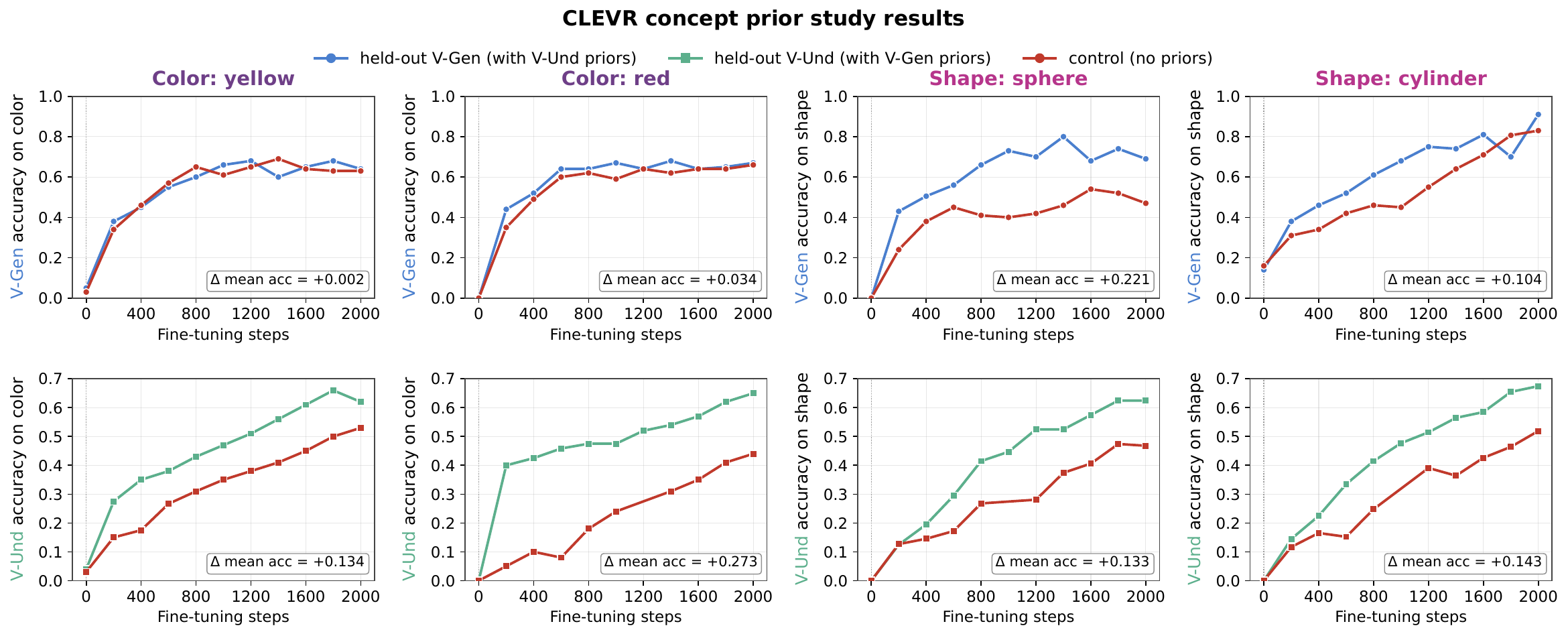}
    \caption{\textbf{Concept recovery via fine-tuning.} We measure how quickly models learn a missing low-level concept. \textbf{Top row:} Prior exposure via visual understanding provides no acceleration for color generation, but leaves a usable prior that accelerates shape generation. \textbf{Bottom row:} Prior exposure via visual generation acts as a booster, accelerating visual understanding learning across both color and shape.}
    \label{fig:prior}
\end{figure*}

\paragraph{Results.}
As depicted in Figure~\ref{fig:prior}, while low-level concepts fail at zero-shot transfer, examining their fine-tuning trajectories reveals a stark and unexpected asymmetry in latent priors:

\textbf{Generation provides a strong latent prior for low-level concept understanding.} As shown in the bottom row of Figure~\ref{fig:prior}, models that previously encountered a specific color or shape strictly through generative training (green curves) recover understanding capabilities significantly faster than the control group (red curves). Across all tested low-level concepts, the prior group achieves a notably higher area under the curve ($\Delta$ mean acc ranging from +0.133 to +0.273). This indicates that the generative objective forces the model to learn robust, fine-grained visual representations (e.g., exact pixel distributions of a color or texture of a shape) that, while not immediately accessible for zero-shot VQA, serve as a highly reusable foundation once the model is shown how to map them to language.

\textbf{Understanding provides minimal priors for low-level concept generation.} Conversely, the top row demonstrates that prior exposure through visual understanding offers virtually no advantage for visual generation. For colors (yellow, red), the recovery curves of the prior group (blue) and the no-prior control group (red) overlap almost perfectly ($\Delta$ mean acc $\approx$ 0). For shapes, there is only a marginal improvement. This suggests that the discriminative features learned via understanding are often too abstract or sparse to guide the dense, pixel-level reconstruction required for generation. 


\textbf{Why generative synergy is task-selective.} 
Our results help resolve a common contradiction in multimodal learning: why joint generative pretraining often provides little benefit (or even an alignment penalty) on standard understanding tasks, yet sometimes improves understanding performance in spatial, geometric, and physical reasoning related tasks~\citep{hu2025omni,wen2026unig2u,su2026generation}.

Standard benchmarks focus mostly on high-level semantic classification, which only requires sparse and abstract representations. For such tasks, the dense, pixel-level details learned from generation are mostly redundant. In contrast, spatial and structural tasks demand a precise understanding of depth, boundaries, and perspective. The generative objective forces the model to learn these dense visual features, building a richer geometric representation. As shown in our results on low-level property priors, these generation-learned priors can be effectively utilized within understanding streams. Consequently, while this geometric foundation may remain dormant in zero-shot settings, it serves as a highly receptive prior that can be rapidly activated to advance understanding tasks that require dense features.

\section{Synergy vs. Competition in Unified Pretraining}
\label{sec:synergy}

A unified model must allocate a single set of parameters across visual generation, visual understanding, and language modeling. Whether these objectives reinforce one another (\textbf{synergy}) or compete for capacity (\textbf{competition}) determines whether unification is a net positive or merely a structural convenience. We dissect this question along three complementary axes: the data and task complexity of each modality (\S 4.1), parameters sharing between modalities (\S 4.2), and generalization across multiple vision encoder designs (\S~\ref{subsec:encoder}). The first asks whether harder visual or linguistic data taxes the other modality more; the second asks whether weight-sharing strategies can convert latent competition into synergy; the third evaluates whether these interaction patterns remain consistent under different visual representation spaces.

\finding{3}{Modality interaction is governed by task complexity and parameter sharing, generalizing across vision tokenization designs. Simple tasks act as cross-modal boosters, whereas complex tasks induce capacity competition that outweighs synergy. Architecturally, sharing attention and normalization layers fosters cross-modal synergy, while decoupling feed-forward networks mitigates capacity competition.}

\subsection{Data and task complexity}
\label{subsec:complexity}

Whether two modalities cooperate or compete during unified pretraining is not solely governed by architecture; it is deeply tied to the intrinsic complexity of the data they are asked to learn. We hypothesize that a noisy, low-information visual stream imposes fundamentally different demands on a shared parameter space than a highly structured natural-language stream. Consequently, the magnitude of cross-modal interference, or the potential for synergy, predictably scales with the semantic richness and difficulty of the tasks. To systematically probe this, we isolate the task complexity of one modality while holding the other completely fixed.

\begin{figure*}[!h]
    \centering
    \includegraphics[width=\linewidth]{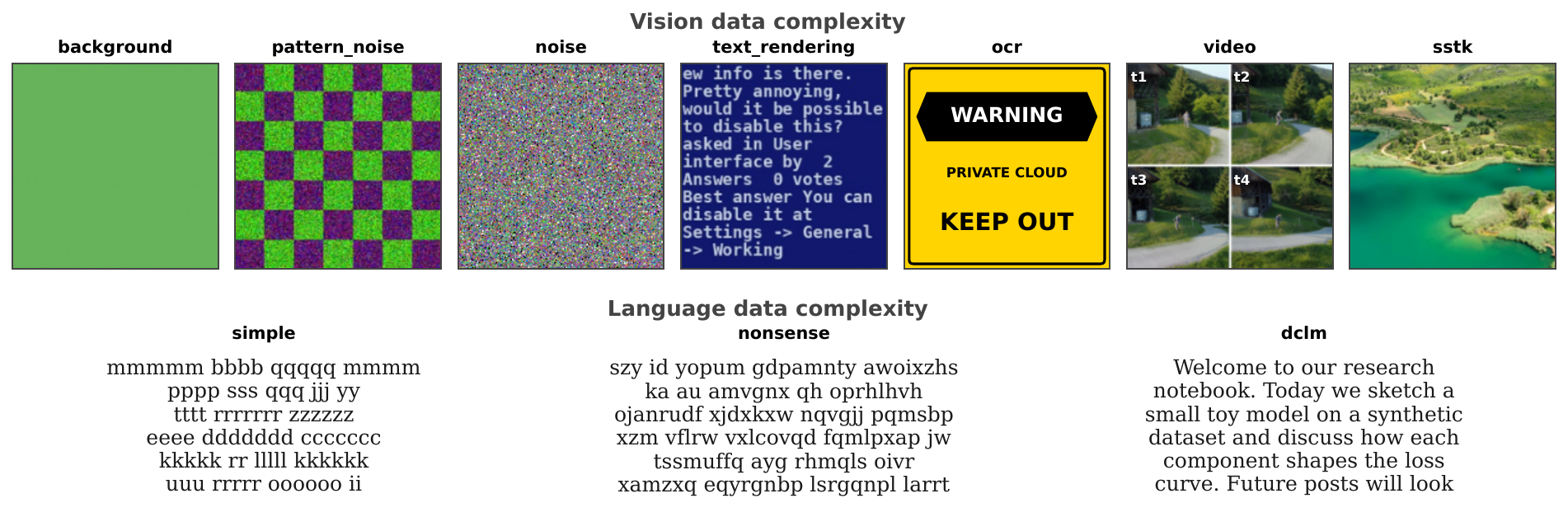}
    \caption{\textbf{Overview of data complexity progressions.} Examples of visual (top) and language (bottom) data used to evaluate the impact of task complexity on modality interactions, ranging from simple synthetic patterns to complex real-world distributions.}
    \label{fig:part12_data_samples}
\end{figure*}

\paragraph{Setting.}
Models are trained from scratch for a fixed compute budget of $100$B tokens, evenly split ($50$B/$50$B) between language and primarily unconditional image generation. To establish baselines, we also train unimodal references on $50$B tokens of their respective modalities. We further verify this baseline choice by testing a $100$B token two-modality run where the counterpart modality is fully processed but its loss weight is set to zero, which yielded highly consistent results. We design two symmetric complexity progressions (illustrated in Figure~\ref{fig:part12_data_samples}). Note that while the escalation in data complexity across these stages represents a broad, general trend, it is not an absolute or strictly monotonic measure.

\begin{itemize}
    \item \textbf{Vision-task complexity data:} Holding the language stream constant at 50\% DCLM, we progressively escalate the difficulty of the visual targets along seven rungs of structural complexity:
    \begin{itemize}
        \item \textbf{(i) Pure backgrounds (\texttt{background})}, synthetic images filled with a single solid RGB color.
        \item \textbf{(ii) Patterns (\texttt{pattern\_noise})}, synthetic images with gradients, checkerboards, and stripes.
        \item \textbf{(iii) Noise (\texttt{noise})}, images of unstructured stochastic noise (Gaussian, salt-and-pepper, speckle) at different intensities.
        \item \textbf{(iv) Text-rendering (\texttt{text\_rendering})}, canvases on which DCLM text snippets are rendered with random fonts, sizes, and color palettes.
        \item \textbf{(v) OCR images (\texttt{ocr})}, 
        OCR corpus of real photographs containing scene text (signage, posters, logos).
        \item \textbf{(vi) Video (\texttt{video})}, 
        instruction-style video--text clip corpus, sampled at 8 frames per clip.
        \item \textbf{(vii) SSTK (\texttt{sstk})}, a corpus of real natural images.
    \end{itemize}
    We measure the cross-modal impact on language by reporting text perplexity.

    \item \textbf{Language-task complexity data:} Holding the visual stream constant (SSTK natural images), we systematically upgrade the language stream along three rungs:
    \begin{itemize}
        \item \textbf{(i) Simple language (\texttt{simple})}, synthetic text in which every word is a single lowercase letter repeated a handful of times.
        \item \textbf{(ii) Nonsense (\texttt{nonsense})}, synthetic text (alphabet a to z) in which every word is a random consonant-biased letter string.
        \item \textbf{(iii) DCLM (\texttt{dclm})}, the full DCLM web-text mixture.
    \end{itemize}
    We measure the cross-modal impact on vision via held-out validation diffusion loss.
\end{itemize}

Evaluating cross-modal effects requires the target distribution of the measured modality to remain static. Diffusion loss is directly comparable across the language variations because the visual targets (SSTK) never change. Similarly, text perplexity is strictly comparable across the vision variations because the text targets (DCLM) are identical. We therefore rely exclusively on vision $\rightarrow$ language ppl and language $\rightarrow$ vision diffusion loss to quantify interference and synergy.

\begin{figure*}[!h]
    \centering
    \includegraphics[width=\linewidth]{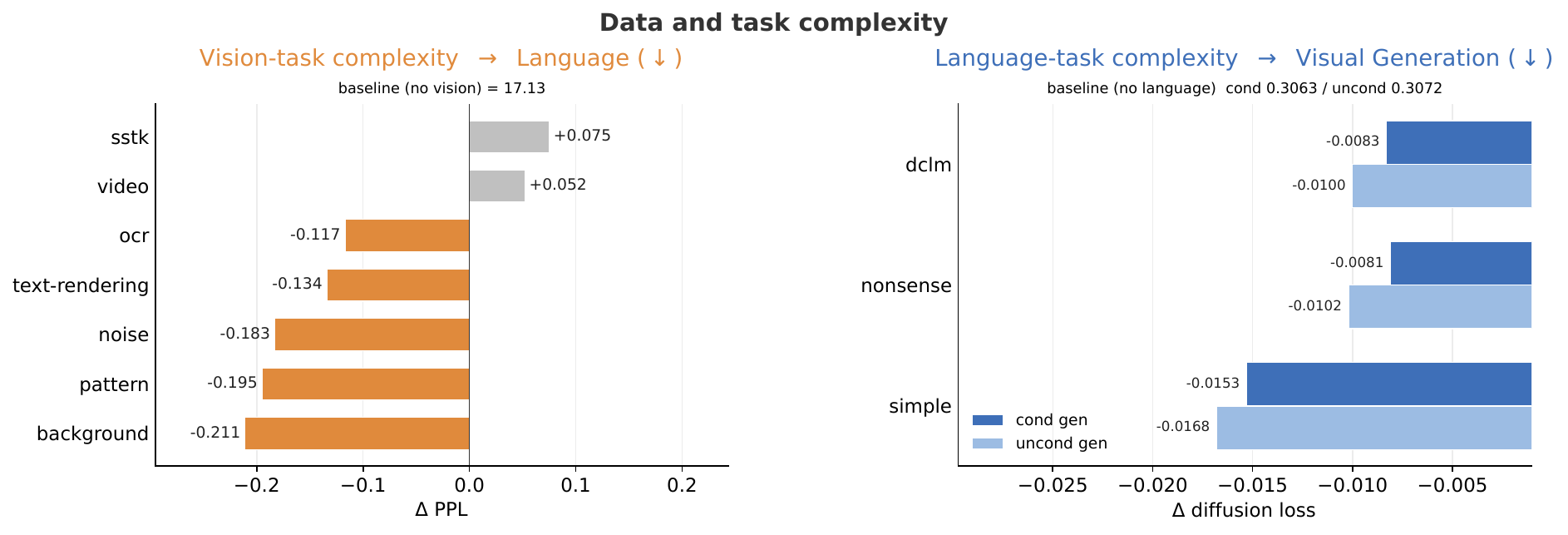}
    \caption{\textbf{Impact of task complexity on modality interaction.} \textbf{Left:} Escalating visual task complexity gradually turns synergy into competition. Simple visual tasks (e.g., backgrounds, noise) improve language modeling, whereas complex visual distributions (SSTK, video) degrade text perplexity. \textbf{Right:} Introducing language universally aids visual generation, but the simplest linguistic distribution provides the maximum synergistic boost.}
    \label{fig:part12_data}
\end{figure*}

\paragraph{Results.} As shown in Figure~\ref{fig:part12_data}, varying the task complexity reveals a stark dynamic between parameter competition and cross-modal synergy:

\textbf{Simple tasks act as cross-modal boosters.} Counterintuitively, asking a model to learn a low-complexity task in one modality actually improves its performance on the other. On the vision side (Figure~\ref{fig:part12_data}, left), pairing language with extremely simple visual targets (such as backgrounds with $\Delta$ ppl $-0.211$ or noise with $\Delta$ ppl $-0.183$) results in significantly better language perplexity than training a pure language model without vision at all. Symmetrically, on the language side (Figure~\ref{fig:part12_data}, right), introducing a highly simplified distribution yields the largest improvement in visual generation loss ($\Delta$ $-0.0153$ for conditional and $-0.0168$ for unconditional generation), outperforming both the no-language baseline and the more complex DCLM text mixture.

\textbf{Complex tasks induce capacity competition.} As the semantic richness of the data scales up, the observable synergistic effect diminishes. When the visual stream is upgraded to complex real-world data like video and SSTK natural images, text perplexity noticeably degrades compared to the unimodal baseline ($\Delta$ PPL $+0.052$ and $+0.075$, respectively). This demonstrates that while an underlying cross-modal synergy inherently exists, forcing both modalities to model high-entropy, complex real-world distributions causes them to fiercely compete for the network's finite parameter capacity. As task difficulty increases, this capacity competition eventually outweighs the synergistic benefits.

These results reveal that unified pretraining is defined by a delicate interplay of two opposing forces: cross-modal synergy and capacity competition. Since both forces coexist within the same training loop, we next investigate where these synergies and competition manifest within the model's internal components.

\subsection{Parameter-sharing in Transformers for promoting synergy}
\label{subsec:archsynergy}

To understand how the architecture of the network dictates these modality interactions, we dissect the standard Transformer block. Specifically, we evaluate which components require specialization to mitigate capacity competition, and which must remain shared to foster cross-modal synergy.

\paragraph{Setting.} We freeze the data mixture at a 50\%/50\% split (DCLM and SSTK) and deconstruct the standard Transformer block into three configurable parameter groups. Each group either remains shared across modalities or splits into modality-specific copies. These groups are: the feed-forward network (\textbf{FFN}), the attention mechanism (\textbf{Attention}, which includes Q/K/V/O projections and cross-modal routing), and the final layer normalization (\textbf{FinalNorm}). All other base components like embeddings remain uniformly shared.

We study five settings to test the effect of parameter sharing:

\begin{itemize}
    \item \textbf{dense:} All three groups are shared. Visual and language tokens process through identical weights and fully attend to each other, serving as the fully coupled baseline.
    \item \textbf{split\_ffn:} Only FFNs are decoupled per modality. Attention and FinalNorms remain shared to test whether isolating computation capacity while preserving cross-modal attention can mitigate competition.
    \item \textbf{split\_ffn\_attn:} Both FFNs and Attention mechanisms are decoupled (disabling cross-modal attention), leaving only FinalNorm shared. This tests the effect of completely restricting cross-modal token interactions.
    \item \textbf{split\_ffn\_norm:} Both FFNs and FinalNorms are decoupled, while Attention remains shared. This tests whether accommodating different modality signal magnitudes via separated normalization provides additional benefits.
    \item \textbf{split\_all:} All three components are separated. This effectively maintains two independent parallel networks within the block, testing the impact of complete parameter isolation.
\end{itemize}

All architectural variants are trained from scratch on the exact same 100B token budget (50B language / 50B visual). To maximize the observable variance between architectural choices, we specifically employ the simple language (for the language stream) and pure backgrounds (for the vision stream). We quantify the effectiveness of each configuration by measuring the performance improvement or degradation it brings compared to unimodal baselines, which inherently represent a complete absence of parameter sharing.

\begin{figure*}[!h]
    \centering
    \includegraphics[width=\linewidth]{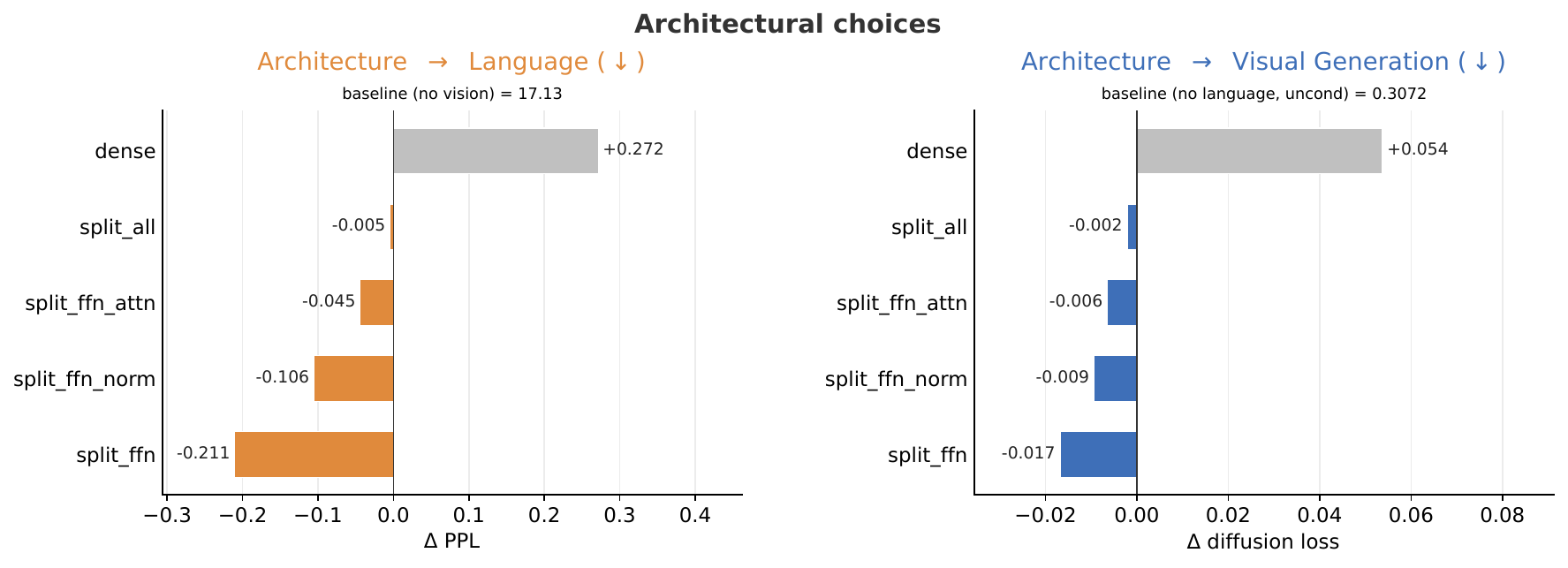}
    \caption{\textbf{Impact of parameter sharing on cross-modal performance.} Fully shared (\texttt{dense}) parameters force modality competition, degrading both language and vision. Decoupling solely the FFNs (\texttt{split\_ffn}) perfectly mitigates this competition while leveraging shared attention to foster strong synergy. Decoupling attention (\texttt{split\_ffn\_attn}) or normalization (\texttt{split\_ffn\_norm}) significantly diminishes these improvements, and fully isolating all parameters (\texttt{split\_all}) yields identical results to baselines.}
    \label{fig:part12_arch}
\end{figure*}

\paragraph{Results.} As depicted in Figure~\ref{fig:part12_arch}, the architectural design fundamentally dictates whether modalities compete or synergize:

\textbf{Dense networks suffer from severe competition.} The standard \texttt{dense} architecture, where language and visual tokens are processed through completely shared weights, exhibits severe modality competition. Compared to their respective unimodal baselines, the dense model suffers a clear degradation, increasing language ppl by 0.272 and visual diffusion loss by 0.0537. This confirms that forcing distinct modalities into the exact same parameter subspace is structurally suboptimal.

\textbf{Decoupling FFNs mitigates competition.} By isolating just the feed-forward networks (\texttt{split\_ffn}), the network mitigates capacity competition while preserving the synergistic benefits. This configuration yields significant cross-modal benefits, improving language ppl by 0.211 and lowering vision loss by 0.0168 beyond the unimodal baselines. This provides support for recent Mixture-of-Transformers (MoT) and Mixture-of-Experts (MoE) designs: modality-specific FFNs provide the necessary isolated capacity to prevent destructive interference~\citep{liang2024mixture,lin2024moma,meituanlongcatteam2026longcatnextlexicalizingmodalitiesdiscrete,tong2026beyond,deng2025bagel}.

\textbf{Sharing attention and norm drives synergy.} We further observe that the cross-modal benefits heavily depend on shared attention and normalization. When we decouple the attention mechanisms (\texttt{split\_attn}), the performance improvements drastically shrink, with the language ppl improvement dropping to 0.045 and the vision loss reduction to 0.0064. Separating the normalization layers (\texttt{split\_norm}) also visibly diminishes the performance improvements. Finally, completely isolating all components (\texttt{split\_all}) results in near-identical results as baselines. At this point, the network performs almost identically to the unimodal baseline.

We hypothesize that shared attention and normalization act as a bridge for synergy: shared attention allows modalities to contextualize each other within a joint representational space, and shared normalization aligns their feature scales. Conversely, FFNs act as modality-specific experts that absorb the distinct computational demands of each stream, preventing destructive interference in the shared substrate.

\subsection{Generalization across multiple vision encoder designs}
\label{subsec:encoder}

To further understand whether the observed modality synergies are tied to a specific visual representation, we evaluate how different vision encoder and tokenization designs affect these interactions. 

\textbf{Setting.} Following the same \texttt{split\_ffn} training protocol as in \S\ref{subsec:archsynergy}, we pretrain models on a 100B token budget (50B language / 50B visual) using our synergy-maximizing probes: pure backgrounds for the vision stream (measuring the change in language perplexity, $\Delta$ PPL) and simple language for the text stream (measuring the relative change in visual generation loss, relative $\Delta$ diffusion loss \%). We compare this setup across the four distinct visual tokenization configurations (or vision encoder/decoder designs) introduced in \S\ref{section:formulation}: the default \textbf{RAE}, \textbf{Raw Pixels}, decoupled \textbf{CLIP + VAE}, and \textbf{AR (UniTok)}.

\begin{figure*}[!h]
    \centering
    \includegraphics[width=\linewidth]{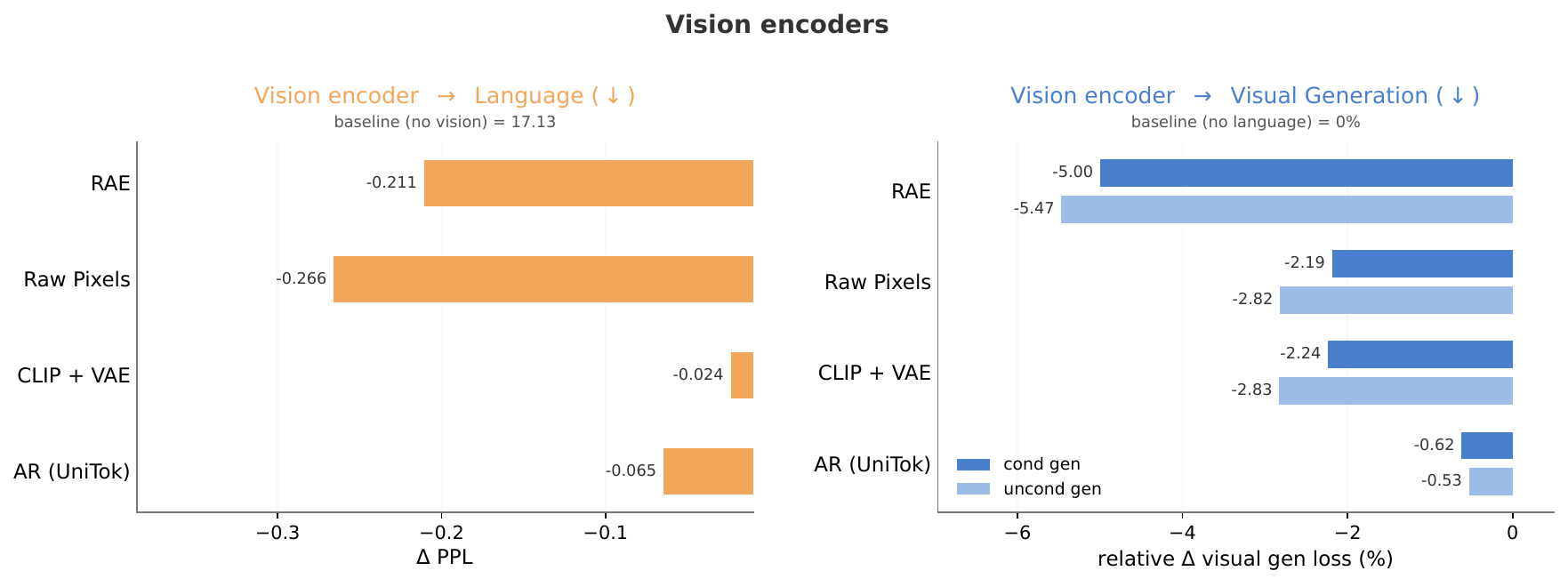}
    \caption{\textbf{Impact of vision encoder designs on modality synergy.} Left: The impact of pairing pure background images with language across different encoder configurations on language perplexity ($\Delta$ PPL). Right: The relative change in diffusion loss (\%) for conditional and unconditional generation when paired with simple language. Modality synergy consistently occurs across all four visual tokenization designs.}
    \label{fig:part12_arch_encoders}
\end{figure*}

\textbf{Results.} As depicted in Figure~\ref{fig:part12_arch_encoders}, cross-modal synergy occurs across all four vision encoder designs. Even when the model is trained on \textbf{Raw Pixels} without any pre-existing semantic prior, pairing it with pure background images yields the largest language perplexity improvement ($\Delta$ PPL of $-0.266$), outperforming the default RAE setup ($-0.211$). Similarly, on the generation side, \textbf{Raw Pixels} achieves notable relative diffusion loss reductions of $-2.19\%$ (conditional) and $-2.82\%$ (unconditional) when trained with simple language. Under the decoupled \textbf{CLIP + VAE} setting, where understanding and generation are structurally segregated into distinct latent spaces, the model still exhibits cross-modal synergy. Although the language perplexity benefit is modest ($\Delta$ PPL of $-0.024$), the visual generation capability receives a substantial boost (relative diffusion loss reductions of $-2.24\%$ and $-2.83\%$). The autoregressive result also exhibits consistent synergy, confirming that cross-modal synergy persists when transitioning to discrete sequence modeling.

These findings strongly align with our real-world modality transfer results detailed in \S\ref{section:flowstudy} and Appendix~\ref{appendix:modalitytransfer} (Figures~\ref{fig:sup_part1_raw}, \ref{fig:sup_part1_vae}, and \ref{fig:sup_part1_ar} real-world modality transfer results on different vision encoders), which show that asymmetric knowledge flow remains highly consistent regardless of visual tokenization choice. A critical question in unified multimodal models is whether the vision representation must reside within a pre-aligned, shared representation space to boost cross-modal synergy. Taken together, our results demonstrate that a pre-aligned (\textbf{Raw Pixels}) or shared vision representation space (\textbf{RAE}) is not an absolute prerequisite for modality transfer or cross-modal synergy, which is also driven by task objectives and parameter-sharing designs in Transformer blocks. Nonetheless, a unified representation space remains beneficial, as it minimizes representational friction and design complexity and thereby boosts the overall efficiency and magnitude of the resulting synergy.

\section{The Necessity of Early Unification}
\label{sec:earlyunification}

\finding{4}{Early and simultaneous unification is highly beneficial. Multimodal integration should ideally occur in the early stages of pretraining, and modalities are better trained jointly. Delaying visual integration consistently degrades native visual capabilities, while isolating training into sequential stages is prone to catastrophic forgetting and does not fully unlock cross-modal synergies.}

Given that vision heavily benefits from language priors, a natural intuition might be to pretrain a pure language model first and align visual capabilities later. In this section, we challenge this common practice. By systematically varying the timing of visual integration (\S~\ref{subsec:early}) and the sequencing (\S~\ref{subsec:sequential}) of modality training, we investigate the underlying temporal mechanics of multimodal learning. Our findings challenge the prevailing late-alignment heuristics~\citep{liu2023visual,bai2025qwen3}, revealing why early and simultaneous unification~\citep{team2024chameleon} is a necessity to prevent vision laziness and unlock more cross-modal synergy (\S~\ref{subsec:visionlaziness}).

\subsection{Early unification vs. late unification}
\label{subsec:early}
\paragraph{Setting.} Holding the total compute budget fixed at 1T tokens, we vary \emph{when} unified training begins by sweeping the number of pure-language tokens consumed beforehand over ${0, 200, 400, 600, 800}$B; the remaining ${1000, 800, 600, 400, 200}$B are then spent in a unified stage with a 50\%/50\% language/vision mix (vision split evenly between captioning and conditional generation). The $0$B point is the unified-from-scratch baseline, while larger values correspond to first training a pure-language model and then continuing with the unified objective from that checkpoint. 

\begin{figure*}[!h]
    \centering
    \includegraphics[width=\linewidth]{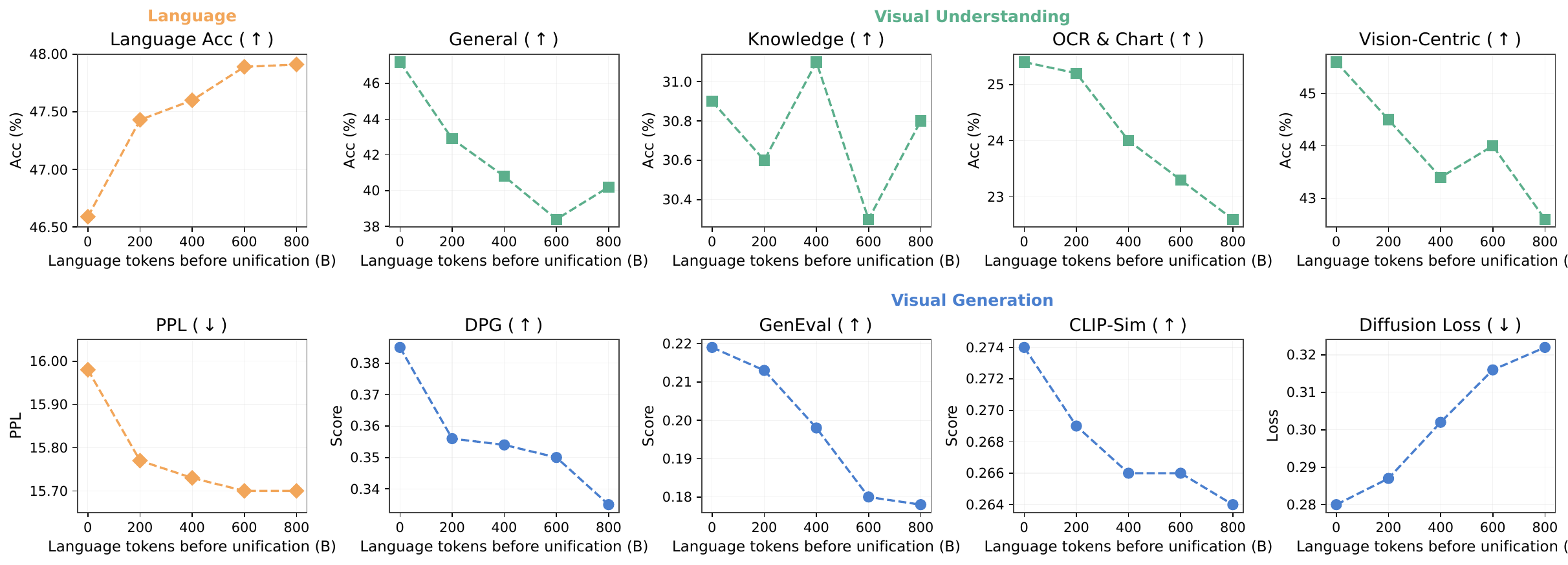}
    \caption{\textbf{Timing of unification training.} The x-axis represents the number of pure language tokens consumed before visual data is introduced to the training mix. While extending the initial pure language phase yields marginal improvements in unimodal text metrics like language accuracy and perplexity, it triggers a steep and consistent decline in performance across all visual understanding and visual generation benchmarks. }
    \label{fig:part6}
\end{figure*}

\paragraph{Results.} As illustrated in Figure~\ref{fig:part6}, the timing of when visual is introduced fundamentally alters the model's final capabilities. On the language side, extending the initial pure-language training phase yields early improvements in both downstream accuracy and perplexity. However, these benefits shrink significantly over time, with performance essentially plateauing between 600B and 800B pretraining tokens.

For the vast majority of visual capabilities, the trend is clear: the earlier visual data is introduced, the better the performance. General, OCR \& Chart, Vision-Centric VQA tasks, and all visual generation metrics show a steep and consistent decline as the pure-language phase is prolonged. Overall, results show a window for multi-modal unification. While a minimal initial pure-language phase can provide a foundational text prior without severely harming vision, pushing unification later into training offers less benefit to language while systematically crippling the model's visual understanding and generative fidelity.

\subsection{Sequential vs. joint training}
\label{subsec:sequential}

\begin{figure*}[!h]
    \centering
    \includegraphics[width=\linewidth]{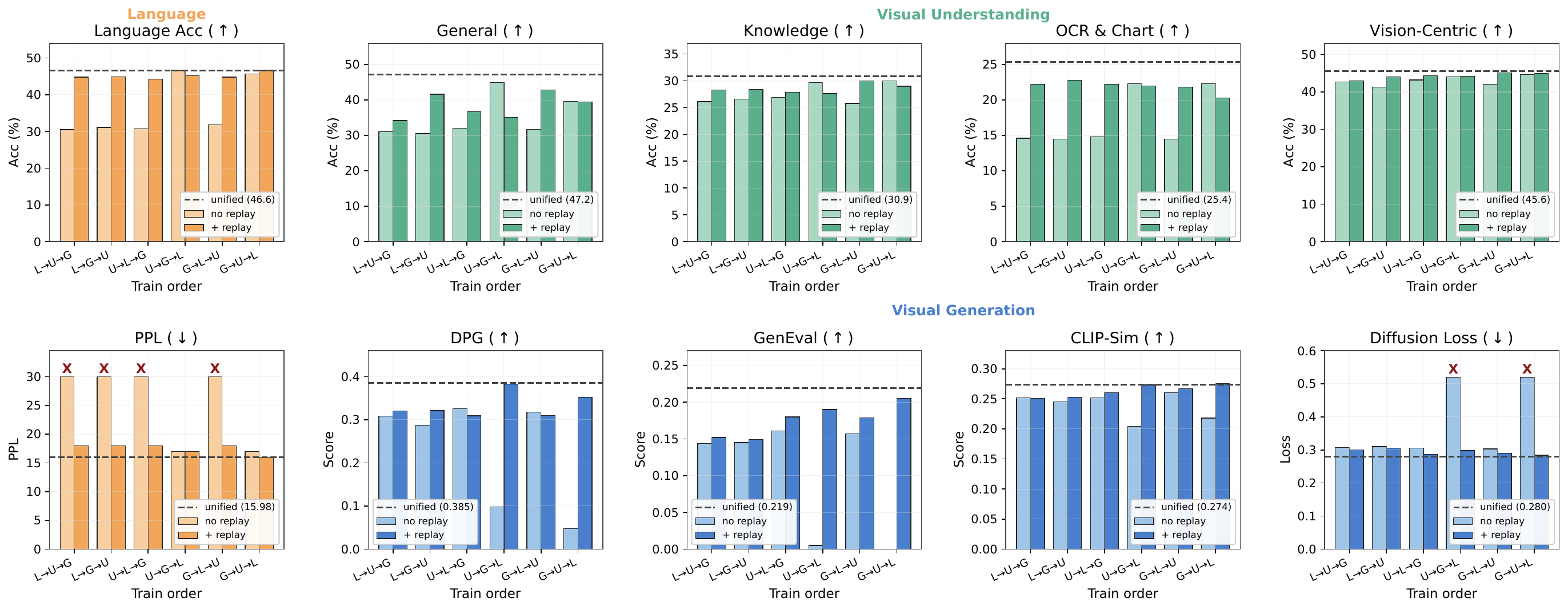}
    \caption{\textbf{Impact of sequential versus joint pretraining across various modality orderings.} The charts display the performance of six distinct sequential training paths. Solid bars denote strict sequential training, while patterned bars indicate training with a 12.5\% replay buffer of previously seen modalities. The horizontal dashed line represents the simultaneous joint training baseline. The results clearly show that joint training dominates all sequential approaches across almost every metric. Although data replay slightly mitigates catastrophic forgetting, it fails to match the cross-modal synergies.}
    \label{fig:part5}
\end{figure*}

\paragraph{Settings.} Late-fusion style training often implicitly assumes a language-first progression, with vision introduced only afterward. More broadly, this raises a natural question: since humans typically acquire vision before language~\citep{steinberg1975reading,smith2005development,orhan2024learning,vong2024grounded,bambach2018toddler}, would a vision-first training order be more effective than a language-first one? Beyond studying unified timing, we therefore examine whether multimodal learning can be decomposed into discrete sequential stages. Holding our 1T token budget and a standard 50/25/25 modality mix constant for Language, Understanding, and Generation, we enforce three separate training stages and test all six possible modality orderings. We run each sequence both with and without a 12.5\% data replay mechanism designed to mitigate forgetting. These runs are then compared against a fully joint baseline, shown as a horizontal dashed line in Figure~\ref{fig:part5}.

\paragraph{Results.} The empirical results show a decisive advantage for simultaneous joint training. Regardless of the sequence order, isolated training fails to match the performance of joint training on visual tasks. A human-like vision-first progression (e.g., G$\rightarrow$ U $\rightarrow$ L) does not outperform the unified baseline in all dimensions.

Furthermore, while retaining a 12.5\% replay buffer of previously seen modalities (to mitigate catastrophic forgetting~\citep{wu2025mitigating,zhai2023investigating}) slightly elevates performance compared to strict sequential training, it still falls drastically short of the joint baseline. This suggests that modalities do not merely coexist: they actively co-evolve, and treating them as separate curriculum steps reduces the mechanics of modality synergy.

\subsection{Vision laziness in late alignment}
\label{subsec:visionlaziness}

\finding{5}{Late alignment introduces vision laziness: the longer the language warm-up, the less the vision pathway commits itself. We show this through four independent mechanistic measurements.}

The previous studies established that delaying multimodal unification systematically degrades visual performance. Here we ask \emph{why}: what changes inside the model when the language trunk is allowed to harden before vision is introduced? We probe this question along four orthogonal axes, all on the same controlled set of checkpoints, and find a single coherent mechanism: the vision pathway becomes increasingly underdeveloped and increasingly disconnected from the language manifold the more pretrained the language trunk is at the moment of unification.

\paragraph{Setting.} We hold the data mix, training recipe, and
downstream training budget constant across five runs and vary only the
language warm-up: each model is initialised from a pure-language
checkpoint at $\{0, 200, 400, 600, 800\}$~B language tokens and then
fine-tuned for an additional 200~B-token continuation on a 50\%/50\%
language/vision mix that mirrors a typical late-alignment regime. To
test whether vision laziness is task-specific, we run two parallel
experiments: a generation sweep
(vision side trained purely on conditional image generation) and an
understanding sweep (vision side trained purely on image captioning).
We use a split-FFN architecture in which the image-side feed-forward
branch (\texttt{img\_ffn}) is randomly initialised in every
run regardless of starting point. Cross-checkpoint differences in its
weights or activations are therefore a clean, scale-fair measure of
how much vision the model committed to learning, free of the confound
that the language pathway inherits different amounts of pretraining.

We probe vision laziness along four axes:
\begin{itemize}
    \item \textbf{A: Training-time activation L2 of \texttt{img\_ffn}.} For every training step we record the L2 norm of the
    \texttt{img\_ffn} and report
    average across all 100k optimization steps. This captures how
    loud the model's hidden states are while it is actually being
    trained on a vision-conditioned objective.
    \item \textbf{B: Embedding L2 norm of the special image-wrapper
    tokens.} The wrapper tokens (\texttt{<}begin\_of\_img\texttt{>} and
    \texttt{<}end\_of\_img\texttt{>}) are routed through the language FFN, so
    the L2 magnitude of their embedding rows in
    \texttt{tok\_embeddings.weight} reports how much the language
    pathway has reshaped itself to accommodate the image segment.
    Smaller magnitudes indicate that the wrapper has been pushed away
    from typical language-vocab scales.
    \item \textbf{C: Inference-time per-element activation of
    \texttt{img\_ffn}.} We hook the image-side FFN during
    a real vision forward pass — the diffusion image-generation loop
    on the generation side, an image-conditioned caption forward on
    the understanding side, and record the L2 norm of its output
    tensor, normalised by sqaure root of the number of elements to obtain
    a per-element RMS that is comparable across the other axes. This
    is a function-level readout of how much computation the vision
    branch performs when the model is actually using vision.
    \item \textbf{D: Inference-time attention on image tokens.} For
    each layer (head-averaged) we measure how much of the model's
    attention is actually placed on the image tokens during a real
    vision forward pass, which is the fraction of attention that
    image-patch queries direct at other image patches on the
    generation side, and the fraction that text queries direct at the
    image patches on the understanding side. Larger values mean the
    model is genuinely focusing on the image when it
    generates or answers; smaller values mean the image is in favour of the surrounding text context. This
    is a direct readout of where the vision pathway puts its
    attention, complementing the magnitude-based readouts in A and C.
\end{itemize}

\begin{figure*}[!h]
    \centering
    \includegraphics[width=\linewidth]{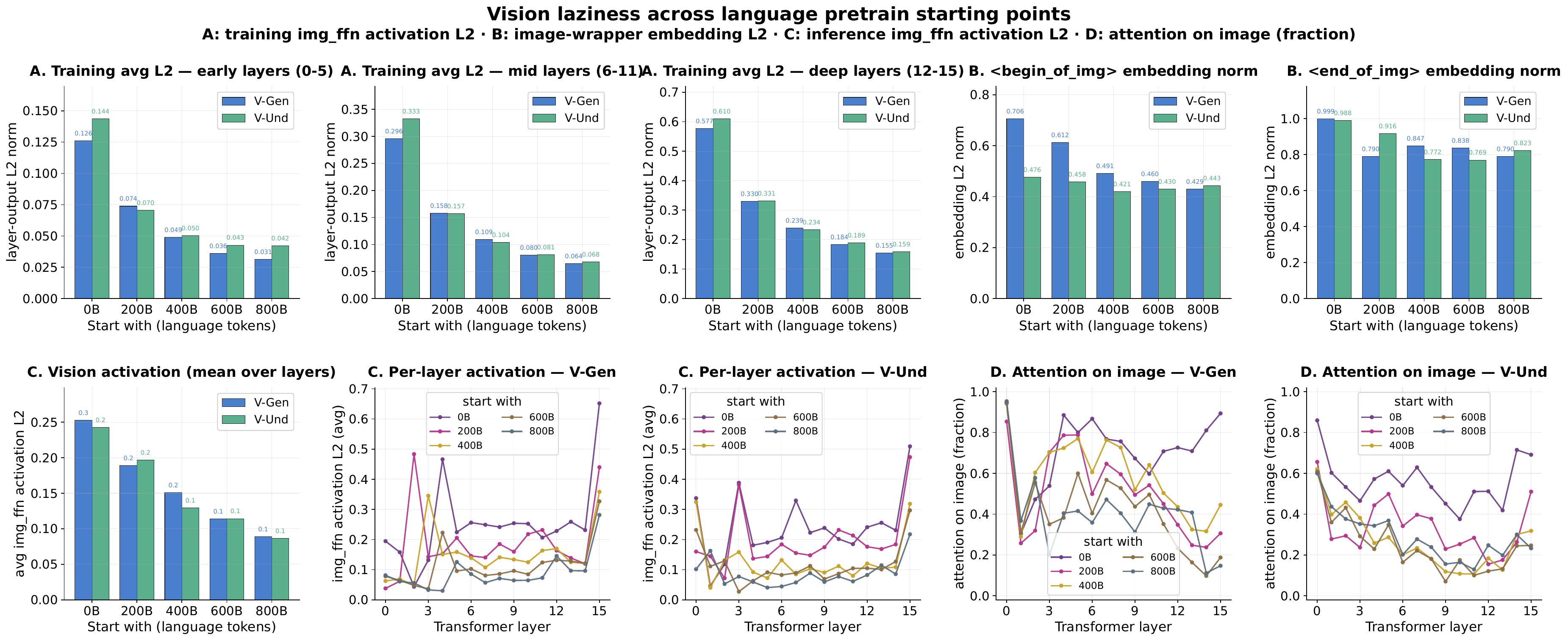}
    \caption{\textbf{Results of vision laziness in late alignment.} We measure the visual pathway's commitment across varying language starting points (0B to 800B tokens) for visual generation (V-Gen) and understanding (V-Und). \textbf{(A)} Training-time activation L2 norms of vision-specific FFNs consistently decrease across layers as the language warm-up lengthens. \textbf{(B)} Embedding norms of special image-wrapper tokens shrink in late-aligned models, showing weaker integration into the language manifold. \textbf{(C)} Inference-time activations of the vision FFNs drop drastically when visual unification is delayed. \textbf{(D)} The fraction of attention placed on image tokens at inference time, from image queries during generation and from text queries during understanding, likewise drops with longer language warm-ups, meaning the model focuses less on the image itself. Across all metrics, longer pure-language pretraining reduces the model's utilization of its visual parameters.}
    \label{fig:laziness}
\end{figure*}

\paragraph{Results.} 
 As shown in Figure~\ref{fig:laziness}, delaying multimodal unification fundamentally alters the internal mechanics of the model, systematically inducing vision laziness. Across all four mechanistic axes, we observe a stark decline in the visual pathway's activity and integration as the pure-language pretraining phase lengthens. First, training-time activations (Figure~\ref{fig:laziness} A) and inference-time per-layer activations (Figure~\ref{fig:laziness} C) reveal that the visual feed-forward networks become progressively "quieter". Models initialized with a heavily pretrained language trunk (e.g., 600B or 800B tokens) exhibit significantly lower layer-output magnitudes compared to the unified-from-scratch baseline (0B), indicating that the vision module contributes less to the representations. Furthermore, the embedding norms of the special wrapper tokens (Figure~\ref{fig:laziness} B) shrink with longer language warm-ups, suggesting that a hardened language pathway resists accommodating the new visual modality, pushing the wrapper tokens away from typical expressive feature scales. Finally, the inference-time attention readout (Figure~\ref{fig:laziness} D) shows that the late-aligned model also stops looking at the image: the fraction of attention that image-patch queries place on other image patches during generation, and the fraction that text queries place on the image patches during understanding, both fall sharply with longer language warm-ups, particularly in the mid and deep layers. So the model directs less of its attention to the image tokens, defaulting instead to the surrounding text context and relying more on language priors.

Collectively, these results may also explain the visual performance drop observed in \S~\ref{subsec:early}. In other words, the stronger the underlying LLM, the more adept it becomes at leveraging its advanced language capabilities to "shortcut" visual problems, a tendency heavily characterized as visual-language prior bias~\citep{han2025learning,luo2024probing,brown2025benchmark,ben2026mirage}. While relying on such language heuristics is not detrimental for all tasks, it prevents the model from treating vision as an equally important, first-class modality. When the language trunk is already highly optimized before vision is introduced, the joint optimization landscape is heavily dominated by existing language priors. This phenomenon may partially explain why many powerful multimodal LLMs still occasionally act blind or make elementary visual errors~\citep{tong2024eyes,han2025learning}: their vision pathways are simply not fully engaged. The impact of this visual under-optimization is qualitatively shown in generation tasks, where late-aligned models tend to produce simplified shapes and fail in detailed attribute binding (Appendix~\ref{sec:additional_generation_results}). This dynamic also represents a prominent form of cross-modal bias, echoing the physics-based phenomenological characterization in~\citet{kim2026physics} where optimization trajectories are dominated by pre-existing representational shortcuts. Early unification is therefore not merely a scheduling choice, but a mechanical necessity required to force the active, unified co-evolution of modalities from scratch, also reflecting conclusions from native multimodal scaling laws~\citep{shukor2025scaling}. 



\section{Designing Unified Pretraining Recipes }
\label{sec:recipes}

In this section, we synthesize our empirical findings into actionable, high-performance recipes for unified multimodal pretraining. Our prior analysis of knowledge flow (\S~\ref{section:flowstudy}) reveals a clear asymmetry: language serves as a universal booster, and visual understanding acts as a strong prior for generation, but generation offers minimal backward transfer. This implies that an optimal recipe should allocate the vast majority of the compute budget to language and understanding to efficiently "drive" generative capabilities. We first validate this asymmetric data mixture hypothesis (\S~\ref{subsec:mixingratio}), and then integrate all our architecture and dynamic insights to validate these recipes at scale (\S~\ref{subsec:scaling}).

\subsection{Optimizing data mixing ratios}
\label{subsec:mixingratio}
\finding{6}{Visual generation is highly data efficient when trained jointly with more understanding and language. This highly asymmetric data mixture, combined with early unification and MoE architectures, enables unified multimodal models to scale efficiently to achieve strong overall performance.} 

\paragraph{Settings.} To validate whether dominant language and understanding priors can effectively drive generation, we execute an extensive grid search of the mixing ratios across 1T token pretraining runs. This search is organized along three axes. First, the "Fix MM" setting sweeps the language fraction from 10\% to 90\% while keeping the remaining visual data split evenly between understanding and generation. Second, the "Fix Lan" setting anchors language at 50\% and sweeps the internal balance of understanding versus generation. Finally, having identified 70\% as an optimal point for language preservation, the "Next" sweep anchors language at 70\% and fine-tunes the internal visual ratio.

\begin{table*}[t]
\centering
\sisetup{
  round-mode=places,
  detect-weight,
  mode=text,
  round-pad=false
}
\footnotesize
\setlength{\tabcolsep}{3.4pt}
\begin{tabular}{lS[table-format=2.0]S[table-format=2.0]S[table-format=2.0]S[table-format=2.2,round-precision=2]S[table-format=2.2,round-precision=2]S[table-format=2.1,round-precision=1]S[table-format=2.1,round-precision=1]S[table-format=2.1,round-precision=1]S[table-format=2.1,round-precision=1]S[table-format=2.1,round-precision=1]S[table-format=1.3,round-precision=3]S[table-format=1.3,round-precision=3]S[table-format=1.3,round-precision=3]S[table-format=1.4,round-precision=4]}
\toprule
{} & \multicolumn{3}{c}{\textbf{Mix \%}}
  & \multicolumn{2}{c}{\textcolor{LangCol}{\textbf{Language}}}
  & \multicolumn{5}{c}{\textcolor{UndCol}{\textbf{Visual Understanding}}}
  & \multicolumn{4}{c}{\textcolor{GenCol}{\textbf{Visual Generation}}} \\
\cmidrule(lr){2-4}
\cmidrule(lr){5-6}
\cmidrule(lr){7-11}
\cmidrule(lr){12-15}
{} & {L} & {U} & {G}
  & {\textcolor{LangCol}{PPL $\downarrow$}} & {\textcolor{LangCol}{Acc $\uparrow$}}
  & {\textcolor{UndCol}{Gen $\uparrow$}} & {\textcolor{UndCol}{Know $\uparrow$}} & {\textcolor{UndCol}{OCR $\uparrow$}} & {\textcolor{UndCol}{V-Ctr $\uparrow$}} & {\textcolor{UndCol}{Avg $\uparrow$}}
  & {\textcolor{GenCol}{DPG $\uparrow$}} & {\textcolor{GenCol}{GenEval $\uparrow$}} & {\textcolor{GenCol}{CLIP-Sim $\uparrow$}} & {\textcolor{GenCol}{DiffLoss $\downarrow$}} \\
\midrule
\multirow{9}{*}{\textbf{Fix MM}} & 10 & 45 & 45 & 19.27 & 41.89 & 37.9 & 26.9 & 22.9 & 42.9 & 32.7 & 0.326 & 0.148 & 0.256 & 0.2984 \\
 & 20 & 40 & 40 & 17.51 & 44.46 & 45.7 & 30.8 & 24.0 & 44.0 & 36.1 & 0.395 & 0.189 & 0.273 & 0.2802 \\
 & 30 & 35 & 35 & 16.73 & 45.03 & 43.1 & 29.9 & 24.9 & 42.2 & 35.0 & 0.361 & 0.183 & 0.273 & 0.2826 \\
 & 40 & 30 & 30 & 16.31 & 45.79 & 46.1 & 31.2 & 24.2 & 45.6 & 36.8 & 0.331 & 0.186 & 0.269 & 0.2946 \\
 & 50 & 25 & 25 & 15.98 & 46.59 & 47.2 & 30.9 & 25.4 & 44.6 & 37.0 & 0.385 & 0.219 & 0.274 & 0.2804 \\
 & 60 & 20 & 20 & 15.81 & 46.70 & 44.6 & \bfseries 33.0 & 24.8 & 43.9 & 36.6 & 0.387 & 0.203 & 0.275 & 0.2883 \\
 & 70 & 15 & 15 & 15.68 & 46.99 &  48.1 & 32.3 & 25.2 & 46.6 & 38.1 & 0.399 & 0.219 & 0.273 & 0.2996 \\
 & 80 & 10 & 10 & 15.57 & 46.85 & 45.9 & 32.8 & 24.0 & 45.2 & 37.0 & 0.388 & 0.216 & 0.271 & 0.2868 \\
 & 90 &  5 &  5 & \bfseries 15.48 & \bfseries 48.08 & 43.7 & 31.8 & 21.4 & 46.3 & 35.8 & 0.336 & 0.204 & 0.273 & 0.2894 \\
\midrule
\multirow{9}{*}{\textbf{Fix Lan}} & 50 &  5 & 45 & 16.05 & 45.26 & 43.8 & 29.4 & 23.7 & 43.1 & 35.0 & 0.358 & 0.206 & 0.273 & \bfseries 0.2785 \\
 & 50 & 10 & 40 & 16.03 & 46.18 & 45.9 & 31.0 & 24.2 &  45.6 & 36.7 & 0.401 & 0.221 & \bfseries 0.281 & 0.2801 \\
 & 50 & 15 & 35 & 16.06 & 46.42 & 46.6 & 32.1 & 23.4 & 45.4 & 36.9 & 0.370 & 0.193 & 0.273 & 0.2834 \\
 & 50 & 20 & 30 & 16.02 & 46.17 & 46.8 & 31.2 & 23.5 & 45.2 & 36.7 & 0.390 & 0.199 & 0.274 & 0.2909 \\
 & 50 & 25 & 25 & 15.98 & 46.59 & 47.2 & 30.9 & 25.4 & 44.6 & 37.0 & 0.385 & 0.219 & 0.274 & 0.2804 \\
 & 50 & 30 & 20 & 16.01 & 46.05 & 45.5 & 31.1 & 25.2 & 44.7 & 36.6 & 0.370 & 0.208 & 0.273 & 0.2893 \\
 & 50 & 35 & 15 & 16.05 & 45.85 & 47.0 & 32.5 & 23.8 & 46.0 & 37.3 & 0.399 & 0.199 & 0.275 & 0.2821 \\
 & 50 & 40 & 10 & 16.00 & 46.14 & 47.7 & 32.9 & \bfseries 26.5 & 45.7 & 38.2 &  0.420 & 0.216 & 0.276 & 0.2874 \\
 & 50 & 45 &  5 & 16.03 & 46.14 & 46.9 & 32.3 & 25.8 & 46.1 & 37.8 & 0.392 & 0.200 & 0.269 & 0.2931 \\
 \midrule
\multirow{5}{*}{\textbf{Next}} & 70 &  5 & 25 & 15.67 & 46.55 & 47.0 & 32.9 & 23.9 & 43.5 & 36.8 & 0.375 & 0.204 & 0.271 & 0.2823 \\
 & 70 & 10 & 20 & 15.71 & 46.34 & 46.8 & 30.6 & 22.7 & 44.3 & 36.1 & 0.358 & 0.206 & 0.273 & 0.2855 \\
 & 70 & 15 & 15 & 15.68 & 46.99 & 48.1 & 32.3 & 25.2 & 46.6 & 38.1 & 0.399 & 0.219 & 0.273 & 0.2996 \\
 & 70 & 20 & 10 & 15.65 & 46.65 & 48.0 & 31.9 & 26.3 & 46.3 & 38.1 & 0.401 & 0.221 & 0.272 & 0.2934 \\
 & 70 & 25 &  5 & 15.68 & 46.86 & \bfseries 48.3 & 32.7 & 25.8 & \bfseries 47.1 & \bfseries 38.5 & \bfseries 0.450 & \bfseries 0.237 & 0.275 & 0.2868 \\ 
\bottomrule
\end{tabular}
\caption{\textbf{Extensive grid search of data mixing ratios across three axes.} We evaluate models on Language, Visual Understanding, and Visual Generation. The searches confirm that while language requires a dominant token share, visual capabilities peak at highly specific and asymmetrical ratios. The optimal configuration emerges in the "Next" sweep at a 70/25/5 split for Language, Understanding, and Generation respectively.}
\label{tab:part7_mixing_ratio}
\end{table*}

\paragraph{Results.} The sweeps detailed in Table~\ref{tab:part7_mixing_ratio} reveal asymmetric data volume requirements across different tasks. Language requires the vast majority of the compute budget to maintain core reasoning skills. Visual understanding benefits from a moderate allocation of around 25\%. Interestingly, visual generation reaches near-peak performance with surprisingly little data. As shown in the "Next" sweep, reducing the generation allocation to just 5\% dramatically boosts visual understanding. General VQA hits 48.3 in this setting while simultaneously achieving the highest generation evaluation scores.  These results validate that pretraining unified models relies on respecting modality asymmetry. Because visual generation provides minimal backward transfer (\S3), forcing a balanced data mixture simply wastes training capacity on low-yield tokens. By heavily skewing the training budget toward language and understanding, the model acquires foundational priors that effectively drive generation.

\subsection{Scaling recipes}
\label{subsec:scaling}

To validate our findings at scale, we evaluate each of our three core findings and design choices—data mixture recipes, model architecture, and unification timing—through controlled, single-variable comparisons:

(1) \textbf{Knowledge flow (\S~\ref{section:flowstudy}):} Guided by the finding that knowledge flow is asymmetric, we apply the identified skewed data mixture (L70/U25/G5, representing Language, Visual Understanding, and Visual Generation ratios, respectively). To isolate the impact of this recipe, we compare it against a Balanced Recipe Baseline (L50/U25/G25) while keeping the MoE architecture and early unification strategy identical.

(2) \textbf{Architecture (\S~\ref{sec:synergy}):} To satisfy the conditions for modality synergy, where attention and normalizations are shared while FFNs are separated, we transition naturally to an MoE architecture. Specifically, we scale to a 13.5B parameter MoE model (with 1.5B active parameters per token)~\citep{tong2026beyond} consisting of 256 experts, routing to the top 16 experts per token. To enforce modality decoupling, we fix two of the active experts to be modality-specific (one dedicated to language and one to vision), while the remaining 14 are dynamically routed. We compare this setup against a Dense Baseline (a 3.5B dense model) trained under the same balanced data mixture (L50/U25/G25) and early unification training setting.

(3) \textbf{Early unification (\S~\ref{sec:earlyunification}):} We enforce early unification and simultaneous training from scratch to encourage native visual learning. To assess the importance of early-stage integration, we compare this against a Late-Fusion Baseline using the same MoE and balanced data recipe (L50/U25/G25). In this baseline, training begins with language-only data, and vision tokens are introduced only after 60\% of the training progress. To ensure a fair comparison, the total volume of vision tokens is kept identical to the early unification setup by packing them more densely in the remaining 40\% of the training.

\begin{table*}[t]
\centering
\sisetup{
  round-mode=places,
  detect-weight,
  mode=text,
  round-pad=false
}
\footnotesize
\setlength{\tabcolsep}{3.4pt}
\begin{tabular}{l
  S[table-format=2.2,round-precision=2]
  S[table-format=2.2,round-precision=2]
  S[table-format=2.2,round-precision=2]
  S[table-format=2.2,round-precision=2]
  S[table-format=2.2,round-precision=2]
  S[table-format=2.2,round-precision=2]
  S[table-format=2.2,round-precision=2]
  S[table-format=1.3,round-precision=3]
  S[table-format=1.3,round-precision=3]
  S[table-format=1.3,round-precision=3]
  S[table-format=1.3,round-precision=3]
}
\toprule
\multirow{2}{*}{\textbf{Model}}
  & \multicolumn{2}{c}{\textcolor{LangCol}{\textbf{Language}}}
  & \multicolumn{5}{c}{\textcolor{UndCol}{\textbf{Visual Understanding}}}
  & \multicolumn{4}{c}{\textcolor{GenCol}{\textbf{Visual Generation}}} \\
\cmidrule(lr){2-3} \cmidrule(lr){4-8} \cmidrule(lr){9-12}
  & {\textcolor{LangCol}{PPL $\downarrow$}}
  & {\textcolor{LangCol}{Acc $\uparrow$}}
  & {\textcolor{UndCol}{Gen $\uparrow$}} & {\textcolor{UndCol}{Know $\uparrow$}}
  & {\textcolor{UndCol}{OCR $\uparrow$}} & {\textcolor{UndCol}{V-Ctr $\uparrow$}}
  & {\textcolor{UndCol}{Avg $\uparrow$}}
  & {\textcolor{GenCol}{DPG $\uparrow$}} & {\textcolor{GenCol}{GenEval $\uparrow$}}
  & {\textcolor{GenCol}{CLIP-Sim $\uparrow$}} & {\textcolor{GenCol}{DiffLoss $\downarrow$}} \\
\midrule
Balanced Recipe
  & 11.97 & 52.86
  & 51.50 & 38.90 & 25.15 & 50.14 & 41.42
  & 0.676 & 0.467 & 0.310 & \bfseries 0.261 \\
Dense Model
  & 12.14 & 52.03
  & 50.12 & 36.66 & 25.43 & 49.74 & 40.49
  & 0.667 & 0.459 & 0.308 & 0.266 \\
Late-Fusion 
  & 12.25 & 51.78
  & 49.89 & 37.03 & 26.22 & 49.50 & 40.66
  & 0.672 & 0.471 & 0.308 & 0.269 \\
\textbf{Full}
  & \bfseries 11.67 & \bfseries 54.31
  & \bfseries 53.63 & \bfseries 40.11 & \bfseries 27.23 & \bfseries 51.33 & \bfseries 43.08
  & \bfseries 0.689 & \bfseries 0.482 & \bfseries 0.312 & 0.272 \\
\bottomrule
\end{tabular}
\caption{\textbf{Scaling results and controlled baseline comparisons.} We evaluate our model against three controlled baselines to validate our main design choices: data mixture recipes (Balanced Recipe), architecture design style (Dense Model), and vision alignment strategy (Late-Fusion).}
\label{tab:part8_sweep_compare}
\end{table*}

\textbf{Results.} We pretrain our 13.5B MoE models on a 2T token budget to validate each design choice through controlled, single-variable comparisons, using the Balanced Recipe as our primary baseline reference. Results are shown in Table~\ref{tab:part8_sweep_compare}. 

\paragraph{Knowledge flow (Asymmetric vs. Balanced Recipe).} We compare Full (L70/U25/G5 mix) against the Balanced Recipe (L50/U25/G25). Shifting to this asymmetric mix improves language accuracy (54.31\% vs. 52.86\%) and the visual understanding average (43.08\% vs. 41.42\%). Text-to-image alignment also improves (GenEval rises to 0.482 from 0.467; DPG to 0.689 from 0.676) despite using five times fewer generative tokens. While this asymmetric mix leads to a minor increase in diffusion loss (0.272 vs. 0.261), the pure image generative quality remains highly consistent; indeed, a Fréchet Inception Distance (FID)~\citep{fid} evaluation on 50k generated images yields 5.234 for the asymmetric mix, which is highly competitive with the 5.131 achieved by the balanced recipe. These overall results suggest that strong language and understanding priors effectively bootstrap visual generation. 

Pretraining everything from scratch enables the model to acquire strong visual generation capabilities even with a small fraction of generation tokens, as language and visual understanding priors naturally transfer to bootstrap generative modeling early on. Nevertheless, increasing the generation ratio in later stage of pretraining or introducing a dedicated visual generation midtraining stage remains beneficial for achieving better generative modeling.

\paragraph{Architecture (MoE vs. Dense).} To evaluate architectural choices, we compare the Balanced Recipe (MoE) against the Dense Model under a constant data mix and early unification. The MoE-based Balanced Recipe consistently outperforms the dense baseline, improving language accuracy (52.86\% vs. 52.03\%) and the visual understanding average (41.42\% vs. 40.49\%), while lowering visual generation diffusion loss (0.261 vs. 0.266). These results support the design of shared attention and normalization with modality-specific FFNs, wherein the MoE model outperforms the 3.5B Dense Model with 1.5B active parameters.

\paragraph{Early unification (Early vs. Late).} To isolate the impact of training dynamics, we compare the Balanced Recipe (early-fusion training) against the Late-Fusion baseline under a constant MoE setup. Early unified pretraining provides clear advantages over late alignment, yielding higher language accuracy (52.86\% vs. 51.78\%) and a higher visual understanding average (41.42\% vs. 40.66\%). Early unification also achieves a lower visual generation diffusion loss (0.261 vs. 0.269), indicating that early joint training allows modalities to co-evolve more effectively. We provide qualitative text-to-image generation comparisons between these two configurations in Appendix~\ref{sec:additional_generation_results}.

In summary, these large-scale evaluations consistently validate the insights derived from our controlled, smaller-scale studies. The alignment of our findings suggests that the principles of asymmetric knowledge flow, architectural decoupling, and early unification generalize reliably to larger compute regimes. 



\section{Related Work}
\label{section:related}

\subsection{Unified models}
Multimodal models have rapidly advanced beyond treating vision simply as a conditional input for text models~\citep{li2024aria,shukor2025scaling,wu2026scalingnativemultimodalpretraining}. Instead, current research actively pursues unified architectures capable of simultaneous visual comprehension and generation~\citep{zhang2025unified,xiao2025mindomni,geng2025x,xu2025tbac,xin2025lumina,li2025onecat,wang2025lightbagel,wei2025univideo,wang2025ovis,li2025uniworld,nguyen2025oneflow,metaquery,li2025synergen,zhang2026nextflow,dai2023emu,emu2,ge2024seed,dong2023dreamllm,tong2024metamorph,metaquery,han2025tv2tv,fu2026lanceunifiedmultimodalmodeling,chen2025blip3o,wang2026representation,wang2026uniddtunifyingmultimodalunderstanding,lin2026gearguidedendtoendautoregression,lin2025exploring}. Early unified paradigms relied on quantizing images into discrete visual tokens~\citep{vqvae,vqvae2,rqvae} for autoregressive modeling~\citep{lu2022unified,aghajanyan2022cm3,team2024chameleon,lu2024unified,emu3,schlarmann2025fuselip}. More recently, hybrid methods such as Transfusion~\citep{zhou2024transfusion,lmfusion,tong2026beyond} have successfully combined continuous diffusion mechanisms with discrete language prediction, establishing a highly effective standard for joint modality modeling. In parallel, advancements in unified vision representation spaces have also emerged to bridge perception and generation~\citep{scale-rae-2026,zheng2025diffusion,han2026vision,li2025manzanosimplescalableunified,meituanlongcatteam2026longcatnextlexicalizingmodalitiesdiscrete,uniflow,jiao2025unitoken,atoken,liu2026tuna,anlin2026vision,zheng2025hita,gui2025adapting,jia2025dino,chen2026ideal,zhao2025qlip,wang2024image,pan2026repfusionleveragingmultimodalpriors,singh2026raev2,jin2026latentum,zhang2026hydra,yu2026rae,peng2026uniar,li2026sparsemanticpixelselfalignmentadaptive,wu2024vila}.
Building on these advancements, our work delves into the physics of multimodal pretraining, hoping to pave the way for the design and scaling of future unified models.

\subsection{Vision integration} A prevailing methodology for constructing multimodal systems relies on late-fusion. This involves stitching together pretrained LLMs~\citep{touvron2023llama, grattafiori2024llama3} with independently pretrained visual encoders for visual understanding~\citep{li2023blip, liu2023visual, alayrac2022flamingo,bai2025qwen3} via adapters and post-hoc fine-tuning. Because rich visual signals are forced to conform to a pre-existing language space, this approach inherently bottlenecks capabilities~\citep{tong2024cambrian}. To overcome this, the field has increasingly shifted toward early-fusion, integrating vision from the very beginning of training to allow visual and language representations to co-evolve. These early-fusion efforts focused primarily on deep, native visual understanding~\citep{team2026kimi, meta2025llama, comanici2025gemini,thinkingmachines2026interactionmodels}. The paradigm has expanded beyond visual understanding towards visual generation as well.  By jointly modeling perception and creation, this unification enables the system to evolve towards world models~\citep{tong2026beyond}. This deeper synergy unlocks stronger visual interactive capabilities, which are essential for seamlessly supporting complex, multi-modal reasoning like interleaved generation~\citep{wise,yang2026omni} and real-world tasks like robotics~\citep{hu2026bagelvla,intelligence2026pi,aditi2026cosmos3omnimodalworld}. In this work, we explore the underlying physics of multimodal pretraining by systematically investigating unification timing and modality sequences, demonstrating that early unification prevents vision laziness and encourages stronger visual learning.

\subsection{Knowledge flow} Recent benchmarking observations highlight that unified multimodal models often surpass specialized text-to-image systems on complex generation tasks~\citep{wise, li2026ueval,wang2026quantifying,yang2026reasoning,wang2026visgym}, hinting that language and understanding capabilities fundamentally benefit visual generation~\citep{tong2024metamorph,mogao,han2025learning}. We formalize this phenomenon by rigorously dissecting the knowledge flow between modalities using both web-scale real-world distributions and strictly controlled synthetic environments. At the high level, we show that capability transfer is highly asymmetric: language acts as a universal booster, and visual understanding heavily drives generation, whereas visual generation offers minimal backward transfer. Furthermore, our controlled synthetic study uncovers that this cross-modal knowledge flow is deeply concept-dependent. Guided by these findings, which show that generative modeling requires strong understanding and language priors, we derive highly asymmetric data mixing recipes for unified pretraining at scale.

\subsection{Architecture designs in unified models} Within the design space of unified multimodal models, maximizing cross-modal synergy while minimizing modality competition remains a central challenge. A critical debate persists regarding the optimal degree of parameter sharing. Some approaches utilize a monolithic transformer for all modalities~\citep{zhou2024transfusion, janus, ma2025janusflow, chen2025januspro,wu2024liquid}, whereas others forcibly isolate parameters via modality-specific feed-forward networks (FFNs)~\citep{lmfusion, lin2024moma} or separate attention blocks~\citep{liang2024mixture,deng2025bagel}. At the most decoupled design, frameworks can maintain highly separated networks, sharing almost no parameters and passing LLM-derived text features as conditioning signals to an independent diffusion model for generation~\citep{wu2025qwen,chen2025pixels,metaquery}.
Our investigation explores how task complexity and architectural choices dictate competition and synergy during unified training. By localizing the exact sources of modality interactions, we confirm that shared attention promotes synergy, while decoupled FFNs prevent competition, providing strong empirical support for MoE architectures~\citep{meituanlongcatteam2026longcatnextlexicalizingmodalitiesdiscrete,tong2026beyond}. Furthermore, the design for vision encoders and decoders remains an open question~\citep{han2026vision,fan2025scaling,scale-rae-2026,meituanlongcatteam2026longcatnextlexicalizingmodalitiesdiscrete,liu2026tuna}; to explore this, we systematically investigate three prevalent vision representation designs and evaluate their specific impacts on knowledge flow.

\section{Discussion and Conclusion}
\label{section:conclusion}

In this work, we present a systematic exploration into the underlying physics of unified multimodal pretraining. By moving beyond prevailing heuristics and conducting rigorously controlled experiments across both real-world and synthetic environments, we demystify the fundamental mechanisms of modality knowledge flow, architectural synergy, and training dynamics. We demonstrate that multimodal transfer is inherently asymmetric and concept-dependent, explain the necessity of early joint training to prevent vision laziness, and identify data and architectural designs that maximize cross-modal synergy. 

Together, these insights establish a more solid foundation for training the next generation of unified foundation models, which are increasingly adopting early-fusion architectures to co-evolve representations from the very beginning~\citep{comanici2025gemini,meta2025llama,team2026kimi,thinkingmachines2026interactionmodels}. A defining takeaway from our study is the highly efficient, scalable manner in which diverse modalities can co-evolve. Rather than treating generative modeling as an isolated, resource-heavy objective to be aligned post-hoc, we show that it can be powerfully bootstrapped. Because discriminative understanding and language act as strong foundational priors, our optimized recipes suggest that strong generative fidelity can be achieved with only a small fraction of the training budget. By demonstrating that generative capabilities can be unlocked at a low computational cost, we show that native generation can be seamlessly integrated into general-purpose foundation models without the traditional computational penalty. This asymmetric scaling strategy, combined with decoupled parameter allocation and early-fusion co-evolution, provides a concrete, computationally efficient pathway to future models supporting any-to-any I/O.

Looking forward, resolving the delicate balance between synergy and competition becomes even more critical as we transition toward more omnimodal foundation models supporting arbitrary input-output formats~\citep{aditi2026cosmos3omnimodalworld,yang2026omni,tong2026beyond}. The integration of vision established in this work should naturally extend to continuous video streams, audio, 3D, actions, and even more modalities. By applying our findings on early joint training and parameter decoupling to these richer sensory streams, we can move closer to general-purpose models with highly versatile I/O capabilities. Ultimately, these pretraining principles provide a scalable blueprint for systems with native, high-dimensional I/O, establishing a foundation for a more unified and expressive form of multimodal intelligence.  We envision these principles serving as a stepping stone toward scalable multimodal modeling, driving the evolution of multimodal systems from passive multimodal observers into predictive, interactive world models.

\section{Acknowledgment}
\label{section:ack}

We would like to thank Zeyuan Allen-Zhu for fruitful discussions on the vision-related aspects of the physics of language models. We are also grateful to Emily Dinan, John Nguyen, Amir Bar, Xiaochuang Han, Weijia Shi, Boyang Zheng, Jianyuan Wang, Yuwei Niu, Weiyang Jin, Chaorui Deng, Xichen Pan, Ellis Brown, Marjan Ghazvininejad, Jakob Verbeek, Bo Zheng, Songlin Yang, Lili Yu, Dayiheng Liu, Yuren Cong, An Yang, Brian Karrer, Koustuv Sinha, Haoqi Fan, Zihan Ding, Nick Hawes, Andrea Vedaldi, Luke Zettlemoyer, Mary Williamson, and Saining Xie for their helpful discussions and support.

\clearpage
\newpage

\appendix 

\section{Limitations and Future Work}
While this work provides a systematic, bottom-up exploration of unified multimodal pretraining, it is subject to several limitations that open exciting avenues for future research. Because our insights are derived from the fundamental aspects of modality interactions, we hypothesize that our core principles (such as asymmetric knowledge flow, the necessity of early unification, and architectural decoupling) will largely hold true for more complex modalities and extreme scales. However, these frontiers remain to be empirically validated.

First, our empirical study is mostly confined to the domain of text and static images, leaving the exploration of dynamic and continuous modalities, such as video and audio, as an open question. Video introduces an additional temporal dimension, which alters data complexity and computational demands. While we expect our static-image findings to serve as a highly transferable baseline, investigating exactly how the principles of knowledge flow and modality synergy apply to video generation and understanding, and whether temporal reasoning requires different architectural decouplings or data mixing curricula, remains a crucial next step for developing predictive world models.

Second, while our proposed pretraining recipes are validated at a substantial scale (e.g., 13.5B parameters trained on 2T tokens), the underlying physics of modality interaction may continue to evolve at the extreme scales of frontier models (e.g., more than 1T parameters). As parameter capacity increases, the threshold at which modalities transition from synergy to competition might shift. More intriguingly, extreme scaling could unlock entirely new forms of cross-modal interactions. For instance, while we currently observe a strict asymmetry where visual generation provides minimal backward transfer to understanding, a massive model might possess enough capacity to utilize generative modeling as an internal world simulator. At such scales, emergent bidirectional knowledge flows could occur, allowing generative processes to actively assist in complex physical reasoning or advanced language tasks. Future work may explore these phenomena at larger scales to determine whether unified pretraining recipes require further recalibration at the frontier.


\section{Modality Transfer on Real-world Data Additional Results}
\label{appendix:modalitytransfer}

In \S\ref{subsec:part1_start} of the main text, we analyze the modality knowledge flow using our default RAE visual tokenizer/encoder. To verify that our findings on knowledge flow asymmetry and neutrality are applicable to different visual tokenizer designs, we replicate the real-world modality transfer experiments under three alternative vision designs: (1) \textbf{Raw Pixels}, (2) \textbf{CLIP + VAE}, and (3) \textbf{AR (UniTok)}. For these supplementary transfer evaluations, we primarily focus on conditional visual generation.

\begin{figure*}[!h]
    \centering
    \includegraphics[width=\linewidth]{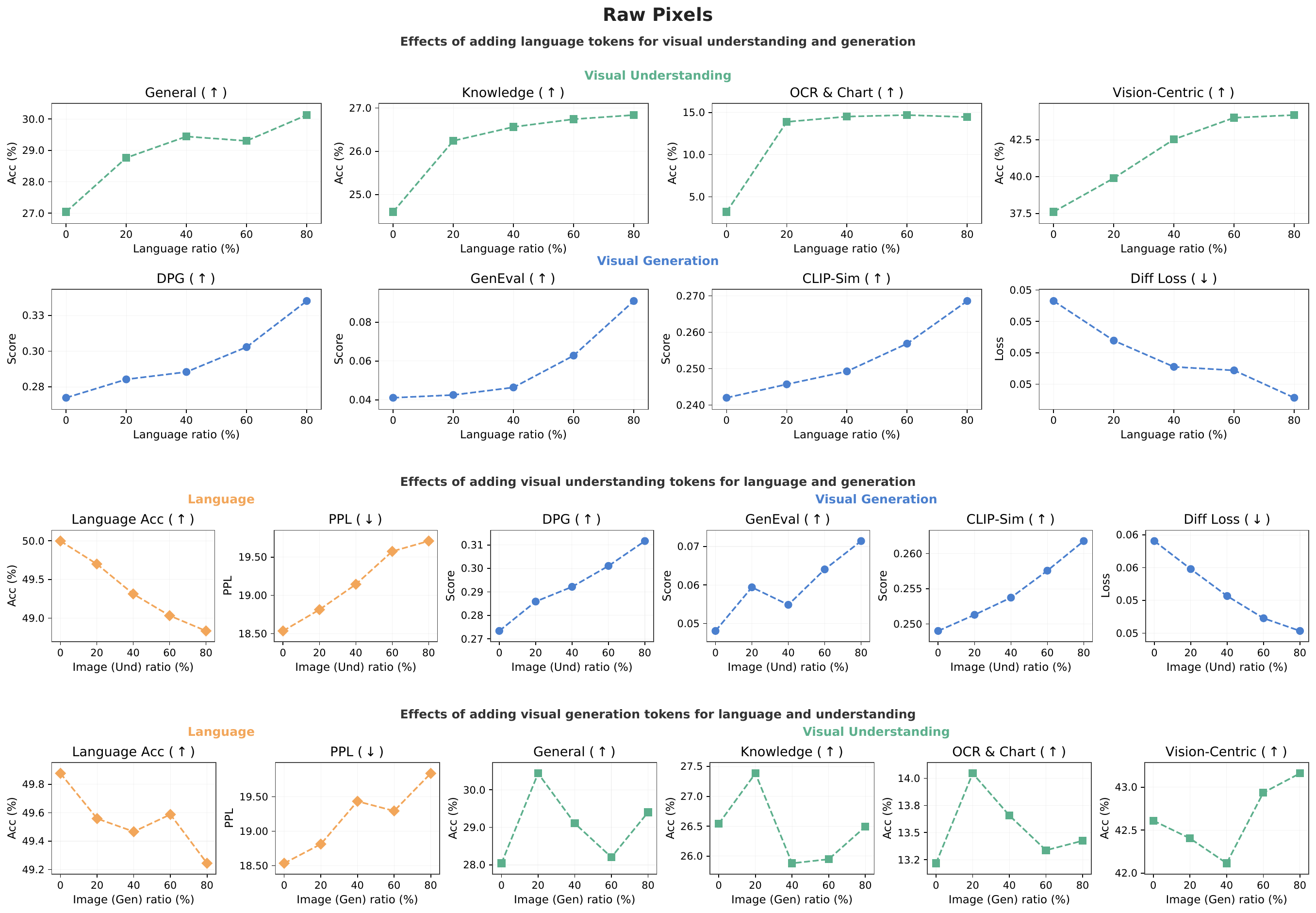}
    \caption{\textbf{Modality transfer results using Raw Pixels.} We replicate the real-world knowledge flow transfer experiments using raw pixels. The results mirror the main text: language universally boosts vision (top), understanding significantly benefits generation (middle), and generation yields a highly stable, neutral effect on language and visual understanding tasks (bottom).}
    \label{fig:sup_part1_raw}
\end{figure*}

\begin{figure*}[!h]
    \centering
    \includegraphics[width=\linewidth]{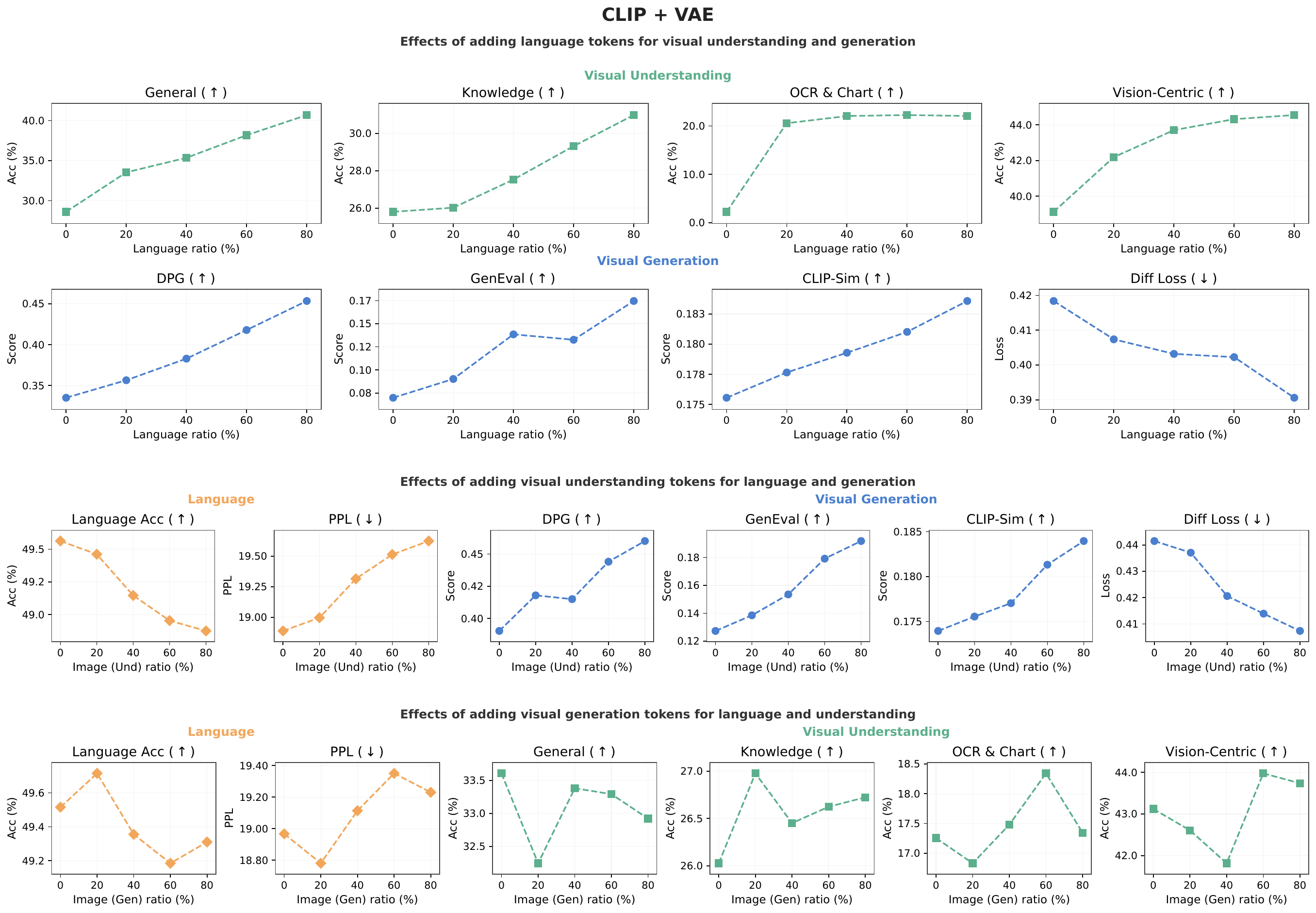}
    \caption{\textbf{Modality transfer results using CLIP + VAE.} We replicate the real-world modality transfer experiments utilizing a SigLIP-2 encoder for visual understanding and an SD3 VAE for generation. The transfer dynamics remain highly consistent, showing that language acts as a universal booster, understanding acts as a strong prior for generation, and generation has a stable, neutral, or slightly fluctuating impact on other abilities.}
    \label{fig:sup_part1_vae}
\end{figure*}

\begin{figure*}[!h]
    \centering
    \includegraphics[width=\linewidth]{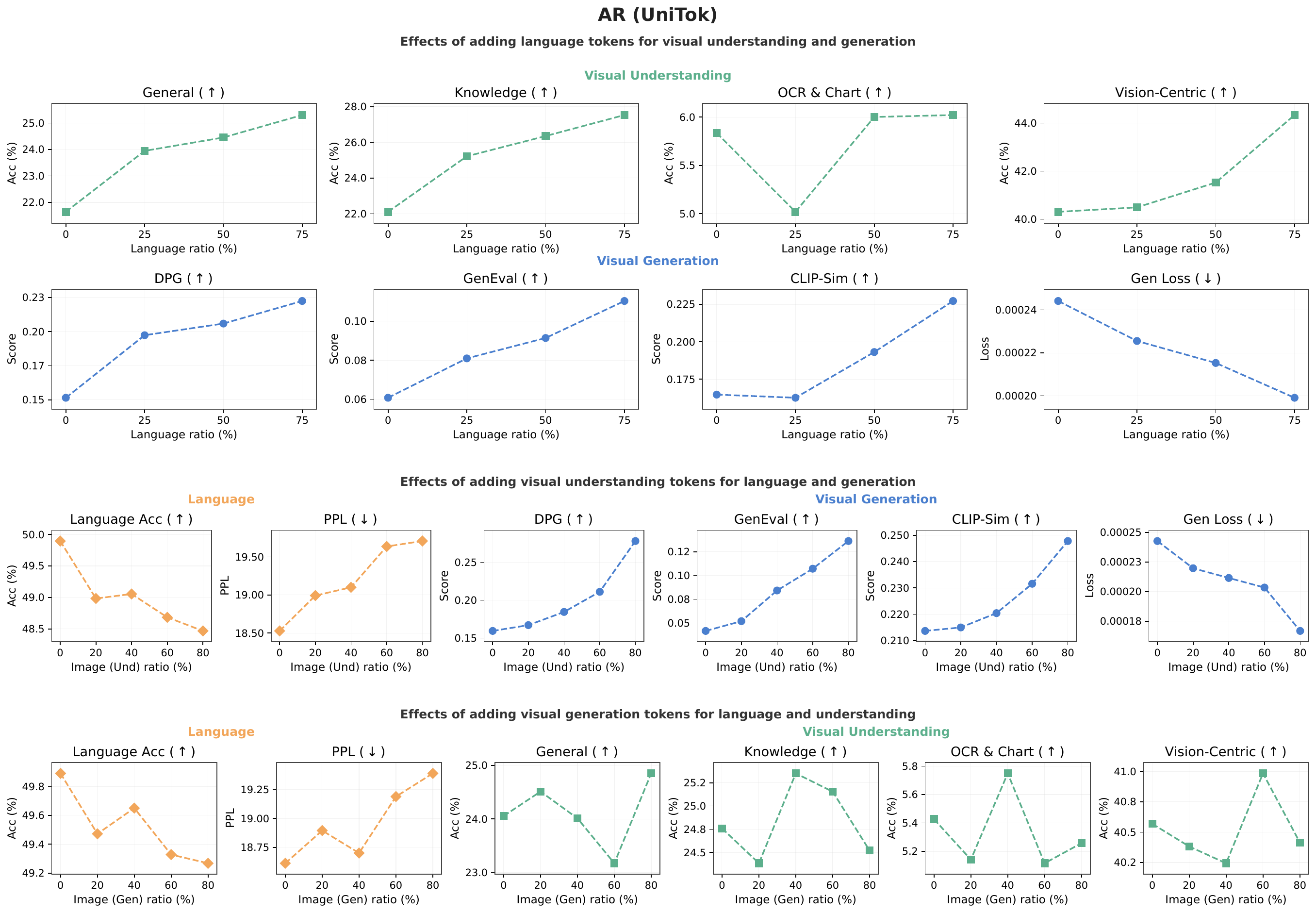}
    \caption{\textbf{Modality transfer results with AR modeling.} We replicate the real-world modality transfer experiments utilizing discrete visual tokenization and autoregressive next-token prediction. Consistent with our continuous diffusion findings, the results demonstrate that language acts as a universal booster (top), visual understanding serves as a strong prior to drive generation (middle), and visual generation exerts a highly stable, neutral effect on other capabilities (bottom). Note that in the language-to-vision transfer experiments (top row), we record slightly different data points.}
    \label{fig:sup_part1_ar}
\end{figure*}

As illustrated in Figure \ref{fig:sup_part1_raw} (Raw Pixels), Figure \ref{fig:sup_part1_vae} (CLIP + VAE), and Figure \ref{fig:sup_part1_ar} (AR modeling), the empirical results exhibit highly consistent behaviors across all three tokenization designs. Across all designs, we consistently observe:
\begin{itemize}
    \item \textbf{Language as a universal booster:} Across alternative setups, increasing the language data ratio from 0\% to 80\% monotonically and significantly improves all visual understanding metrics as well as generative quality.
    \item \textbf{Understanding as a prior for generation:} Scaling the proportion of visual understanding tokens consistently lowers the diffusion loss and enhances text-to-image alignment scores under three settings, confirming the effective transfer of visual understanding knowledge to generative processes.
    \item \textbf{Neutral effect of visual generation:} Increasing the visual generation compute budget yields a largely neutral, highly stable effect on other capabilities. Rather than triggering any systematic degradation, both language and visual understanding benchmarks exhibit only minor, non-directional fluctuations across the entire range of generation ratios. 
\end{itemize}

These supplementary evaluations confirm that modality knowledge flows are more tied to the learning objectives themselves rather than specific tokenization designs.

\section{CLEVR Generation and Evaluation Details}

\subsection{Generation}

\paragraph{Synthetic dataset environment.} We build our synthetic dataset on top of the CLEVR rendering pipeline. To increase the complexity of visual features, we extend the standard CLEVR asset library by expanding the color and shape vocabularies:
\begin{itemize}
    \item \textbf{Expanded Colors:} Standard colors plus additional custom hues, including \texttt{red}, \texttt{blue}, \texttt{green}, \texttt{gray}, \texttt{brown}, \texttt{purple}, \texttt{cyan}, and \texttt{yellow}.
    \item \textbf{Expanded Shapes:} Standard primitives (\texttt{cube}, \texttt{sphere}, \texttt{cylinder}) along with newly authored shapes: \texttt{cross}, \texttt{ellipse}, \texttt{pentagon}, \texttt{rectangle}, and \texttt{triangle}.
\end{itemize}
Blender renders each scene based on a programmatically sampled scene graph that defines the color, shape, size, and 3D position of every object. Each scene contains between 1 and 7 objects. To ensure clean visual rendering, we programmatically enforce distance constraints during placement to prevent physical intersections or visual overlap. In total, we generate roughly one million scenes for the training mixture and reserve a held-out set as the evaluation split.

\paragraph{Dense and sparse captions.} For each rendered image, we generate a dense caption (prompted to detail every object's shape, size, color, and position) and a sparse caption (prompted to briefly describe the image) using Qwen3-VL-8B-Instruct (temperature 0.2, top-p 0.8, a maximum of 256 tokens). We mix dense and sparse captions together during training. Each caption serves both visual understanding (image $\rightarrow$ text) and visual generation (text $\rightarrow$ image).

\paragraph{Five target concepts.} To analyze knowledge flow, we focus our systematic ablation and evaluation on five specific conceptual axes:
\begin{enumerate}
    \item \textbf{Color:} Discriminating or rendering the color of target objects.
    \item \textbf{Shape:} Recognizing or generating the precise geometry of objects.
    \item \textbf{Relation:} Comprehending spatial directional offsets between objects. We compute the ground-truth directions analytically using the 3D coordinate metadata in the scene graph to support both cardinal and diagonal relations (e.g., \texttt{left}, \texttt{right}, \texttt{front-left}, \texttt{behind-right}).
    \item \textbf{Size:} Categorizing absolute object scales (\texttt{large}, \texttt{medium}, \texttt{small}) based on the rendered dimension, where questions query either the absolute scale of a single object or relative comparisons between multiple objects.
    \item \textbf{Count:} Quantifying the total number of objects in the scene.
\end{enumerate}

\paragraph{VQA pair construction.} We derive VQA pairs directly from the scene graph metadata, ensuring noise-free ground truth by construction. We format these pairs as open-ended, short-answer questions that revolve around the five target concepts. Specifically, we generate questions across several structural types:

\begin{itemize}
    \item \textbf{Attribute identification} queries the color, shape, or scale of a target object (e.g., "What is the shape of the blue object?"). We enforce a uniqueness filter on the referent attributes to ensure the target object is uniquely identifiable, discarding the question if ambiguity exists.
    \item \textbf{Counting and size comparison} queries the total number of objects in the scene (e.g., "How many objects are in the image?"), the count of objects matching specific descriptors (e.g., "How many green cubes are there?"), absolute size categorization, or relative size comparisons between objects (e.g., "Is the sphere larger than the cylinder?").
    \item \textbf{Spatial reasoning} covers single-hop and multi-hop spatial relationships. For single-hop queries, we directly ask for the directional orientation of one object relative to another (e.g., "Where is the sphere relative to the cube?"), where camera-relative directions are analytically computed using 3D coordinate offsets to produce diagonal outputs such as \texttt{front-left} or \texttt{behind-right}. For multi-hop queries, we precompute directional neighbors to formulate sequential paths (e.g., "What color is the object behind the object to the left of the cube?").
\end{itemize}

The corresponding answers are formatted as single-word or short-phrase text generation targets (e.g., \texttt{red}, \texttt{cylinder}, \texttt{three}, or \texttt{front-left}). To prevent training bias, we balance the question generation process so that the total number of VQA pairs for each of the five target concepts is roughly equal.

Each instance derived from a rendered scene consists of one image, two captions (dense and sparse), and one to three VQA questions.

\subsection{Evaluation}

We evaluate concept transfer using a controlled leave-one-concept-out study across two symmetric directions. For every held-out concept $c$, we use $100$ evaluation samples.

\paragraph{Understanding\,$\rightarrow$\,Generation.}
For each held-out concept value $c$ (e.g., color \texttt{yellow}, shape \texttt{sphere}, relation \texttt{front-left}, count \texttt{4+}, size \texttt{large}), we surgically remove any scene whose scene-graph metadata contains the concept $c$ from the generation training data. Consequently, the model never observes any training image containing the ablated target (e.g., no yellow objects), while we leave the understanding data (captions used as image $\rightarrow$ text inputs, and VQA pairs) completely untouched. 

We then ask whether the model can still produce $c$ when prompted, by feeding it ablate prompts that explicitly request the held-out concept. For $c=\text{\texttt{yellow}}$, for example, every ablate prompt mentions a yellow object, sampled from templates such as "\emph{A yellow triangle is on the front-right side of the image}" or "\emph{A yellow rectangle is to the right of a cyan ellipse}"; for $c=\text{\texttt{sphere}}$, each prompt places a sphere in the scene; for $c=\text{\texttt{front-left}}$, each prompt requests a front-left arrangement; for $c=\text{\texttt{4+}}$, each prompt requests a scene containing at least four objects; and for $c=\text{\texttt{large}}$, each prompt requests a large object. The neutral control set draws from the same templates but uses values that remain present in the training mixture. 

We score the resulting images using a VLM-judge protocol. Specifically, Qwen3-VL-8B-Instruct directly processes the generated image alongside the original generative prompt, answering a binary yes/no question targeting the held-out axis (e.g., "does the image depict the correct \textsc{color}\,/\,\textsc{shape}\,/\,\textsc{relation}\,/\,\textsc{size}\,/\,\textsc{count}?" ). We report per-axis accuracy on the ablate set restricted to the axis that we hold out, making the metric directly sensitive to whether the concept successfully appears in the generated pixels.

\paragraph{Generation\,$\rightarrow$\,Understanding.}
The symmetric study removes $c$ from the understanding stream (both captions and VQA pairs that reference $c$) while leaving the generation stream untouched. We then ask whether the model can still answer questions about $c$ at evaluation time. Concretely, we draw the $100$ ablation VQA examples from the held-out scenes whose ground-truth answer is $c$, meaning the model only scores a point when it generates $c$ itself. 

For $c=\text{\texttt{yellow}}$, this setup includes questions such as "\emph{What is the color of the object?}" on a single yellow-object scene, "\emph{What is the color of the triangle?}" on a multi-object scene whose unique triangle is yellow, or "\emph{What is the color of the object to the left of the gray cube?}" . We apply the same construction to the other concepts: for $c=\text{\texttt{sphere}}$, we sample shape questions whose answer is \texttt{sphere}; for $c=\text{\texttt{front-left}}$, we sample spatial-relation questions whose ground-truth direction is \texttt{front-left}; for $c=\text{\texttt{4+}}$, we sample counting questions whose ground-truth count is \texttt{four}; and for $c=\text{\texttt{large}}$, we sample size questions whose ground-truth label is \texttt{large}. We report the average VQA accuracy over these 100 questions for each concept.

\section{CLEVR Additional Results}
\label{sec:clevr_additional_results}

In this section, we provide supplementary quantitative and qualitative results on the synthetic CLEVR dataset to support the analyses and findings presented in \S\ref{subsec:clevr}. 

\begin{figure*}[!h]
    \centering
    \includegraphics[width=\linewidth]{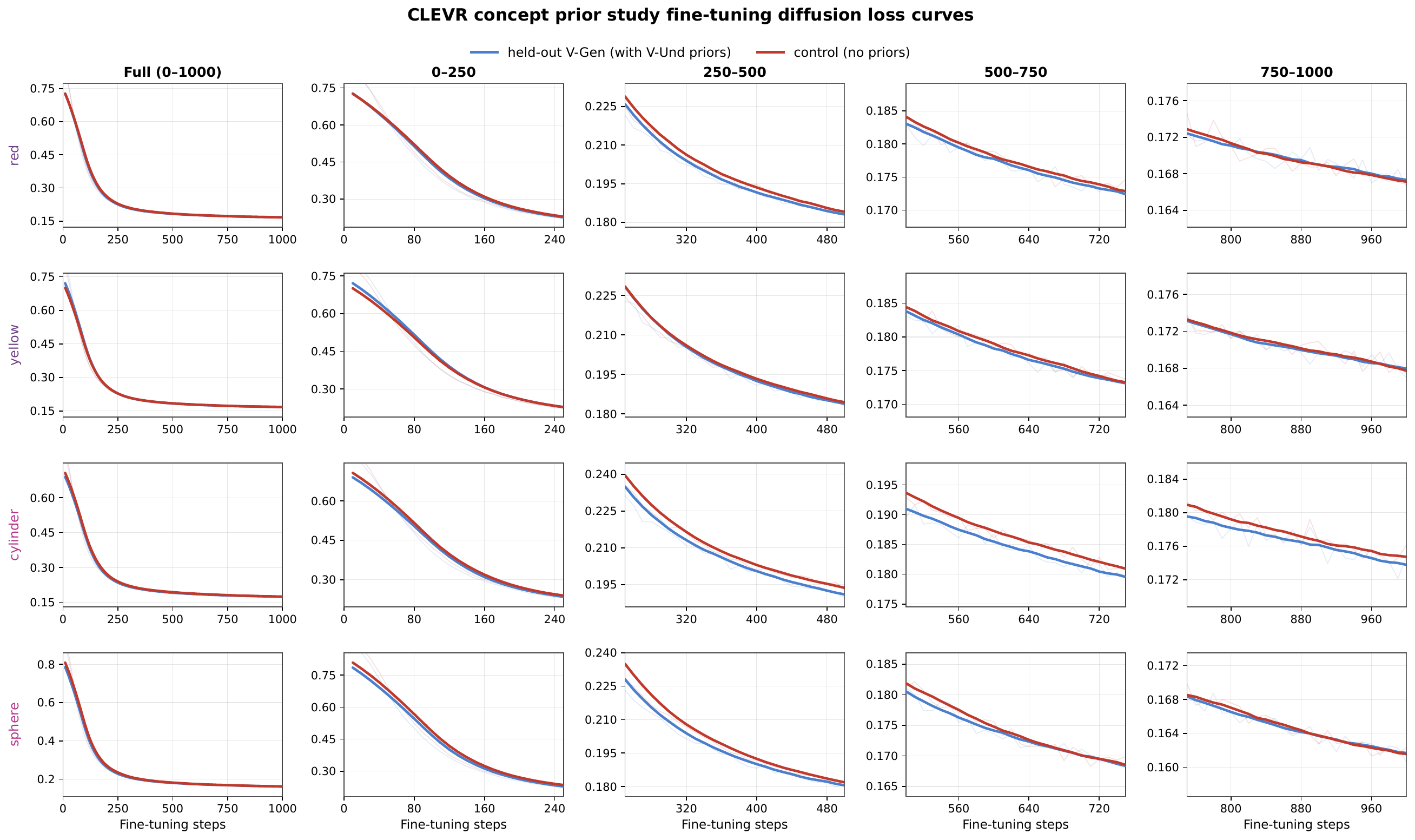}
    \caption{\textbf{Fine-tuning diffusion loss curves for concept recovery.} We show the training trajectories of the diffusion loss during fine-tuning on the generative task for four low-level concepts: red, yellow, cylinder, and sphere. The curves compare the model with visual understanding priors (blue lines) against the control group with no prior exposure (red lines) across different step intervals. The trajectories show that prior exposure via visual understanding provides no observable benefit for subsequent color generation (overlapping closely), but yields a marginal, minor acceleration for shape generation.}
    \label{fig:sup_loss_curve1}
\end{figure*}

\begin{figure*}[!h]
    \centering
    \includegraphics[width=\linewidth]{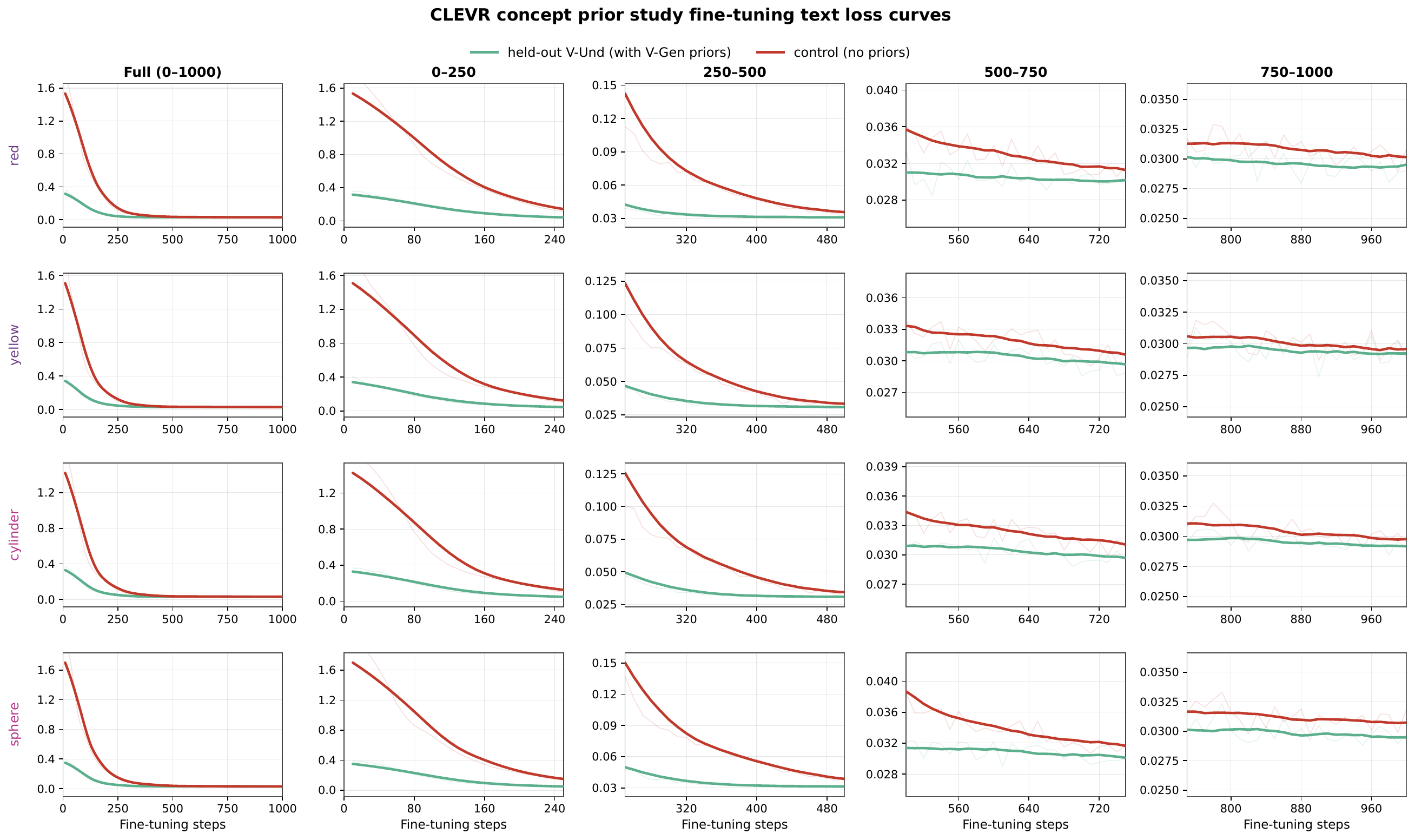}
    \caption{\textbf{Fine-tuning text loss curves for concept recovery.} We show the cross-entropy training loss trajectories during fine-tuning on the understanding (VQA) task for four low-level concepts: red, yellow, cylinder, and sphere. The curves compare the model with visual generation priors (green lines) against the control group with no prior exposure (red lines). These results demonstrate that generative training establishes fine-grained priors in low-level attributes (e.g., pixel-level color and texture distributions) for visual understanding. While not immediately accessible zero-shot, these latent representations serve as strong priors that accelerate learning when mapping visual features to visual understanding.}
    \label{fig:sup_loss_curve2}
\end{figure*}

First, to offer a more granular look at the concept prior study, we plot the detailed training trajectories across the first 1000 steps of optimization process. Figure~\ref{fig:sup_loss_curve1} displays the step-by-step diffusion loss curves when recovering held-out low-level concepts (color and shape) for the generative task under the presence of understanding priors, serving as the supplementary result for the top row of Figure~\ref{fig:prior}. Symmetrically, Figure~\ref{fig:sup_loss_curve2} shows the text cross-entropy loss curves when utilizing generative priors to recover these same concepts for the understanding (VQA) task, serving as the extended data for the bottom row of Figure~\ref{fig:prior}. 

\begin{figure*}[!h]
    \centering
    \includegraphics[width=\linewidth]{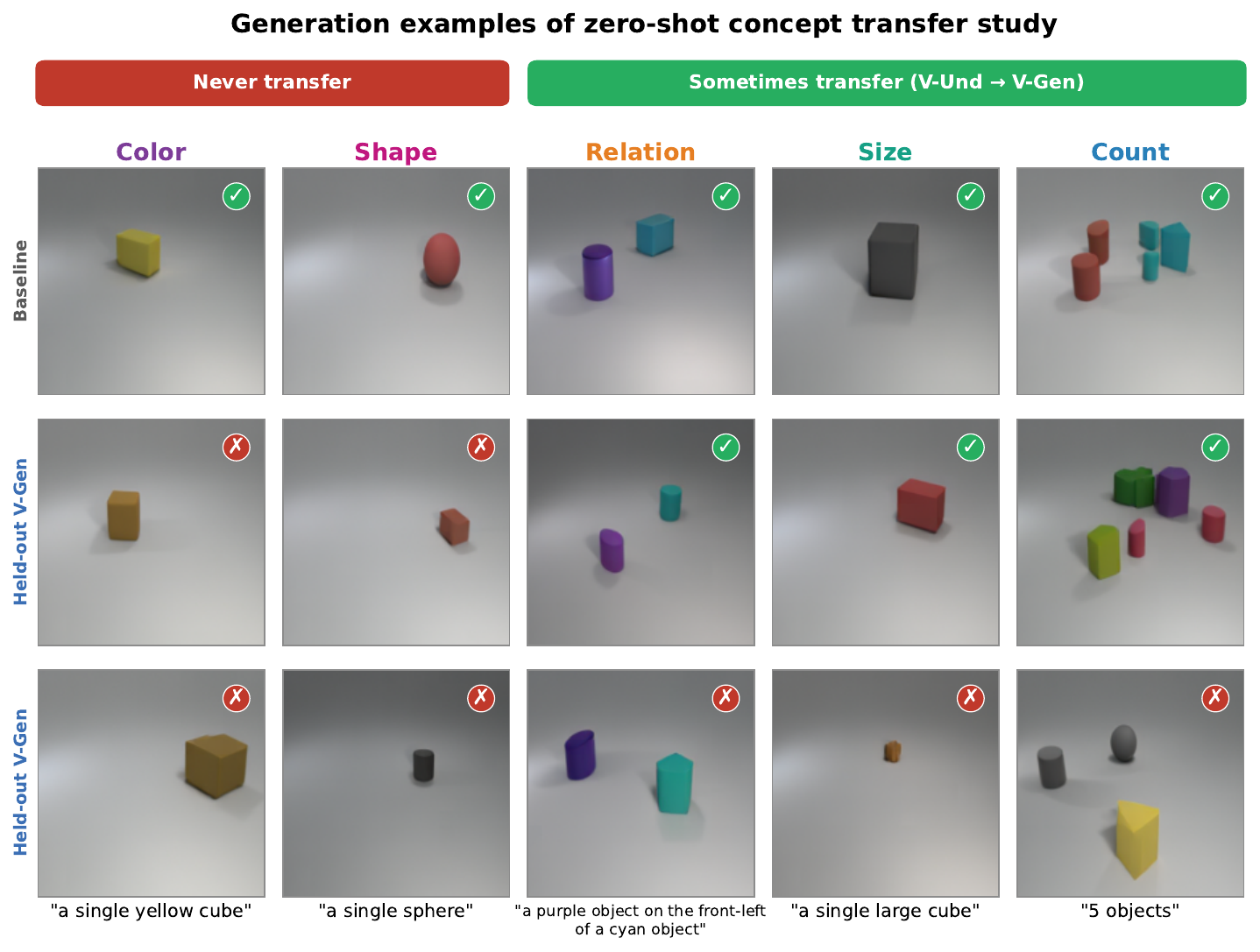}
    \caption{\textbf{Qualitative examples of zero-shot concept transfer.} We present synthetic CLEVR scenes generated under different pretraining configurations. The baseline model (top row) represents joint training on all modalities and concepts. For models where specific concepts are held out from the generation stream but preserved in the understanding stream (middle and bottom rows), zero-shot generation exhibits partial transfer, sometimes successfully rendering higher-level structural concepts such as spatial relations, size, and object count (green checkmarks), but consistently failing to generate low-level attributes like color and shape (red crosses), showcasing the asymmetric, concept-dependent nature of cross-modal transfer.}
    \label{fig:sup_clevr_transfer}
\end{figure*}

Second, to complement the quantitative zero-shot transfer rates shown in Figure~\ref{fig:clevr_ablation} and discussed in \S\ref{subsubsec:clevr_zeroshot}, we provide qualitative visual examples in Figure~\ref{fig:sup_clevr_transfer}. These samples showcase the model's generation capabilities under different ablation settings, visually illustrating where structural knowledge transfer succeeds (such as spatial relations, size, and object count) and where low-level semantic transfer fails (such as color and shape).

\section{Additional Text-to-Image Generation Results}
\label{sec:additional_generation_results}

\begin{figure*}[!h]
    \centering
    \includegraphics[width=0.9\linewidth]{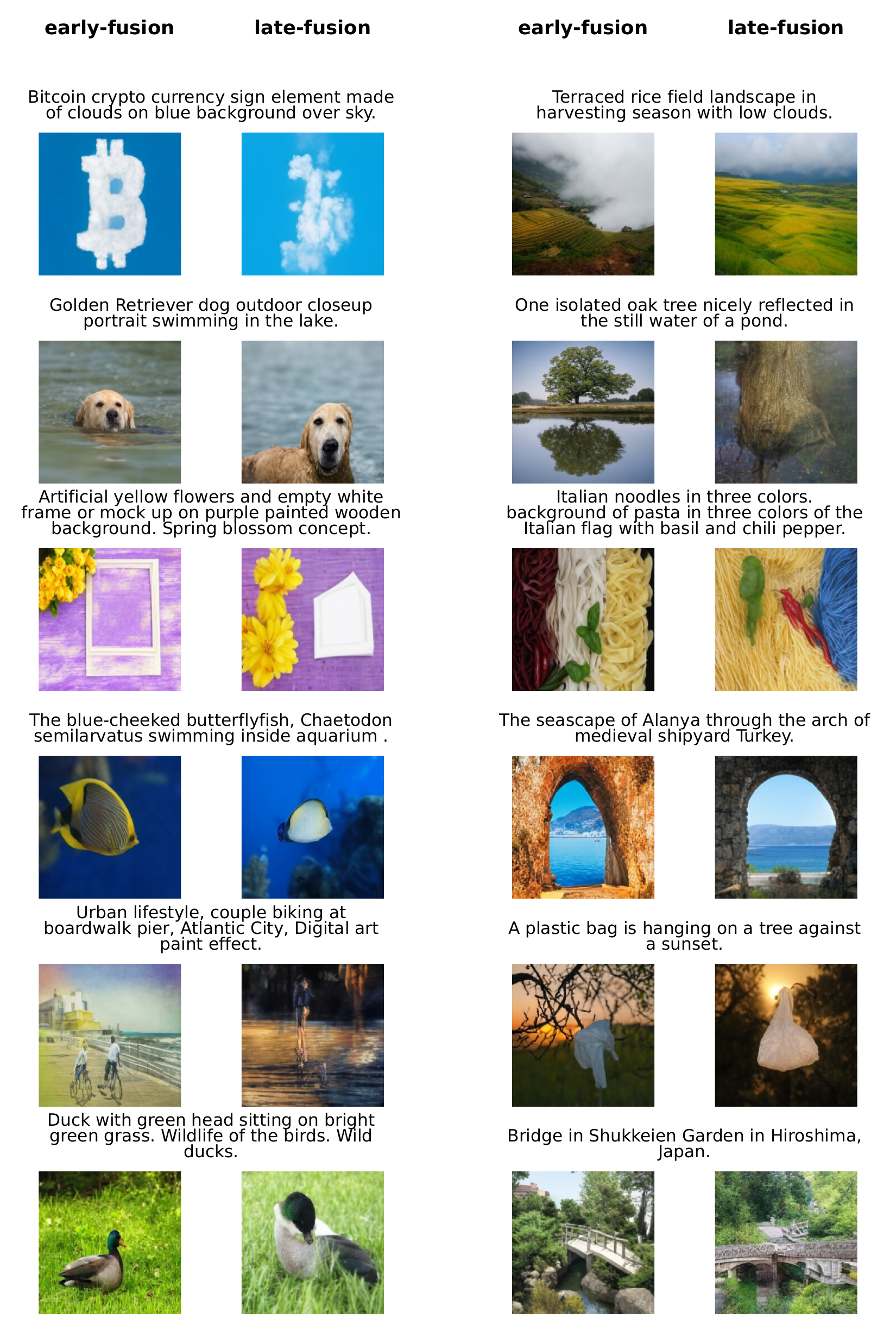}
    \caption{\textbf{Qualitative comparisons of text-to-image generation between early-fusion and late-fusion.} The early-fusion model generally exhibits closer adherence to multi-attribute constraints and sharper geometric structures, whereas the late-fusion baseline tends to produce simplified or abstract shapes.}
    \label{fig:sup_gen}
\end{figure*}

We qualitatively compare text-to-image generation between our early unification model (Balanced Recipe) and the late alignment baseline (Late Fusion). Both are 13.5B MoE models trained under the matched compute budget described in \S~\ref{subsec:scaling}. Figure~\ref{fig:sup_gen} showcases their generated images across diverse prompts.

\paragraph{Qualitative analysis of generative samples.}
A direct comparison of the generated samples shows noticeable differences in how both models handle prompt instructions. For multi-attribute binding prompts, such as \textit{``Italian noodles in three colors...''} or \textit{``Duck with green head sitting on bright green grass,''} the early-fusion model tends to maintain clearer boundaries between distinct elements. In the noodle example, it generates separate red, white, and green segments, whereas the late-fusion model tends to blend these colors together into a generic pasta dish. Similar trends appear in geometric and symbolic representations like \textit{``Bitcoin crypto currency sign element made of clouds,''} where the early-fusion model outlines a recognizable shape while the late-fusion model outputs standard unstructured clouds. Additionally, for prompts demanding specific contextual elements such as reflections or hanging objects, the early-fusion model attempts to render these details more consistently, whereas the late-fusion model sometimes omits them or renders them as highly ambiguous shapes.

\paragraph{Vision laziness in visual generation.}
These qualitative variations are consistent with the vision laziness phenomenon discussed in \S~\ref{subsec:visionlaziness}. When the language pathway is allowed to harden during an extended pure-language pretraining phase, the model may fail to fully optimize its visual pathways, potentially resulting in the simpler geometries and less defined textures observed in the late-fusion outputs. This possible under-optimization of the visual components is highly aligned with the quieter activations shown in Figure~\ref{fig:laziness}. Consequently, the late-fusion model appears to rely primarily on coarse language priors, rendering a standard subject while neglecting specific descriptive adjectives like \textit{``green head''} or \textit{``nicely reflected.''} While qualitative samples cannot definitively prove the underlying optimization dynamics, the observed degradation in multi-attribute binding and structural details in late-fusion aligns well with the quantitative measurements of reduced visual pathway activity.

\clearpage
\label{references}
\addcontentsline{toc}{section}{References}
\bibliographystyle{arxiv-numbered}
\bibliography{paper}

\begin{thebibliography}{162}
\providecommand{\natexlab}[1]{#1}
\providecommand{\url}[1]{\texttt{#1}}
\expandafter\ifx\csname urlstyle\endcsname\relax
  \providecommand{\doi}[1]{doi: #1}\else
  \providecommand{\doi}{doi: \begingroup \urlstyle{rm}\Url}\fi

\bibitem[Aghajanyan et~al.(2022)Aghajanyan, Huang, Ross, Karpukhin, Xu, Goyal, Okhonko, Joshi, Ghosh, Lewis, et~al.]{aghajanyan2022cm3}
[1] Aghajanyan, A., Huang, B., Ross, C., Karpukhin, V., Xu, H., Goyal, N., Okhonko, D., Joshi, M., Ghosh, G., Lewis, M., et~al.
\newblock Cm3: A causal masked multimodal model of the internet.
\newblock \emph{arXiv preprint arXiv:2201.07520}, 2022.

\bibitem[Alayrac et~al.(2022)Alayrac, Donahue, Luc, Miech, Barr, Hasson, Lenc, Mensch, Millican, Reynolds, et~al.]{alayrac2022flamingo}
[2] Alayrac, J.-B., Donahue, J., Luc, P., Miech, A., Barr, I., Hasson, Y., Lenc, K., Mensch, A., Millican, K., Reynolds, M., et~al.
\newblock Flamingo: a visual language model for few-shot learning.
\newblock In \emph{NeurIPS}, 2022.

\bibitem[{Allen-Zhu}(2024)]{AllenZhu-icml2024-tutorial}
[3] {Allen-Zhu}, Z.
\newblock {ICML 2024 Tutorial: Physics of Language Models}, July 2024.
\newblock Project page: \url{https://physics.allen-zhu.com/}.

\bibitem[Anlin et~al.(2026)Anlin, Wen, Zhang, Ma, Wang, Yu, Zhang, and Qi]{anlin2026vision}
[4] Anlin, Z., Wen, X., Zhang, X., Ma, C., Wang, T., Yu, G., Zhang, X., and Qi, X.
\newblock Vision foundation models as effective visual tokenizers for autoregressive generation.
\newblock \emph{Advances in Neural Information Processing Systems}, 38:\penalty0 62656--62675, 2026.

\bibitem[Bambach et~al.(2018)Bambach, Crandall, Smith, and Yu]{bambach2018toddler}
[5] Bambach, S., Crandall, D., Smith, L., and Yu, C.
\newblock Toddler-inspired visual object learning.
\newblock \emph{Advances in neural information processing systems}, 31, 2018.

\bibitem[Ben-Levi et~al.(2026)Ben-Levi, Goldfeder, Zhao, Lapid, LeVi, Roush, Shwartz-Ziv, and Lipson]{ben2026mirage}
[6] Ben-Levi, D., Goldfeder, J., Zhao, W., Lapid, R., LeVi, A., Roush, A.~G., Shwartz-Ziv, R., and Lipson, H.
\newblock Mirage probes: How vision models fake visual understanding.
\newblock \emph{arXiv preprint arXiv:2606.13870}, 2026.

\bibitem[Bisk et~al.(2020)Bisk, Zellers, Gao, Choi, et~al.]{bisk2020piqa}
[7] Bisk, Y., Zellers, R., Gao, J., Choi, Y., et~al.
\newblock Piqa: Reasoning about physical commonsense in natural language.
\newblock In \emph{Proceedings of the AAAI conference on artificial intelligence}, volume~34, pp.\  7432--7439, 2020.

\bibitem[Brazil et~al.(2023)Brazil, Kumar, Straub, Ravi, Johnson, and Gkioxari]{brazil2023omni3d}
[8] Brazil, G., Kumar, A., Straub, J., Ravi, N., Johnson, J., and Gkioxari, G.
\newblock Omni3d: A large benchmark and model for 3d object detection in the wild.
\newblock In \emph{CVPR}, 2023.

\bibitem[Brown et~al.(2025)Brown, Yang, Yang, Fergus, and Xie]{brown2025benchmark}
[9] Brown, E., Yang, J., Yang, S., Fergus, R., and Xie, S.
\newblock Benchmark designers should" train on the test set" to expose exploitable non-visual shortcuts.
\newblock \emph{arXiv preprint arXiv:2511.04655}, 2025.

\bibitem[Chen et~al.(2025{\natexlab{a}})Chen, Xue, Xu, Pan, Yang, Qin, Yan, Zhou, Chen, Huang, et~al.]{chen2025blip3o}
[10] Chen, J., Xue, L., Xu, Z., Pan, X., Yang, S., Qin, C., Yan, A., Zhou, H., Chen, Z., Huang, L., et~al.
\newblock Blip3o-next: Next frontier of native image generation.
\newblock \emph{arXiv preprint arXiv:2510.15857}, 2025{\natexlab{a}}.

\bibitem[Chen et~al.(2025{\natexlab{b}})Chen, Wu, Liu, Pan, Liu, Xie, Yu, and Ruan]{chen2025januspro}
[11] Chen, X., Wu, Z., Liu, X., Pan, Z., Liu, W., Xie, Z., Yu, X., and Ruan, C.
\newblock Janus-pro: Unified multimodal understanding and generation with data and model scaling.
\newblock \emph{arXiv preprint arXiv:2501.17811}, 2025{\natexlab{b}}.

\bibitem[Chen et~al.(2025{\natexlab{c}})Chen, Han, Bai, Tong, Kokkinos, and Torr]{chen2025pixels}
[12] Chen, Y., Han, J., Bai, T., Tong, S., Kokkinos, F., and Torr, P.
\newblock From pixels to feelings: Aligning mllms with human cognitive perception of images.
\newblock \emph{arXiv preprint arXiv:2511.22805}, 2025{\natexlab{c}}.

\bibitem[Chen et~al.(2026)Chen, Diao, Wang, Kong, Ren, He, Jiang, and Wu]{chen2026ideal}
[13] Chen, Y., Diao, Z., Wang, J., Kong, L., Ren, Y., He, B., Jiang, Y.-G., and Wu, Z.
\newblock Ideal: In-depth alignment makes a discrete representation autoencoder.
\newblock \emph{arXiv preprint arXiv:2606.11096}, 2026.

\bibitem[Clark et~al.(2019)Clark, Lee, Chang, Kwiatkowski, Collins, and Toutanova]{clark2019boolq}
[14] Clark, C., Lee, K., Chang, M.-W., Kwiatkowski, T., Collins, M., and Toutanova, K.
\newblock Boolq: Exploring the surprising difficulty of natural yes/no questions.
\newblock \emph{arXiv preprint arXiv:1905.10044}, 2019.

\bibitem[Clark et~al.(2018)Clark, Cowhey, Etzioni, Khot, Sabharwal, Schoenick, and Tafjord]{arc-ce}
[15] Clark, P., Cowhey, I., Etzioni, O., Khot, T., Sabharwal, A., Schoenick, C., and Tafjord, O.
\newblock Think you have solved question answering? try arc, the ai2 reasoning challenge.
\newblock \emph{arXiv preprint arXiv:1803.05457}, 2018.

\bibitem[Dai et~al.(2023)Dai, Hou, Ma, Tsai, Wang, Wang, Zhang, Vandenhende, Wang, Dubey, et~al.]{dai2023emu}
[16] Dai, X., Hou, J., Ma, C.-Y., Tsai, S., Wang, J., Wang, R., Zhang, P., Vandenhende, S., Wang, X., Dubey, A., et~al.
\newblock Emu: Enhancing image generation models using photogenic needles in a haystack.
\newblock \emph{arXiv preprint arXiv:2309.15807}, 2023.

\bibitem[Deng et~al.(2025)Deng, Zhu, Li, Gou, Li, Wang, Zhong, Yu, Nie, Song, Shi, and Fan]{deng2025bagel}
[17] Deng, C., Zhu, D., Li, K., Gou, C., Li, F., Wang, Z., Zhong, S., Yu, W., Nie, X., Song, Z., Shi, G., and Fan, H.
\newblock Emerging properties in unified multimodal pretraining.
\newblock \emph{arXiv preprint arXiv:2505.14683}, 2025.

\bibitem[Diao et~al.(2026)Diao, Wu, Deng, Wang, Bai, Wu, Fan, Ye, Tong, Fan, Li, Wang, Cao, Lin, Yang, Cai, Niu, Zhu, Liu, Lv, Yu, Xie, Wang, Fan, Li, Lu, Ni, Xu, Liang, Shi, Dai, Wang, Qian, Gao, Liu, Sun, Shen, Wang, Ma, Yang, Xie, Li, Zhong, Kong, Shi, Gao, Yao, Wang, Bai, Lin, Yin, Sun, Gong, Wang, Lu, Yang, Liu, and Lin]{diao2026sensenovau1unifyingmultimodalunderstanding}
[18] Diao, H., Wu, P., Deng, H., Wang, J., Bai, S., Wu, S., Fan, W., Ye, W., Tong, W., Fan, X., Li, Y., Wang, Y., Cao, Z., Lin, Z., Yang, Z., Cai, Z., Niu, Y., Zhu, Y., Liu, B., Lv, C., Yu, H., Xie, H., Wang, H., Fan, J., Li, J., Lu, J., Ni, J., Xu, J., Liang, K., Shi, L., Dai, L., Wang, L., Qian, O., Gao, P., Liu, P., Sun, Q., Shen, R., Wang, R., Ma, S., Yang, S., Xie, S., Li, S., Zhong, T., Kong, X., Shi, X., Gao, Y., Yao, Y., Wang, Y., Bai, Z., Lin, Z., Yin, Z., Sun, W., Gong, R., Wang, Q., Lu, L., Yang, L., Liu, Z., and Lin, D.
\newblock Sensenova-u1: Unifying multimodal understanding and generation with neo-unify architecture, 2026.
\newblock URL \url{https://arxiv.org/abs/2605.12500}.

\bibitem[Dong et~al.(2024)Dong, Han, Peng, Qi, Ge, Yang, Zhao, Sun, Zhou, Wei, et~al.]{dong2023dreamllm}
[19] Dong, R., Han, C., Peng, Y., Qi, Z., Ge, Z., Yang, J., Zhao, L., Sun, J., Zhou, H., Wei, H., et~al.
\newblock Dreamllm: Synergistic multimodal comprehension and creation.
\newblock In \emph{ICLR}, 2024.

\bibitem[Fan et~al.(2025)Fan, Tong, Zhu, Sinha, Liu, Chen, Rabbat, Ballas, LeCun, Bar, et~al.]{fan2025scaling}
[20] Fan, D., Tong, S., Zhu, J., Sinha, K., Liu, Z., Chen, X., Rabbat, M., Ballas, N., LeCun, Y., Bar, A., et~al.
\newblock Scaling language-free visual representation learning.
\newblock In \emph{ICCV}, 2025.

\bibitem[Fu et~al.(2025)Fu, Chen, Shen, Qin, Zhang, Lin, Yang, Zheng, Li, Sun, Wu, Ji, Shan, and He]{fu2023mme}
[21] Fu, C., Chen, P., Shen, Y., Qin, Y., Zhang, M., Lin, X., Yang, J., Zheng, X., Li, K., Sun, X., Wu, Y., Ji, R., Shan, C., and He, R.
\newblock {MME}: A comprehensive evaluation benchmark for multimodal large language models.
\newblock In \emph{NeurIPS D\&B Track}, 2025.

\bibitem[Fu et~al.(2026)Fu, Huang, Wu, Jiang, Huo, Li, Song, Ding, Guo, He, Fu, Mao, and Zhang]{fu2026lanceunifiedmultimodalmodeling}
[22] Fu, F., Huang, M., Wu, S., Jiang, Y., Huo, Y., Li, H., Song, Y., Ding, F., Guo, J., He, Q., Fu, Z., Mao, Z., and Zhang, Y.
\newblock Lance: Unified multimodal modeling by multi-task synergy, 2026.
\newblock URL \url{https://arxiv.org/abs/2605.18678}.

\bibitem[Ge et~al.(2023)Ge, Ge, Zeng, Wang, and Shan]{ge2023planting}
[23] Ge, Y., Ge, Y., Zeng, Z., Wang, X., and Shan, Y.
\newblock Planting a seed of vision in large language model.
\newblock \emph{arXiv preprint arXiv:2307.08041}, 2023.

\bibitem[Ge et~al.(2024)Ge, Zhao, Zhu, Ge, Yi, Song, Li, Ding, and Shan]{ge2024seed}
[24] Ge, Y., Zhao, S., Zhu, J., Ge, Y., Yi, K., Song, L., Li, C., Ding, X., and Shan, Y.
\newblock {SEED-X:} multimodal models with unified multi-granularity comprehension and generation.
\newblock \emph{arXiv preprint arXiv:2404.14396}, 2024.

\bibitem[Gemini(2025)]{comanici2025gemini}
[25] Gemini.
\newblock Gemini 2.5: Pushing the frontier with advanced reasoning, multimodality, long context, and next generation agentic capabilities.
\newblock \emph{arXiv preprint arXiv:2507.06261}, 2025.

\bibitem[Geng et~al.(2025)Geng, Wang, Ma, Li, Rao, Gu, Zhong, Lu, Hu, Zhang, et~al.]{geng2025x}
[26] Geng, Z., Wang, Y., Ma, Y., Li, C., Rao, Y., Gu, S., Zhong, Z., Lu, Q., Hu, H., Zhang, X., et~al.
\newblock X-omni: Reinforcement learning makes discrete autoregressive image generative models great again.
\newblock \emph{arXiv preprint arXiv:2507.22058}, 2025.

\bibitem[Ghosh et~al.(2023)Ghosh, Hajishirzi, and Schmidt]{geneval}
[27] Ghosh, D., Hajishirzi, H., and Schmidt, L.
\newblock Geneval: An object-focused framework for evaluating text-to-image alignment.
\newblock \emph{Advances in Neural Information Processing Systems}, 36:\penalty0 52132--52152, 2023.

\bibitem[GPT4o(2024)]{OpenAI2024gpt4o}
[28] GPT4o.
\newblock gpt4o, 2024.

\bibitem[Gui et~al.(2025)Gui, Schusterbauer, Phan, Krause, Susskind, Bautista, and Ommer]{gui2025adapting}
[29] Gui, M., Schusterbauer, J., Phan, T., Krause, F., Susskind, J., Bautista, M.~A., and Ommer, B.
\newblock Adapting self-supervised representations as a latent space for efficient generation.
\newblock \emph{arXiv preprint arXiv:2510.14630}, 2025.

\bibitem[Han et~al.(2026{\natexlab{a}})Han, Chen, Zhao, Wang, Zhao, Yang, He, Yue, and Jiang]{han2026vision}
[30] Han, J., Chen, H., Zhao, Y., Wang, H., Zhao, Q., Yang, Z., He, H., Yue, X., and Jiang, L.
\newblock Vision as a dialect: Unifying visual understanding and generation via text-aligned representations.
\newblock \emph{Advances in Neural Information Processing Systems}, 38:\penalty0 158430--158459, 2026{\natexlab{a}}.

\bibitem[Han et~al.(2026{\natexlab{b}})Han, Tong, Fan, Ren, Sinha, Torr, and Kokkinos]{han2025learning}
[31] Han, J., Tong, S., Fan, D., Ren, Y., Sinha, K., Torr, P., and Kokkinos, F.
\newblock Learning to see before seeing: Demystifying llm visual priors from language pre-training.
\newblock In \emph{ICLR}, 2026{\natexlab{b}}.

\bibitem[Han et~al.(2025)Han, Emad, Hall, Nguyen, Padthe, Robbins, Bar, Chen, Drozdzal, Elbayad, et~al.]{han2025tv2tv}
[32] Han, X., Emad, Y., Hall, M., Nguyen, J., Padthe, K., Robbins, L., Bar, A., Chen, D., Drozdzal, M., Elbayad, M., et~al.
\newblock Tv2tv: A unified framework for interleaved language and video generation.
\newblock \emph{arXiv preprint arXiv:2512.05103}, 2025.

\bibitem[Hessel et~al.(2021)Hessel, Holtzman, Forbes, Le~Bras, and Choi]{clipscore}
[33] Hessel, J., Holtzman, A., Forbes, M., Le~Bras, R., and Choi, Y.
\newblock Clipscore: A reference-free evaluation metric for image captioning.
\newblock In \emph{Proceedings of the 2021 Conference on Empirical Methods in Natural Language Processing}, pp.\  7514--7528, 2021.

\bibitem[Heusel et~al.(2017)Heusel, Ramsauer, Unterthiner, Nessler, and Hochreiter]{fid}
[34] Heusel, M., Ramsauer, H., Unterthiner, T., Nessler, B., and Hochreiter, S.
\newblock {GANs} trained by a two time-scale update rule converge to a local nash equilibrium.
\newblock In \emph{NeurIPS}, 2017.

\bibitem[Hiippala et~al.(2021)Hiippala, Alikhani, Haverinen, Kalliokoski, Logacheva, Orekhova, Tuomainen, Stone, and Bateman]{hiippala2021ai2d}
[35] Hiippala, T., Alikhani, M., Haverinen, J., Kalliokoski, T., Logacheva, E., Orekhova, S., Tuomainen, A., Stone, M., and Bateman, J.~A.
\newblock Ai2d-rst: A multimodal corpus of 1000 primary school science diagrams.
\newblock \emph{Language Resources and Evaluation}, 55:\penalty0 661--688, 2021.

\bibitem[Hu et~al.(2025)Hu, Zhao, Chen, Qiu, Liu, Xu, Luo, Zhang, and Lu]{hu2025omni}
[36] Hu, J., Zhao, S., Chen, Q.-G., Qiu, X., Liu, J., Xu, Z., Luo, W., Zhang, K., and Lu, Y.
\newblock Omni-view: Unlocking how generation facilitates understanding in unified 3d model based on multiview images.
\newblock \emph{arXiv preprint arXiv:2511.07222}, 2025.

\bibitem[Hu et~al.(2024)Hu, Wang, Fang, Fu, Cheng, and Yu]{dpgbench}
[37] Hu, X., Wang, R., Fang, Y., Fu, B., Cheng, P., and Yu, G.
\newblock Ella: Equip diffusion models with llm for enhanced semantic alignment.
\newblock \emph{arXiv preprint arXiv:2403.05135}, 2024.

\bibitem[Hu et~al.(2026)Hu, Zhang, Luo, Guo, Chen, Sun, Feng, Lu, Chen, Zhang, et~al.]{hu2026bagelvla}
[38] Hu, Y., Zhang, J., Luo, Y., Guo, Y., Chen, X., Sun, X., Feng, K., Lu, Q., Chen, S., Zhang, Y., et~al.
\newblock Bagelvla: Enhancing long-horizon manipulation via interleaved vision-language-action generation.
\newblock \emph{arXiv preprint arXiv:2602.09849}, 2026.

\bibitem[Huang et~al.(2025)Huang, Huang, Feng, Lei, and Lv]{huang2025cross}
[39] Huang, Y., Huang, C., Feng, D., Lei, W., and Lv, J.
\newblock Cross-model transferability among large language models on the platonic representations of concepts.
\newblock In \emph{Proceedings of the 63rd Annual Meeting of the Association for Computational Linguistics (Volume 1: Long Papers)}, pp.\  3686--3704, 2025.

\bibitem[Hudson \& Manning(2019)Hudson and Manning]{hudson2019gqa}
[40] Hudson, D.~A. and Manning, C.~D.
\newblock Gqa: A new dataset for real-world visual reasoning and compositional question answering.
\newblock In \emph{CVPR}, 2019.

\bibitem[Huh et~al.(2024)Huh, Cheung, Wang, and Isola]{huh2024platonic}
[41] Huh, M., Cheung, B., Wang, T., and Isola, P.
\newblock The platonic representation hypothesis.
\newblock In \emph{ICML}, 2024.

\bibitem[Intelligence et~al.(2026)Intelligence, Ai, Amin, Aniceto, Balakrishna, Balke, Black, Bokinsky, Cao, Charbonnier, et~al.]{intelligence2026pi}
[42] Intelligence, P., Ai, B., Amin, A., Aniceto, R., Balakrishna, A., Balke, G., Black, K., Bokinsky, G., Cao, S., Charbonnier, T., et~al.
\newblock Pi0.7: a steerable generalist robotic foundation model with emergent capabilities.
\newblock \emph{arXiv preprint arXiv:2604.15483}, 2026.

\bibitem[Jia et~al.(2025)Jia, Li, Shu, Zheng, Fan, Guo, Shi, Lu, Li, Guo, et~al.]{jia2025dino}
[43] Jia, M., Li, M., Shu, Z., Zheng, A., Fan, L., Guo, J., Shi, T., Lu, D., Li, Z., Guo, X., et~al.
\newblock Dino-tok: Adapting dino for visual tokenizers.
\newblock \emph{arXiv preprint arXiv:2511.20565}, 2025.

\bibitem[Jiao et~al.(2025)Jiao, Qiu, Jie, Chen, Chen, Ma, and Jiang]{jiao2025unitoken}
[44] Jiao, Y., Qiu, H., Jie, Z., Chen, S., Chen, J., Ma, L., and Jiang, Y.-G.
\newblock Unitoken: Harmonizing multimodal understanding and generation through unified visual encoding.
\newblock In \emph{CVPR}, 2025.

\bibitem[Jin et~al.(2026)Jin, Zhou, Yang, Zhang, Liu, Zhu, and Deng]{jin2026latentum}
[45] Jin, J., Zhou, Z., Yang, X., Zhang, H., Liu, P., Zhu, J., and Deng, Z.
\newblock Latentum: Unleashing the potential of interleaved cross-modal reasoning via a latent-space unified model.
\newblock \emph{arXiv preprint arXiv:2604.02097}, 2026.

\bibitem[Johnson et~al.(2017)Johnson, Hariharan, Van Der~Maaten, Fei-Fei, Lawrence~Zitnick, and Girshick]{johnson2017clevr}
[46] Johnson, J., Hariharan, B., Van Der~Maaten, L., Fei-Fei, L., Lawrence~Zitnick, C., and Girshick, R.
\newblock Clevr: A diagnostic dataset for compositional language and elementary visual reasoning.
\newblock In \emph{CVPR}, 2017.

\bibitem[Joshi et~al.(2017)Joshi, Choi, Weld, and Zettlemoyer]{joshi2017triviaqa}
[47] Joshi, M., Choi, E., Weld, D.~S., and Zettlemoyer, L.
\newblock Triviaqa: A large scale distantly supervised challenge dataset for reading comprehension.
\newblock In \emph{Proceedings of the 55th Annual Meeting of the Association for Computational Linguistics (Volume 1: Long Papers)}, pp.\  1601--1611, 2017.

\bibitem[K2.5(2026)]{team2026kimi}
[48] K2.5.
\newblock Kimi k2.5: Visual agentic intelligence.
\newblock \emph{arXiv preprint arXiv:2602.02276}, 2026.

\bibitem[Kim et~al.(2026)Kim, Kang, Choi, Ji, Woo, Chung, Han, and Han]{kim2026physics}
[49] Kim, H., Kang, S., Choi, Y., Ji, S., Woo, J., Chung, H., Han, S.~C., and Han, K.
\newblock Physics-based phenomenological characterization of cross-modal bias in multimodal models.
\newblock \emph{arXiv preprint arXiv:2602.20624}, 2026.

\bibitem[Kwiatkowski et~al.(2019)Kwiatkowski, Palomaki, Redfield, Collins, Parikh, Alberti, Epstein, Polosukhin, Devlin, Lee, et~al.]{kwiatkowski2019natural}
[50] Kwiatkowski, T., Palomaki, J., Redfield, O., Collins, M., Parikh, A., Alberti, C., Epstein, D., Polosukhin, I., Devlin, J., Lee, K., et~al.
\newblock Natural questions: a benchmark for question answering research.
\newblock \emph{Transactions of the Association for Computational Linguistics}, 7:\penalty0 453--466, 2019.

\bibitem[Lee et~al.(2022)Lee, Kim, Kim, Cho, and Han]{rqvae}
[51] Lee, D., Kim, C., Kim, S., Cho, M., and Han, W.-S.
\newblock Autoregressive image generation using residual quantization.
\newblock In \emph{CVPR}, 2022.

\bibitem[Li et~al.(2026{\natexlab{a}})Li, Yin, Chai, Fu, and Liu]{li2026ueval}
[52] Li, B., Yin, Y., Chai, W., Fu, X., and Liu, Z.
\newblock Ueval: A benchmark for unified multimodal generation.
\newblock \emph{arXiv preprint arXiv:2601.22155}, 2026{\natexlab{a}}.

\bibitem[Li et~al.(2024{\natexlab{a}})Li, Liu, Wu, Wang, Shen, Qu, Niu, Zhou, Huang, Li, et~al.]{li2024aria}
[53] Li, D., Liu, Y., Wu, H., Wang, Y., Shen, Z., Qu, B., Niu, X., Zhou, F., Huang, C., Li, Y., et~al.
\newblock Aria: An open multimodal native mixture-of-experts model.
\newblock \emph{arXiv preprint arXiv:2410.05993}, 2024{\natexlab{a}}.

\bibitem[Li et~al.(2025{\natexlab{a}})Li, Peng, Wang, Peng, Chen, Weng, Wang, Cai, Dai, and Xiong]{li2025onecat}
[54] Li, H., Peng, X., Wang, Y., Peng, Z., Chen, X., Weng, R., Wang, J., Cai, X., Dai, W., and Xiong, H.
\newblock Onecat: Decoder-only auto-regressive model for unified understanding and generation.
\newblock \emph{arXiv preprint arXiv:2509.03498}, 2025{\natexlab{a}}.

\bibitem[Li et~al.(2025{\natexlab{b}})Li, Tian, Shao, Zhu, Wang, Zhu, Dou, Wang, Li, Lu, et~al.]{li2025synergen}
[55] Li, H., Tian, C., Shao, J., Zhu, X., Wang, Z., Zhu, J., Dou, W., Wang, X., Li, H., Lu, L., et~al.
\newblock Synergen-vl: Towards synergistic image understanding and generation with vision experts and token folding.
\newblock In \emph{CVPR}, 2025{\natexlab{b}}.

\bibitem[Li et~al.(2026{\natexlab{b}})Li, Chen, Zhu, Huang, Cai, Jiang, Hu, and Chen]{li2026sparsemanticpixelselfalignmentadaptive}
[56] Li, H., Chen, H., Zhu, C., Huang, X., Cai, J., Jiang, X., Hu, Y., and Chen, L.
\newblock Spar: Semantic-pixel self-alignment and adaptive routing for unified multimodal models, 2026{\natexlab{b}}.
\newblock URL \url{https://arxiv.org/abs/2606.23041}.

\bibitem[Li et~al.(2023)Li, Li, Savarese, and Hoi]{li2023blip}
[57] Li, J., Li, D., Savarese, S., and Hoi, S.
\newblock Blip-2: Bootstrapping language-image pre-training with frozen image encoders and large language models.
\newblock In \emph{ICML}, 2023.

\bibitem[Li et~al.(2024{\natexlab{b}})Li, Fang, Smyrnis, Ivgi, Jordan, Gadre, Bansal, Guha, Keh, Arora, Garg, Xin, Muennighoff, Heckel, Mercat, Chen, Gururangan, Wortsman, Albalak, Bitton, Nezhurina, Abbas, Hsieh, Ghosh, Gardner, Kilian, Zhang, Shao, Pratt, Sanyal, Ilharco, Daras, Marathe, Gokaslan, Zhang, Chandu, Nguyen, Vasiljevic, Kakade, Song, Sanghavi, Faghri, Oh, Zettlemoyer, Lo, El-Nouby, Pouransari, Toshev, Wang, Groeneveld, Soldaini, Koh, Jitsev, Kollar, Dimakis, Carmon, Dave, Schmidt, and Shankar]{li2024datacomplm}
[58] Li, J., Fang, A., Smyrnis, G., Ivgi, M., Jordan, M., Gadre, S., Bansal, H., Guha, E., Keh, S., Arora, K., Garg, S., Xin, R., Muennighoff, N., Heckel, R., Mercat, J., Chen, M., Gururangan, S., Wortsman, M., Albalak, A., Bitton, Y., Nezhurina, M., Abbas, A., Hsieh, C.-Y., Ghosh, D., Gardner, J., Kilian, M., Zhang, H., Shao, R., Pratt, S., Sanyal, S., Ilharco, G., Daras, G., Marathe, K., Gokaslan, A., Zhang, J., Chandu, K., Nguyen, T., Vasiljevic, I., Kakade, S., Song, S., Sanghavi, S., Faghri, F., Oh, S., Zettlemoyer, L., Lo, K., El-Nouby, A., Pouransari, H., Toshev, A., Wang, S., Groeneveld, D., Soldaini, L., Koh, P.~W., Jitsev, J., Kollar, T., Dimakis, A.~G., Carmon, Y., Dave, A., Schmidt, L., and Shankar, V.
\newblock Datacomp-lm: In search of the next generation of training sets for language models, 2024{\natexlab{b}}.

\bibitem[Li \& He(2025)Li and He]{li2025jit}
[59] Li, T. and He, K.
\newblock Back to basics: Let denoising generative models denoise.
\newblock \emph{arXiv preprint arXiv:2511.13720}, 2025.

\bibitem[Li et~al.(2025{\natexlab{c}})Li, Qian, Pan, Zhang, Huang, Zhang, Tong, You, Du, Gan, Kim, Jia, Wang, Yang, Gao, Dou, Hu, Gao, Li, Dufter, Wang, Yin, Zhang, Chen, Zhao, Pang, and Chen]{li2025manzanosimplescalableunified}
[60] Li, Y., Qian, R., Pan, B., Zhang, H., Huang, H., Zhang, B., Tong, J., You, H., Du, X., Gan, Z., Kim, H., Jia, C., Wang, Z., Yang, Y., Gao, M., Dou, Z.-Y., Hu, W., Gao, C., Li, D., Dufter, P., Wang, Z., Yin, G., Zhang, Z., Chen, C., Zhao, Y., Pang, R., and Chen, Z.
\newblock Manzano: A simple and scalable unified multimodal model with a hybrid vision tokenizer, 2025{\natexlab{c}}.

\bibitem[Li et~al.(2025{\natexlab{d}})Li, Liu, Zhang, Lin, Yuan, Yan, Ye, Yu, Niu, and Yuan]{li2025uniworld}
[61] Li, Z., Liu, Z., Zhang, Q., Lin, B., Yuan, S., Yan, Z., Ye, Y., Yu, W., Niu, Y., and Yuan, L.
\newblock Uniworld-v2: Reinforce image editing with diffusion negative-aware finetuning and mllm implicit feedback.
\newblock \emph{arXiv preprint arXiv:2510.16888}, 2025{\natexlab{d}}.

\bibitem[Liang et~al.(2024)Liang, Yu, Luo, Iyer, Dong, Zhou, Ghosh, Lewis, Yih, Zettlemoyer, et~al.]{liang2024mixture}
[62] Liang, W., Yu, L., Luo, L., Iyer, S., Dong, N., Zhou, C., Ghosh, G., Lewis, M., Yih, W.-t., Zettlemoyer, L., et~al.
\newblock Mixture-of-transformers: A sparse and scalable architecture for multi-modal foundation models.
\newblock \emph{arXiv preprint arXiv:2411.04996}, 2024.

\bibitem[Liao et~al.(2025)Liao, Liu, Wang, Luo, Zhang, Zhao, Wu, Li, Tian, and Huang]{mogao}
[63] Liao, C., Liu, L., Wang, X., Luo, Z., Zhang, X., Zhao, W., Wu, J., Li, L., Tian, Z., and Huang, W.
\newblock Mogao: An omni foundation model for interleaved multi-modal generation, 2025.

\bibitem[Lin et~al.(2026)Lin, Liu, Lin, Chen, Ge, Lin, Zhang, Yang, Zhong, Bo, and Yuan]{lin2026gearguidedendtoendautoregression}
[64] Lin, B., Liu, Z., Lin, C., Chen, S., Ge, Y., Lin, Y., Zhang, J., Yang, M., Zhong, Z., Bo, L., and Yuan, L.
\newblock Gear: Guided end-to-end autoregression for image synthesis, 2026.
\newblock URL \url{https://arxiv.org/abs/2606.32039}.

\bibitem[Lin et~al.(2025)Lin, Pan, Huang, Hou, Wang, Chen, He, Juefei-Xu, Sun, Fan, et~al.]{lin2025exploring}
[65] Lin, H., Pan, X., Huang, Z., Hou, J., Wang, J., Chen, W., He, Z., Juefei-Xu, F., Sun, J., Fan, Z., et~al.
\newblock Exploring mllm-diffusion information transfer with metacanvas.
\newblock \emph{arXiv preprint arXiv:2512.11464}, 2025.

\bibitem[Lin et~al.(2014)Lin, Maire, Belongie, Hays, Perona, Ramanan, Doll{\'a}r, and Zitnick]{lin2014microsoft}
[66] Lin, T.-Y., Maire, M., Belongie, S., Hays, J., Perona, P., Ramanan, D., Doll{\'a}r, P., and Zitnick, C.~L.
\newblock Microsoft coco: Common objects in context.
\newblock In \emph{Computer vision--ECCV 2014: 13th European conference, zurich, Switzerland, September 6-12, 2014, proceedings, part v 13}, pp.\  740--755. Springer, 2014.

\bibitem[Lin et~al.(2024)Lin, Shrivastava, Luo, Iyer, Lewis, Ghosh, Zettlemoyer, and Aghajanyan]{lin2024moma}
[67] Lin, X.~V., Shrivastava, A., Luo, L., Iyer, S., Lewis, M., Ghosh, G., Zettlemoyer, L., and Aghajanyan, A.
\newblock Moma: Efficient early-fusion pre-training with mixture of modality-aware experts.
\newblock \emph{arXiv preprint arXiv:2407.21770}, 2024.

\bibitem[Liu et~al.(2023{\natexlab{a}})Liu, Li, Wu, and Lee]{liu2023visual}
[68] Liu, H., Li, C., Wu, Q., and Lee, Y.~J.
\newblock Visual instruction tuning.
\newblock In \emph{NeurIPS}, 2023{\natexlab{a}}.

\bibitem[Liu et~al.(2023{\natexlab{b}})Liu, Li, Li, Yu, Huang, Peng, Liu, Chen, Li, Jin, et~al.]{liu2023hidden}
[69] Liu, Y., Li, Z., Li, H., Yu, W., Huang, M., Peng, D., Liu, M., Chen, M., Li, C., Jin, L., et~al.
\newblock On the hidden mystery of ocr in large multimodal models.
\newblock \emph{arXiv preprint arXiv:2305.07895}, 2023{\natexlab{b}}.

\bibitem[Liu et~al.(2024)Liu, Duan, Zhang, Li, Zhang, Zhao, Yuan, Wang, He, Liu, et~al.]{liu2023mmbench}
[70] Liu, Y., Duan, H., Zhang, Y., Li, B., Zhang, S., Zhao, W., Yuan, Y., Wang, J., He, C., Liu, Z., et~al.
\newblock Mmbench: Is your multi-modal model an all-around player?
\newblock In \emph{ECCV}, 2024.

\bibitem[Liu et~al.(2025)Liu, Ren, Liu, Zhou, Chen, Qiu, Huang, An, Yang, Patel, et~al.]{liu2025tuna}
[71] Liu, Z., Ren, W., Liu, H., Zhou, Z., Chen, S., Qiu, H., Huang, X., An, Z., Yang, F., Patel, A., et~al.
\newblock Tuna: Taming unified visual representations for native unified multimodal models.
\newblock \emph{arXiv preprint arXiv:2512.02014}, 2025.

\bibitem[Liu et~al.(2026)Liu, Ren, Huang, Chen, Li, Chen, Ji, He, Schult, Zeng, et~al.]{liu2026tuna}
[72] Liu, Z., Ren, W., Huang, X., Chen, S., Li, T., Chen, M., Ji, Y., He, S., Schult, J., Zeng, B., et~al.
\newblock Tuna-2: Pixel embeddings beat vision encoders for multimodal understanding and generation.
\newblock \emph{arXiv preprint arXiv:2604.24763}, 2026.

\bibitem[Llama3(2024)]{grattafiori2024llama3}
[73] Llama3.
\newblock The llama 3 herd of models.
\newblock \emph{arXiv preprint arXiv:2407.21783}, 2024.

\bibitem[Llama4(2025)]{meta2025llama}
[74] Llama4.
\newblock The llama 4 herd: The beginning of a new era of natively multimodal ai innovation, 2025.

\bibitem[Lu et~al.(2022{\natexlab{a}})Lu, Clark, Zellers, Mottaghi, and Kembhavi]{lu2022unified}
[75] Lu, J., Clark, C., Zellers, R., Mottaghi, R., and Kembhavi, A.
\newblock Unified-io: A unified model for vision, language, and multi-modal tasks.
\newblock In \emph{ICLR}, 2022{\natexlab{a}}.

\bibitem[Lu et~al.(2024)Lu, Clark, Lee, Zhang, Khosla, Marten, Hoiem, and Kembhavi]{lu2024unified}
[76] Lu, J., Clark, C., Lee, S., Zhang, Z., Khosla, S., Marten, R., Hoiem, D., and Kembhavi, A.
\newblock Unified-io 2: Scaling autoregressive multimodal models with vision language audio and action.
\newblock In \emph{CVPR}, 2024.

\bibitem[Lu et~al.(2025)Lu, Song, Xu, Ahn, Wang, Chen, Dehghan, and Yang]{atoken}
[77] Lu, J., Song, L., Xu, M., Ahn, B., Wang, Y., Chen, C., Dehghan, A., and Yang, Y.
\newblock Atoken: A unified tokenizer for vision, 2025.
\newblock URL \url{https://arxiv.org/abs/2509.14476}.

\bibitem[Lu et~al.(2022{\natexlab{b}})Lu, Mishra, Xia, Qiu, Chang, Zhu, Tafjord, Clark, and Kalyan]{lu2022learn}
[78] Lu, P., Mishra, S., Xia, T., Qiu, L., Chang, K.-W., Zhu, S.-C., Tafjord, O., Clark, P., and Kalyan, A.
\newblock Learn to explain: Multimodal reasoning via thought chains for science question answering.
\newblock In \emph{NeurIPS}, 2022{\natexlab{b}}.

\bibitem[Lu et~al.(2023)Lu, Bansal, Xia, Liu, Li, Hajishirzi, Cheng, Chang, Galley, and Gao]{lu2023mathvista}
[79] Lu, P., Bansal, H., Xia, T., Liu, J., Li, C., Hajishirzi, H., Cheng, H., Chang, K.-W., Galley, M., and Gao, J.
\newblock Mathvista: Evaluating mathematical reasoning of foundation models in visual contexts.
\newblock In \emph{ICLR}, 2023.

\bibitem[Luo et~al.(2024)Luo, Cao, Lee, Johnson, and Lee]{luo2024probing}
[80] Luo, T., Cao, A., Lee, G., Johnson, J., and Lee, H.
\newblock Probing visual language priors in vlms.
\newblock \emph{arXiv preprint arXiv:2501.00569}, 2024.

\bibitem[Ma et~al.(2025)Ma, Liu, Chen, Liu, Wu, Wu, Pan, Xie, Zhang, Yu, et~al.]{ma2025janusflow}
[81] Ma, Y., Liu, X., Chen, X., Liu, W., Wu, C., Wu, Z., Pan, Z., Xie, Z., Zhang, H., Yu, X., et~al.
\newblock Janusflow: Harmonizing autoregression and rectified flow for unified multimodal understanding and generation.
\newblock In \emph{CVPR}, 2025.

\bibitem[Masry et~al.(2022)Masry, Long, Tan, Joty, and Hoque]{masry2022chartqa}
[82] Masry, A., Long, D.~X., Tan, J.~Q., Joty, S., and Hoque, E.
\newblock Chartqa: A benchmark for question answering about charts with visual and logical reasoning.
\newblock In \emph{ACL}, 2022.

\bibitem[Mathew et~al.(2021)Mathew, Karatzas, and Jawahar]{mathew2021docvqa}
[83] Mathew, M., Karatzas, D., and Jawahar, C.
\newblock Docvqa: A dataset for vqa on document images.
\newblock In \emph{WACV}, 2021.

\bibitem[Mihaylov et~al.(2018)Mihaylov, Clark, Khot, and Sabharwal]{openbookqa}
[84] Mihaylov, T., Clark, P., Khot, T., and Sabharwal, A.
\newblock Can a suit of armor conduct electricity? a new dataset for open book question answering.
\newblock \emph{arXiv preprint arXiv:1809.02789}, 2018.

\bibitem[Nguyen et~al.(2025)Nguyen, Havasi, Berrada, Zettlemoyer, and Chen]{nguyen2025oneflow}
[85] Nguyen, J., Havasi, M., Berrada, T., Zettlemoyer, L., and Chen, R.~T.
\newblock Oneflow: Concurrent mixed-modal and interleaved generation with edit flows.
\newblock \emph{arXiv preprint arXiv:2510.03506}, 2025.

\bibitem[Niu et~al.(2025{\natexlab{a}})Niu, Jin, Liao, Feng, Jin, Lin, Li, Zhu, Yu, and Yuan]{niu2025does}
[86] Niu, Y., Jin, W., Liao, J., Feng, C., Jin, P., Lin, B., Li, Z., Zhu, B., Yu, W., and Yuan, L.
\newblock Does understanding inform generation in unified multimodal models? from analysis to path forward.
\newblock \emph{arXiv preprint arXiv:2511.20561}, 2025{\natexlab{a}}.

\bibitem[Niu et~al.(2025{\natexlab{b}})Niu, Ning, Zheng, Jin, Lin, Jin, Liao, Feng, Ning, Zhu, et~al.]{wise}
[87] Niu, Y., Ning, M., Zheng, M., Jin, W., Lin, B., Jin, P., Liao, J., Feng, C., Ning, K., Zhu, B., et~al.
\newblock Wise: A world knowledge-informed semantic evaluation for text-to-image generation.
\newblock \emph{arXiv preprint arXiv:2503.07265}, 2025{\natexlab{b}}.

\bibitem[Oord et~al.(2017)Oord, Vinyals, and Kavukcuoglu]{vqvae}
[88] Oord, A. v.~d., Vinyals, O., and Kavukcuoglu, K.
\newblock Neural discrete representation learning.
\newblock In \emph{NeurIPS}, 2017.

\bibitem[Orhan \& Lake(2024)Orhan and Lake]{orhan2024learning}
[89] Orhan, A.~E. and Lake, B.~M.
\newblock Learning high-level visual representations from a child’s perspective without strong inductive biases.
\newblock \emph{Nature Machine Intelligence}, 6\penalty0 (3):\penalty0 271--283, 2024.

\bibitem[Pan et~al.(2025)Pan, Shukla, Singh, Zhao, Mishra, Wang, Xu, Chen, Li, Juefei-Xu, et~al.]{metaquery}
[90] Pan, X., Shukla, S.~N., Singh, A., Zhao, Z., Mishra, S.~K., Wang, J., Xu, Z., Chen, J., Li, K., Juefei-Xu, F., et~al.
\newblock Transfer between modalities with metaqueries.
\newblock \emph{arXiv preprint arXiv:2504.06256}, 2025.

\bibitem[Pan et~al.(2026)Pan, Singh, Shukla, Fan, Mishra, and Xie]{pan2026repfusionleveragingmultimodalpriors}
[91] Pan, X., Singh, A., Shukla, S.~N., Fan, X., Mishra, S.~K., and Xie, S.
\newblock Repfusion: Leveraging multimodal priors for denoising in representation space, 2026.
\newblock URL \url{https://arxiv.org/abs/2606.14700}.

\bibitem[Peng et~al.(2026)Peng, Meng, Cai, Zhuang, Yang, Fang, Wu, Lin, Wu, and Bai]{peng2026uniar}
[92] Peng, W., Meng, L., Cai, Y., Zhuang, X., Yang, Y., Fang, R., Wu, C., Lin, J., Wu, Z., and Bai, S.
\newblock Unified multimodal autoregressive modeling with shared context --- visual tokenizer is key to unification.
\newblock In \emph{ICML}, 2026.

\bibitem[Qwen3-vl(2025)]{bai2025qwen3}
[93] Qwen3-vl.
\newblock Qwen3-vl technical report.
\newblock \emph{arXiv preprint arXiv:2511.21631}, 2025.

\bibitem[Qwen3.5-omni(2026)]{team2026qwen3}
[94] Qwen3.5-omni.
\newblock Qwen3.5-omni technical report.
\newblock \emph{arXiv preprint arXiv:2604.15804}, 2026.

\bibitem[Razavi et~al.(2019)Razavi, van~den Oord, and Vinyals]{vqvae2}
[95] Razavi, A., van~den Oord, A., and Vinyals, O.
\newblock Generating diverse high-fidelity images with vq-vae-2.
\newblock In \emph{NeurIPS}, 2019.

\bibitem[Reddy et~al.(2019)Reddy, Chen, and Manning]{reddy2019coqa}
[96] Reddy, S., Chen, D., and Manning, C.~D.
\newblock Coqa: A conversational question answering challenge.
\newblock \emph{Transactions of the Association for Computational Linguistics}, 7:\penalty0 249--266, 2019.

\bibitem[Sakaguchi et~al.(2021)Sakaguchi, Bras, Bhagavatula, and Choi]{sakaguchi2021winogrande}
[97] Sakaguchi, K., Bras, R.~L., Bhagavatula, C., and Choi, Y.
\newblock Winogrande: An adversarial winograd schema challenge at scale.
\newblock \emph{Communications of the ACM}, 64\penalty0 (9):\penalty0 99--106, 2021.

\bibitem[Sap et~al.(2019)Sap, Rashkin, Chen, Le~Bras, and Choi]{sap2019social}
[98] Sap, M., Rashkin, H., Chen, D., Le~Bras, R., and Choi, Y.
\newblock Social iqa: Commonsense reasoning about social interactions.
\newblock In \emph{Proceedings of the 2019 conference on empirical methods in natural language processing and the 9th international joint conference on natural language processing (EMNLP-IJCNLP)}, pp.\  4463--4473, 2019.

\bibitem[Schlarmann et~al.(2025)Schlarmann, Croce, Flammarion, and Hein]{schlarmann2025fuselip}
[99] Schlarmann, C., Croce, F., Flammarion, N., and Hein, M.
\newblock Fuselip: Multimodal embeddings via early fusion of discrete tokens.
\newblock \emph{arXiv preprint arXiv:2506.03096}, 2025.

\bibitem[Shi et~al.(2024)Shi, Han, Zhou, Liang, Lin, Zettlemoyer, and Yu]{lmfusion}
[100] Shi, W., Han, X., Zhou, C., Liang, W., Lin, X.~V., Zettlemoyer, L., and Yu, L.
\newblock Lmfusion: Adapting pretrained language models for multimodal generation.
\newblock \emph{arXiv preprint arXiv:2412.15188}, 2024.

\bibitem[Shi et~al.(2026)Shi, Dong, Ding, Wang, Zhu, Zhou, Liu, Tian, Wang, Wang, et~al.]{shi2026realunify}
[101] Shi, Y., Dong, Y., Ding, Y., Wang, Y., Zhu, X., Zhou, S., Liu, W., Tian, H., Wang, R., Wang, H., et~al.
\newblock Realunify: Do unified models truly benefit from unification? a comprehensive benchmark.
\newblock In \emph{Proceedings of the IEEE/CVF Conference on Computer Vision and Pattern Recognition}, pp.\  22488--22497, 2026.

\bibitem[Shukor et~al.(2025)Shukor, Fini, da~Costa, Cord, Susskind, and El-Nouby]{shukor2025scaling}
[102] Shukor, M., Fini, E., da~Costa, V. G.~T., Cord, M., Susskind, J., and El-Nouby, A.
\newblock Scaling laws for native multimodal models.
\newblock In \emph{ICCV}, 2025.

\bibitem[Singh et~al.(2019)Singh, Natarajan, Shah, Jiang, Chen, Batra, Parikh, and Rohrbach]{singh2019towards}
[103] Singh, A., Natarajan, V., Shah, M., Jiang, Y., Chen, X., Batra, D., Parikh, D., and Rohrbach, M.
\newblock Towards vqa models that can read.
\newblock In \emph{CVPR}, 2019.

\bibitem[Singh et~al.(2026)Singh, Zheng, Wu, Zhang, Shechtman, and Xie]{singh2026raev2}
[104] Singh, J., Zheng, B., Wu, Z., Zhang, R., Shechtman, E., and Xie, S.
\newblock Improved baselines with representation autoencoders.
\newblock \emph{arXiv preprint arXiv:2605.18324}, 2026.

\bibitem[Smith \& Gasser(2005)Smith and Gasser]{smith2005development}
[105] Smith, L. and Gasser, M.
\newblock The development of embodied cognition: Six lessons from babies.
\newblock \emph{Artificial life}, 11\penalty0 (1-2):\penalty0 13--29, 2005.

\bibitem[Steinberg \& Steinberg(1975)Steinberg and Steinberg]{steinberg1975reading}
[106] Steinberg, D.~D. and Steinberg, M.~T.
\newblock Reading before speaking.
\newblock \emph{Visible Language}, 9\penalty0 (3), 1975.

\bibitem[Su et~al.(2026)Su, Wei, Cen, Wang, Chen, Yuan, and Chu]{su2026generation}
[107] Su, Z., Wei, H., Cen, K., Wang, Y., Chen, G., Yuan, C., and Chu, X.
\newblock Generation enhances understanding in unified multimodal models via multi-representation generation.
\newblock \emph{arXiv preprint arXiv:2601.21406}, 2026.

\bibitem[Sun et~al.(2024)Sun, Cui, Zhang, Zhang, Yu, Luo, Wang, Rao, Liu, Huang, and Wang]{emu2}
[108] Sun, Q., Cui, Y., Zhang, X., Zhang, F., Yu, Q., Luo, Z., Wang, Y., Rao, Y., Liu, J., Huang, T., and Wang, X.
\newblock Generative multimodal models are in-context learners.
\newblock In \emph{CVPR}, 2024.

\bibitem[Team(2026{\natexlab{a}})]{aditi2026cosmos3omnimodalworld}
[109] Team, C.~.
\newblock Cosmos 3: Omnimodal world models for physical ai, 2026{\natexlab{a}}.
\newblock URL \url{https://arxiv.org/abs/2606.02800}.

\bibitem[Team(2024)]{team2024chameleon}
[110] Team, C.
\newblock Chameleon: Mixed-modal early-fusion foundation models.
\newblock \emph{arXiv preprint arXiv:2405.09818}, 2024.

\bibitem[Team(2026{\natexlab{b}})]{meituanlongcatteam2026longcatnextlexicalizingmodalitiesdiscrete}
[111] Team, M.~L.
\newblock Longcat-next: Lexicalizing modalities as discrete tokens, 2026{\natexlab{b}}.
\newblock URL \url{https://arxiv.org/abs/2603.27538}.

\bibitem[TML(2026)]{thinkingmachines2026interactionmodels}
[112] TML.
\newblock Interaction models: A scalable approach to human-ai collaboration.
\newblock \emph{Thinking Machines Lab}, May 2026.
\newblock \doi{10.64434/tml.20260511}.
\newblock https://thinkingmachines.ai/blog/interaction-models/.

\bibitem[Tong et~al.(2024{\natexlab{a}})Tong, Brown, Wu, Woo, Middepogu, Akula, Yang, Yang, Iyer, Pan, et~al.]{tong2024cambrian}
[113] Tong, S., Brown, E., Wu, P., Woo, S., Middepogu, M., Akula, S.~C., Yang, J., Yang, S., Iyer, A., Pan, X., et~al.
\newblock Cambrian-1: A fully open, vision-centric exploration of multimodal llms.
\newblock In \emph{NeurIPS}, 2024{\natexlab{a}}.

\bibitem[Tong et~al.(2024{\natexlab{b}})Tong, Liu, Zhai, Ma, LeCun, and Xie]{tong2024eyes}
[114] Tong, S., Liu, Z., Zhai, Y., Ma, Y., LeCun, Y., and Xie, S.
\newblock Eyes wide shut? exploring the visual shortcomings of multimodal llms.
\newblock In \emph{CVPR}, 2024{\natexlab{b}}.

\bibitem[Tong et~al.(2025)Tong, Fan, Zhu, Xiong, Chen, Sinha, Rabbat, LeCun, Xie, and Liu]{tong2024metamorph}
[115] Tong, S., Fan, D., Zhu, J., Xiong, Y., Chen, X., Sinha, K., Rabbat, M., LeCun, Y., Xie, S., and Liu, Z.
\newblock Metamorph: Multimodal understanding and generation via instruction tuning.
\newblock In \emph{ICCV}, 2025.

\bibitem[Tong et~al.(2026{\natexlab{a}})Tong, Fan, Nguyen, Brown, Zhou, Qian, Zheng, Vallaeys, Han, Fergus, Murray, Ghazvininejad, Lewis, Ballas, Bar, Rabbat, Verbeek, Zettlemoyer, Sinha, LeCun, and Xie]{tong2026beyond}
[116] Tong, S., Fan, D., Nguyen, J., Brown, E., Zhou, G., Qian, S., Zheng, B., Vallaeys, T., Han, J., Fergus, R., Murray, N., Ghazvininejad, M., Lewis, M., Ballas, N., Bar, A., Rabbat, M., Verbeek, J., Zettlemoyer, L., Sinha, K., LeCun, Y., and Xie, S.
\newblock Beyond language modeling: An exploration of multimodal pretraining.
\newblock \emph{ICML}, 2026{\natexlab{a}}.

\bibitem[Tong et~al.(2026{\natexlab{b}})Tong, Zheng, Wang, Tang, Ma, Brown, Yang, Fergus, LeCun, and Xie]{scale-rae-2026}
[117] Tong, S., Zheng, B., Wang, Z., Tang, B., Ma, N., Brown, E., Yang, J., Fergus, R., LeCun, Y., and Xie, S.
\newblock Scaling text-to-image diffusion transformers with representation autoencoders.
\newblock \emph{arXiv preprint}, 2026{\natexlab{b}}.

\bibitem[Tong et~al.(2026{\natexlab{c}})Tong, Chang, Yin, Liu, Fang, and Ma]{tong2026reversing}
[118] Tong, Y., Chang, D., Yin, Z., Liu, X., Fang, Y., and Ma, Z.
\newblock Reversing the flow: Generation-to-understanding synergy in large multimodal models.
\newblock In \emph{Proceedings of the IEEE/CVF Conference on Computer Vision and Pattern Recognition}, pp.\  6976--6986, 2026{\natexlab{c}}.

\bibitem[Touvron et~al.(2023)Touvron, Lavril, Izacard, Martinet, Lachaux, Lacroix, Rozi{\`e}re, Goyal, Hambro, Azhar, et~al.]{touvron2023llama}
[119] Touvron, H., Lavril, T., Izacard, G., Martinet, X., Lachaux, M.-A., Lacroix, T., Rozi{\`e}re, B., Goyal, N., Hambro, E., Azhar, F., et~al.
\newblock {LLaMA}: Open and efficient foundation language models.
\newblock \emph{arXiv preprint arXiv:2302.13971}, 2023.

\bibitem[Tschannen et~al.(2025)Tschannen, Gritsenko, Wang, Naeem, Alabdulmohsin, Parthasarathy, Evans, Beyer, Xia, Mustafa, et~al.]{tschannen2025siglip}
[120] Tschannen, M., Gritsenko, A., Wang, X., Naeem, M.~F., Alabdulmohsin, I., Parthasarathy, N., Evans, T., Beyer, L., Xia, Y., Mustafa, B., et~al.
\newblock Siglip 2: Multilingual vision-language encoders with improved semantic understanding, localization, and dense features.
\newblock \emph{arXiv preprint arXiv:2502.14786}, 2025.

\bibitem[Vong et~al.(2024)Vong, Wang, Orhan, and Lake]{vong2024grounded}
[121] Vong, W.~K., Wang, W., Orhan, A.~E., and Lake, B.~M.
\newblock Grounded language acquisition through the eyes and ears of a single child.
\newblock \emph{Science}, 383\penalty0 (6682):\penalty0 504--511, 2024.

\bibitem[Wang et~al.(2026{\natexlab{a}})Wang, Chen, Hu, Chen, Chen, Wiegreffe, and Zhou]{wang2026quantifying}
[122] Wang, C., Chen, Y., Hu, Z., Chen, D., Chen, W., Wiegreffe, S., and Zhou, T.
\newblock Quantifying the gap between understanding and generation within unified multimodal models.
\newblock \emph{arXiv preprint arXiv:2602.02140}, 2026{\natexlab{a}}.

\bibitem[Wang et~al.(2025{\natexlab{a}})Wang, Zhao, Zhang, Cao, Zhan, Duan, Lu, Fu, Chen, Zhao, et~al.]{wang2025ovis}
[123] Wang, G.-H., Zhao, S., Zhang, X., Cao, L., Zhan, P., Duan, L., Lu, S., Fu, M., Chen, X., Zhao, J., et~al.
\newblock Ovis-u1 technical report.
\newblock \emph{arXiv preprint arXiv:2506.23044}, 2025{\natexlab{a}}.

\bibitem[Wang et~al.(2026{\natexlab{b}})Wang, Wu, Wu, Sun, Liu, Yu, Ma, He, He, Hong, et~al.]{wang2026ernie}
[124] Wang, H., Wu, H., Wu, T., Sun, Y., Liu, J., Yu, D., Ma, Y., He, J., He, Z., Hong, D., et~al.
\newblock Ernie 5.0 technical report.
\newblock \emph{arXiv preprint arXiv:2602.04705}, 2026{\natexlab{b}}.

\bibitem[Wang et~al.(2026{\natexlab{c}})Wang, Wang, Pan, Hu, Li, Sun, Deng, Chen, Chen, Tian, et~al.]{wang2026arm}
[125] Wang, J., Wang, X., Pan, J., Hu, X., Li, F., Sun, J., Deng, C., Chen, Z., Chen, Y., Tian, K., et~al.
\newblock Arm: An autoregressive large multimodal model with unified discrete representations.
\newblock \emph{arXiv preprint arXiv:2606.11188}, 2026{\natexlab{c}}.

\bibitem[Wang et~al.(2024{\natexlab{a}})Wang, Zhao, Zhang, Feng, Liu, and Kang]{wang2024image}
[126] Wang, L., Zhao, Y., Zhang, Z., Feng, J., Liu, S., and Kang, B.
\newblock Image understanding makes for a good tokenizer for image generation.
\newblock \emph{Advances in Neural Information Processing Systems}, 37:\penalty0 31015--31035, 2024{\natexlab{a}}.

\bibitem[Wang et~al.(2026{\natexlab{d}})Wang, Li, Chen, Gao, Teng, and Wang]{wang2026uniddtunifyingmultimodalunderstanding}
[127] Wang, S., Li, L., Chen, Y., Gao, R., Teng, Y., and Wang, L.
\newblock Uniddt: Unifying multimodal understanding and generation with decoupled diffusion transformer, 2026{\natexlab{d}}.
\newblock URL \url{https://arxiv.org/abs/2606.16255}.

\bibitem[Wang et~al.(2025{\natexlab{b}})Wang, Isola, and Cheung]{wang2025words}
[128] Wang, S.~L., Isola, P., and Cheung, B.
\newblock Words that make language models perceive.
\newblock \emph{arXiv preprint arXiv:2510.02425}, 2025{\natexlab{b}}.

\bibitem[Wang et~al.(2024{\natexlab{b}})Wang, Zhang, Luo, Sun, Cui, Wang, Zhang, Wang, Li, Yu, et~al.]{emu3}
[129] Wang, X., Zhang, X., Luo, Z., Sun, Q., Cui, Y., Wang, J., Zhang, F., Wang, Y., Li, Z., Yu, Q., et~al.
\newblock Emu3: Next-token prediction is all you need.
\newblock \emph{arXiv preprint arXiv:2409.18869}, 2024{\natexlab{b}}.

\bibitem[Wang et~al.(2026{\natexlab{e}})Wang, Lin, Yang, Zhao, Xiao, He, Zhao, Ding, Wang, Wang, Zhang, Fan, and Liu]{wang2026representation}
[130] Wang, Y., Lin, Z., Yang, C., Zhao, Y., Xiao, F., He, H., Zhao, Q., Ding, Z., Wang, F., Wang, S., Zhang, Y., Fan, H., and Liu, X.
\newblock Representation forcing for bottleneck-free unified multimodal models.
\newblock \emph{arXiv preprint arXiv:2604.21921}, 2026{\natexlab{e}}.

\bibitem[Wang et~al.(2025{\natexlab{c}})Wang, Chen, Gou, Li, Deng, Zhu, Li, Yu, Tu, Fan, et~al.]{wang2025lightbagel}
[131] Wang, Z., Chen, Z., Gou, C., Li, F., Deng, C., Zhu, D., Li, K., Yu, W., Tu, H., Fan, H., et~al.
\newblock Lightbagel: A light-weighted, double fusion framework for unified multimodal understanding and generation.
\newblock \emph{arXiv preprint arXiv:2510.22946}, 2025{\natexlab{c}}.

\bibitem[Wang et~al.(2026{\natexlab{f}})Wang, Zhang, Ge, Lian, Fu, Dunlap, Goldberg, Wang, Stoica, Chan, et~al.]{wang2026visgym}
[132] Wang, Z., Zhang, J., Ge, J., Lian, L., Fu, L., Dunlap, L., Goldberg, K., Wang, X., Stoica, I., Chan, D.~M., et~al.
\newblock Visgym: Diverse, customizable, scalable environments for multimodal agents.
\newblock \emph{arXiv preprint arXiv:2601.16973}, 2026{\natexlab{f}}.

\bibitem[Wei et~al.(2025)Wei, Liu, Ye, Wang, Wang, Wan, Gai, and Chen]{wei2025univideo}
[133] Wei, C., Liu, Q., Ye, Z., Wang, Q., Wang, X., Wan, P., Gai, K., and Chen, W.
\newblock Univideo: Unified understanding, generation, and editing for videos.
\newblock \emph{arXiv preprint arXiv:2510.08377}, 2025.

\bibitem[Wen et~al.(2026)Wen, Li, Zhang, Lei, Chen, Fan, Zhang, Wang, Qiu, Li, et~al.]{wen2026unig2u}
[134] Wen, Z., Li, B., Zhang, W., Lei, J., Chen, X., Fan, Y., Zhang, Q., Wang, Y., Qiu, L., Li, B., et~al.
\newblock Unig2u-bench: Do unified models advance multimodal understanding?
\newblock \emph{arXiv preprint arXiv:2603.03241}, 2026.

\bibitem[Wu et~al.(2025{\natexlab{a}})Wu, Chen, Wu, Ma, Liu, Pan, Liu, Xie, Yu, Ruan, et~al.]{janus}
[135] Wu, C., Chen, X., Wu, Z., Ma, Y., Liu, X., Pan, Z., Liu, W., Xie, Z., Yu, X., Ruan, C., et~al.
\newblock Janus: Decoupling visual encoding for unified multimodal understanding and generation.
\newblock In \emph{CVPR}, 2025{\natexlab{a}}.

\bibitem[Wu et~al.(2025{\natexlab{b}})Wu, Li, Zhou, Lin, Gao, Yan, Yin, Bai, Xu, Chen, et~al.]{wu2025qwen}
[136] Wu, C., Li, J., Zhou, J., Lin, J., Gao, K., Yan, K., Yin, S.-m., Bai, S., Xu, X., Chen, Y., et~al.
\newblock Qwen-image technical report.
\newblock \emph{arXiv preprint arXiv:2508.02324}, 2025{\natexlab{b}}.

\bibitem[Wu et~al.(2026)Wu, Wu, Wang, Wu, Ou, and Yu]{wu2026scalingnativemultimodalpretraining}
[137] Wu, H., Wu, A., Wang, H., Wu, J., Ou, J., and Yu, B.
\newblock Scaling native multimodal pre-training from scratch, 2026.
\newblock URL \url{https://arxiv.org/abs/2607.22043}.

\bibitem[Wu et~al.(2024{\natexlab{a}})Wu, Jiang, Ma, Liu, Zhao, Yuan, Bai, and Bai]{wu2024liquid}
[138] Wu, J., Jiang, Y., Ma, C., Liu, Y., Zhao, H., Yuan, Z., Bai, S., and Bai, X.
\newblock Liquid: Language models are scalable and unified multi-modal generators.
\newblock \emph{arXiv preprint arXiv:2412.04332}, 2024{\natexlab{a}}.

\bibitem[Wu et~al.(2025{\natexlab{c}})Wu, Xiong, Li, Xia, Wang, Wang, Yu, Kim, Rossi, Yao, et~al.]{wu2025mitigating}
[139] Wu, J., Xiong, Y., Li, X., Xia, Y., Wang, R., Wang, Y., Yu, T., Kim, S., Rossi, R.~A., Yao, L., et~al.
\newblock Mitigating visual knowledge forgetting in mllm instruction-tuning via modality-decoupled gradient descent.
\newblock \emph{arXiv preprint arXiv:2502.11740}, 8, 2025{\natexlab{c}}.

\bibitem[Wu et~al.(2024{\natexlab{b}})Wu, Zhang, Chen, Tang, Li, Fang, Zhu, Xie, Yin, Yi, et~al.]{wu2024vila}
[140] Wu, Y., Zhang, Z., Chen, J., Tang, H., Li, D., Fang, Y., Zhu, L., Xie, E., Yin, H., Yi, L., et~al.
\newblock Vila-u: a unified foundation model integrating visual understanding and generation.
\newblock \emph{arXiv preprint arXiv:2409.04429}, 2024{\natexlab{b}}.

\bibitem[xAI(2024)]{grok}
[141] xAI.
\newblock grok, 2024.

\bibitem[Xiao et~al.(2025)Xiao, Song, Chen, Luo, Chen, Gan, Huang, Li, Qi, and Shan]{xiao2025mindomni}
[142] Xiao, Y., Song, L., Chen, Y., Luo, Y., Chen, Y., Gan, Y., Huang, W., Li, X., Qi, X., and Shan, Y.
\newblock Mindomni: Unleashing reasoning generation in vision language models with rgpo.
\newblock In \emph{NeurIPS}, 2025.

\bibitem[Xie et~al.(2025)Xie, Mao, Bai, Zhang, Wang, Lin, Gu, Chen, Yang, and Shou]{showo}
[143] Xie, J., Mao, W., Bai, Z., Zhang, D.~J., Wang, W., Lin, K.~Q., Gu, Y., Chen, Z., Yang, Z., and Shou, M.~Z.
\newblock Show-o: One single transformer to unify multimodal understanding and generation.
\newblock In \emph{ICLR}, 2025.

\bibitem[Xin et~al.(2025)Xin, Qin, Luo, Zhu, Yan, Tai, Lei, Cao, Wang, Wang, et~al.]{xin2025lumina}
[144] Xin, Y., Qin, Q., Luo, S., Zhu, K., Yan, J., Tai, Y., Lei, J., Cao, Y., Wang, K., Wang, Y., et~al.
\newblock Lumina-dimoo: An omni diffusion large language model for multi-modal generation and understanding.
\newblock \emph{arXiv preprint arXiv:2510.06308}, 2025.

\bibitem[Xu et~al.(2025)Xu, Yin, and Chen]{xu2025tbac}
[145] Xu, J., Yin, Y., and Chen, X.
\newblock Tbac-uniimage: Unified understanding and generation by ladder-side diffusion tuning.
\newblock \emph{arXiv preprint arXiv:2508.08098}, 2025.

\bibitem[Yang et~al.(2026{\natexlab{a}})Yang, Lin, Zhao, Xiao, He, Zhao, Deng, Li, Ding, Guo, et~al.]{yang2026omni}
[146] Yang, C., Lin, Z., Zhao, Y., Xiao, F., He, H., Zhao, Q., Deng, C., Li, K., Ding, Z., Guo, Y., et~al.
\newblock Context unrolling in omni models.
\newblock \emph{arXiv preprint arXiv:2604.21921}, 2026{\natexlab{a}}.

\bibitem[Yang et~al.(2026{\natexlab{b}})Yang, Shi, Shui, Wu, Tao, Wang, Lee, Liu, Ma, and Berg-Kirkpatrick]{yang2026reasoning}
[147] Yang, C., Shi, C., Shui, B., Wu, Y., Tao, M., Wang, H., Lee, I.~Y., Liu, Y., Ma, X., and Berg-Kirkpatrick, T.
\newblock From reasoning to pixels: Benchmarking the alignment gap in unified multimodal models.
\newblock \emph{arXiv preprint arXiv:2602.08336}, 2026{\natexlab{b}}.

\bibitem[Yu et~al.(2026)Yu, Xu, Huang, Xue, Huang, Duan, and Zhao]{yu2026rae}
[148] Yu, H., Xu, H., Huang, J., Xue, Z., Huang, H., Duan, N., and Zhao, F.
\newblock Rae-ar: Taming autoregressive models with representation autoencoders.
\newblock \emph{arXiv preprint arXiv:2604.01545}, 2026.

\bibitem[Yue et~al.(2024)Yue, Ni, Zhang, Zheng, Liu, Zhang, Stevens, Jiang, Ren, Sun, et~al.]{yue2023mmmu}
[149] Yue, X., Ni, Y., Zhang, K., Zheng, T., Liu, R., Zhang, G., Stevens, S., Jiang, D., Ren, W., Sun, Y., et~al.
\newblock Mmmu: A massive multi-discipline multimodal understanding and reasoning benchmark for expert agi.
\newblock In \emph{CVPR}, 2024.

\bibitem[Yue et~al.(2026)Yue, Zhang, Zeng, Chen, Wang, Zhuang, Dong, Du, Wang, Wang, and Wang]{uniflow}
[150] Yue, Z., Zhang, H., Zeng, X., Chen, B., Wang, C., Zhuang, S., Dong, L., Du, K., Wang, Y., Wang, L., and Wang, Y.
\newblock Uniflow: A unified pixel flow tokenizer for visual understanding and generation.
\newblock In \emph{ICLR}, 2026.

\bibitem[Zellers et~al.(2019)Zellers, Holtzman, Bisk, Farhadi, and Choi]{zellers2019hellaswag}
[151] Zellers, R., Holtzman, A., Bisk, Y., Farhadi, A., and Choi, Y.
\newblock Hellaswag: Can a machine really finish your sentence?
\newblock \emph{arXiv preprint arXiv:1905.07830}, 2019.

\bibitem[Zhai et~al.(2023)Zhai, Mustafa, Kolesnikov, and Beyer]{siglip}
[152] Zhai, X., Mustafa, B., Kolesnikov, A., and Beyer, L.
\newblock Sigmoid loss for language image pre-training.
\newblock In \emph{Proceedings of the IEEE/CVF international conference on computer vision}, pp.\  11975--11986, 2023.

\bibitem[Zhai et~al.(2024)Zhai, Tong, Li, Cai, Qu, Lee, and Ma]{zhai2023investigating}
[153] Zhai, Y., Tong, S., Li, X., Cai, M., Qu, Q., Lee, Y.~J., and Ma, Y.
\newblock Investigating the catastrophic forgetting in multimodal large language models.
\newblock In \emph{CPAL}, 2024.

\bibitem[Zhang et~al.(2026{\natexlab{a}})Zhang, Qiu, Cui, Song, Li, Li, Huang, Zhang, Li, Wu, et~al.]{zhang2026hydra}
[154] Zhang, G., Qiu, X., Cui, Y., Song, T., Li, C., Li, J., Huang, T., Zhang, X., Li, Y., Wu, J., et~al.
\newblock Hydra-x: Native unified multimodal models with holistic visual tokenizers.
\newblock \emph{arXiv preprint arXiv:2606.13289}, 2026{\natexlab{a}}.

\bibitem[Zhang et~al.(2026{\natexlab{b}})Zhang, Qu, Liu, Chen, Song, Dong, Sun, Li, Wang, Jiang, et~al.]{zhang2026nextflow}
[155] Zhang, H., Qu, L., Liu, Y., Chen, H., Song, Y., Dong, Y., Sun, S., Li, X., Wang, X., Jiang, Y., et~al.
\newblock Nextflow: Unified sequential modeling activates multimodal understanding and generation.
\newblock \emph{arXiv preprint arXiv:2601.02204}, 2026{\natexlab{b}}.

\bibitem[Zhang et~al.(2025)Zhang, Li, Li, Yang, and Cheng]{zhang2025unified}
[156] Zhang, J., Li, T., Li, L., Yang, Z., and Cheng, Y.
\newblock Are unified vision-language models necessary: Generalization across understanding and generation.
\newblock \emph{arXiv preprint arXiv:2505.23043}, 2025.

\bibitem[Zhao et~al.(2025)Zhao, Xue, Reed, Fan, Zhu, Kautz, Yu, Kr{\"a}henb{\"u}hl, and Huang]{zhao2025qlip}
[157] Zhao, Y., Xue, F., Reed, S., Fan, L., Zhu, Y., Kautz, J., Yu, Z., Kr{\"a}henb{\"u}hl, P., and Huang, D.-A.
\newblock Qlip: Text-aligned visual tokenization unifies auto-regressive multimodal understanding and generation.
\newblock \emph{arXiv preprint arXiv:2502.05178}, 2025.

\bibitem[Zheng et~al.(2025)Zheng, Wang, Zhao, Deng, Wang, Zhang, and Qi]{zheng2025hita}
[158] Zheng, A., Wang, H., Zhao, Y., Deng, W., Wang, T., Zhang, X., and Qi, X.
\newblock Hita: Holistic tokenizer for autoregressive image generation.
\newblock \emph{arXiv preprint arXiv:2507.02358}, 2025.

\bibitem[Zheng et~al.(2026)Zheng, Ma, Tong, and Xie]{zheng2025diffusion}
[159] Zheng, B., Ma, N., Tong, S., and Xie, S.
\newblock Diffusion transformers with representation autoencoders.
\newblock In \emph{ICLR}, 2026.

\bibitem[Zhou et~al.(2019)Zhou, Zhao, Puig, Xiao, Fidler, Barriuso, and Torralba]{zhou2019semantic}
[160] Zhou, B., Zhao, H., Puig, X., Xiao, T., Fidler, S., Barriuso, A., and Torralba, A.
\newblock Semantic understanding of scenes through the ade20k dataset.
\newblock \emph{IJCV}, 2019.

\bibitem[Zhou et~al.(2025)Zhou, Yu, Babu, Tirumala, Yasunaga, Shamis, Kahn, Ma, Zettlemoyer, and Levy]{zhou2024transfusion}
[161] Zhou, C., Yu, L., Babu, A., Tirumala, K., Yasunaga, M., Shamis, L., Kahn, J., Ma, X., Zettlemoyer, L., and Levy, O.
\newblock Transfusion: Predict the next token and diffuse images with one multi-modal model.
\newblock In \emph{ICLR}, 2025.

\bibitem[Ziyin \& Chuang(2025)Ziyin and Chuang]{ziyin2025proof}
[162] Ziyin, L. and Chuang, I.
\newblock Proof of a perfect platonic representation hypothesis.
\newblock \emph{arXiv preprint arXiv:2507.01098}, 2025.

\end{thebibliography}

\end{document}